\pdfoutput=1
\documentclass{article} 
\usepackage{colm2024_conference}
\usepackage[bottom]{footmisc}

\usepackage{microtype}
\usepackage{hyperref}
\usepackage{xurl}
\usepackage{tikz}
\usepackage{booktabs}
\definecolor{darkblue}{rgb}{0, 0, 0.5}
\hypersetup{colorlinks=true, citecolor=darkblue, linkcolor=darkblue, urlcolor=darkblue}

\usepackage{color}
\usepackage{makecell}
\usepackage{multicol}
\usepackage{multirow}
\usepackage{enumitem}
\usepackage{amssymb}
\usepackage{pifont}
\usepackage{tikz}
\usetikzlibrary{positioning, shapes, arrows}

\usepackage{pdflscape}
\usepackage{longtable} 
\usetikzlibrary{mindmap,trees} 
\usepackage[edges]{forest}
\usepackage{bm}
\usepackage{rotating}
\usepackage{float}

\definecolor{mygreen}{RGB}{9,136,66}
\newcommand{\cmark}{\textcolor{mygreen}{\ding{51}}}%
\newcommand{\xmark}{\textcolor{red!50}{\ding{55}}}%
\newcommand{\omark}{\textcolor{blue}{\ding{109}}}%

\title{Thus Spake Long-Context Large Language Model}

\colmfinalcopy

\author{%
Xiaoran Liu\textsuperscript{1,2,3}\thanks{\ \ Equal contribution.}, 
Ruixiao Li\textsuperscript{2,3}\footnotemark[1], 
Mianqiu Huang\textsuperscript{2}\footnotemark[1], 
Zhigeng Liu\textsuperscript{2,3}\footnotemark[1], 
Yuerong Song\textsuperscript{2,3}\footnotemark[1], \\ 
\textbf{Qipeng Guo\textsuperscript{1,3\dag}, 
Siyang He\textsuperscript{2,3}, 
Qiqi Wang\textsuperscript{2,3}, 
Linlin Li\textsuperscript{4}, 
Qun Liu\textsuperscript{4}, 
} \\
\textbf{
Ziwei He\textsuperscript{3},
Yaqian Zhou\textsuperscript{2}, 
Xuanjing Huang\textsuperscript{2}, 
Xipeng Qiu\textsuperscript{2,3}\thanks{\ \ Corresponding Author.} %
}\\[.5ex]
\textsuperscript{1}Shanghai AI Lab, \ %
\textsuperscript{2}School of Computer Science, Fudan University, \ %
\\
\textsuperscript{3}Shanghai Innovation Institute,  %
\textsuperscript{4}Huawei Noah's Ark Lab \ %
\\[.5ex]
\texttt{xrliu24@m.fudan.edu.cn}, \ \texttt{guoqipeng@pjlab.org.cn}, \ 
\texttt{xpqiu@fudan.edu.cn}\\
}

\begin{document}

\maketitle

\begin{abstract}
Long context is an important topic in Natural Language Processing (NLP), running through the development of NLP architectures, and offers immense opportunities for Large Language Models (LLMs), giving LLMs the lifelong learning potential akin to humans. Unfortunately, the pursuit of a long context is accompanied by numerous obstacles. Nevertheless, long context remains a core competitive advantage for LLMs. In the past two years, the context length of LLMs has achieved a breakthrough extension to millions of tokens. Moreover, research on long-context LLMs has expanded beyond length extrapolation to a comprehensive focus on architecture, infrastructure, training, and evaluation technologies. \\[1ex]
Inspired by the symphonic poem, \textit{Thus Spake Zarathustra}, we draw an analogy between the journey of extending the context of LLM and the attempts of humans to transcend their mortality. In this survey, we will illustrate how LLM struggles between the tremendous need for a longer context and its equal need to accept the fact that it is ultimately finite. To achieve this, we give a global picture of the lifecycle of long-context LLMs from four perspectives: architecture, infrastructure, training, and evaluation, showcasing the full spectrum of long-context technologies. At the end of this survey, we will present 10 unanswered questions currently faced by long-context LLMs. We hope this survey can serve as a systematic introduction to research on long-context LLMs.
\end{abstract}
\tikzstyle{my-box}=[
	rectangle,
	draw=black, 
	rounded corners,
	text opacity=1,
	minimum height=1.5em,
	minimum width=5em,
	inner sep=2pt,
	align=center,
	fill opacity=.5,
	line width=0.8pt,
]
\tikzstyle{leaf}=[my-box, minimum height=1.5em,
	fill=hidden-pink!80, text=black, align=left,font=\normalsize,
	inner xsep=2pt,
	inner ysep=4pt,
	line width=0.8pt,
]

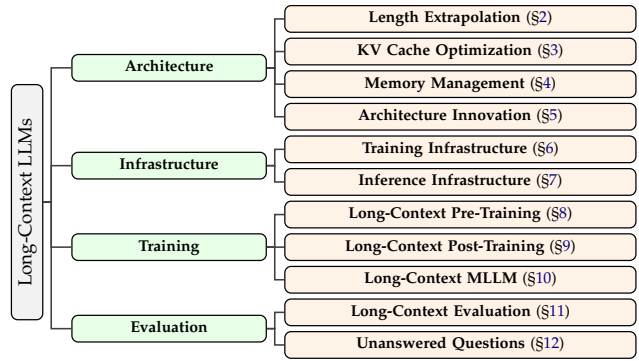
\begin{figure*}[b]
	\centering
	\resizebox{0.6\textwidth}{!}{
		\begin{forest}
			forked edges,
			for tree={
				grow=east,
				reversed=true,
				anchor=base west,
				parent anchor=east,
				child anchor=west,
				base=center,
				font=\large,
				rectangle,
				draw=black, 
				rounded corners,
				align=left,
				text centered,
				minimum width=4em,
				edge+={darkgray, line width=1pt},
				s sep=3pt,
				inner xsep=2pt,
				inner ysep=3pt,
				line width=0.8pt,
				ver/.style={rotate=90, child anchor=north, parent anchor=south, anchor=center, minimum width=15em, fill=gray!10},
			},
			where level=1{text width=12em,font=\normalsize,fill=green!10}{},
			where level=2{text width=22em, align=left, font=\normalsize, fill=orange!10}{},
			where level=3{ }{},
			where level=4{text width=18em,font=\normalsize,}{},
			where level=5{text width=18em,font=\normalsize,}{},
			[
				Long-Context LLMs, ver
				[
					\textbf{Architecture}
					[
						\textbf{Length Extrapolation} (\S\ref{sec2})
					]
					[
						\textbf{KV Cache Optimization} (\S\ref{sec3})
					]
					[
						\textbf{Memory Management} (\S \ref{sec4})
					]
					[
						\textbf{Architecture Innovation} (\S \ref{sec5})
					]
				]
				[
					\textbf{Infrastructure}
					[
						\textbf{Training Infrastructure} (\S \ref{sec6})
					]
					[
						\textbf{Inference Infrastructure} (\S \ref{sec7})
					]
				]
				[
					\textbf{Training}
					[
						\textbf{Long-Context Pre-Training} (\S \ref{sec8})
					]
					[
						\textbf{Long-Context Post-Training} (\S \ref{sec9})
					]
					[
						\textbf{Long-Context MLLM} (\S \ref{sec10})
					]
				]
				[
					\textbf{Evaluation}
					[
						\textbf{Long-Context Evaluation} (\S \ref{sec11})
					]
					[
						\textbf{Unanswered Questions} (\S \ref{sec12})
					]
				]
			]
		\end{forest}
	}
	\caption{An overview of \emph{Thus Spake Long-Context Large Language Model}.}
	\label{fig:overall}
\end{figure*}

\; \; \; \; Video: \url{https://www.bilibili.com/video/BV11h9AYoEYj}. 

\; \; \; \; Github: \url{https://github.com/OpenMOSS/Thus-Spake-Long-Context-LLM}.

\newpage

\section{Introduction}



Research on long-context capability has been an important topic in Natural Language Processing (NLP), reflected in the evolutionary trajectory of mainstream architectures. This evolution shows a consistent progression toward increasing context length, from the Bag-of-Word models~\citep{Harris01081954} with no concept of context to CNNs~\citep{lecun1995convolutional} with local receptive fields, then to LSTMs~\citep{schmidhuber1997long} characterized with an explicit long short-term memory, and currently to Transformer featured with modeling long-range dependencies~\citep{Vaswani2017attention}, as well as the recent discussions on the SSM-Mamba series~\citep{gu2020hippo,gu2023mamba,daotransformers} that challenges the dominance of Transformers from the perspective of history storage. Researchers hope models, especially the Large Language Model (LLM)~\citep{gpt4,Sun2024MOSS,reid2024gemini,meta2024introducing}, can possess life-long context, rather than being limited by a fixed window size.

In a 1k context, LLM may only understand a short fairy tale. In a 4k context, the reading comprehension may be limited to an arXiv paper~\citep{shaham2022scrolls}. In a 32k to 128k context, LLM may process a hundreds-page detective novel in its entirety and successfully infer the identity of the murderer~\citep{xu2024detectiveqa,wang2024novelqa}. When the context length extends to 512k, even a novel as lengthy as Ulysses or a novel series could be input and understood as a whole~\citep{jacobs2023deepspeed}. When the context length reaches 2M, the model may learn new knowledge through many-shot long In-Context Learning (ICL)~\citep{agarwal2024many} or acquire a new language via vocabulary and grammar books~\citep{reid2024gemini}. If the context length becomes infinite, LLM may possess life-long learning capabilities, which may change the existing training paradigm~\citep{sun2024learning,lin2023unlocking}.

Unfortunately, as context length increases, researchers also face various obstacles. From an architectural perspective, the context length of mainstream Transformer architectures is limited not only by the pre-training window size~\citep{presstrain,chen2023extending} but also by the memory and computational overhead of the Key-Value (KV) cache~\citep{kwon2023efficient,xiaoefficient}. From an infrastructural perspective, longer contexts result in greater memory pressure and lower throughput~\citep{chen2024internevo,kwon2023efficient}. From a training perspective, long-context datasets face challenges in both quantity and quality~\citep{lv2024longwanjuan,gao2024train}. From an evaluation perspective, increasing context length reveals more potential problems in LLMs~\citep{agarwal2024many,hsieh2024ruler}, leading to higher requirements for LLM performance~\citep{xu2024detectiveqa,zhang2023movqa}.

\begin{figure}[!t]
    \centering
    \includegraphics[width=\linewidth]{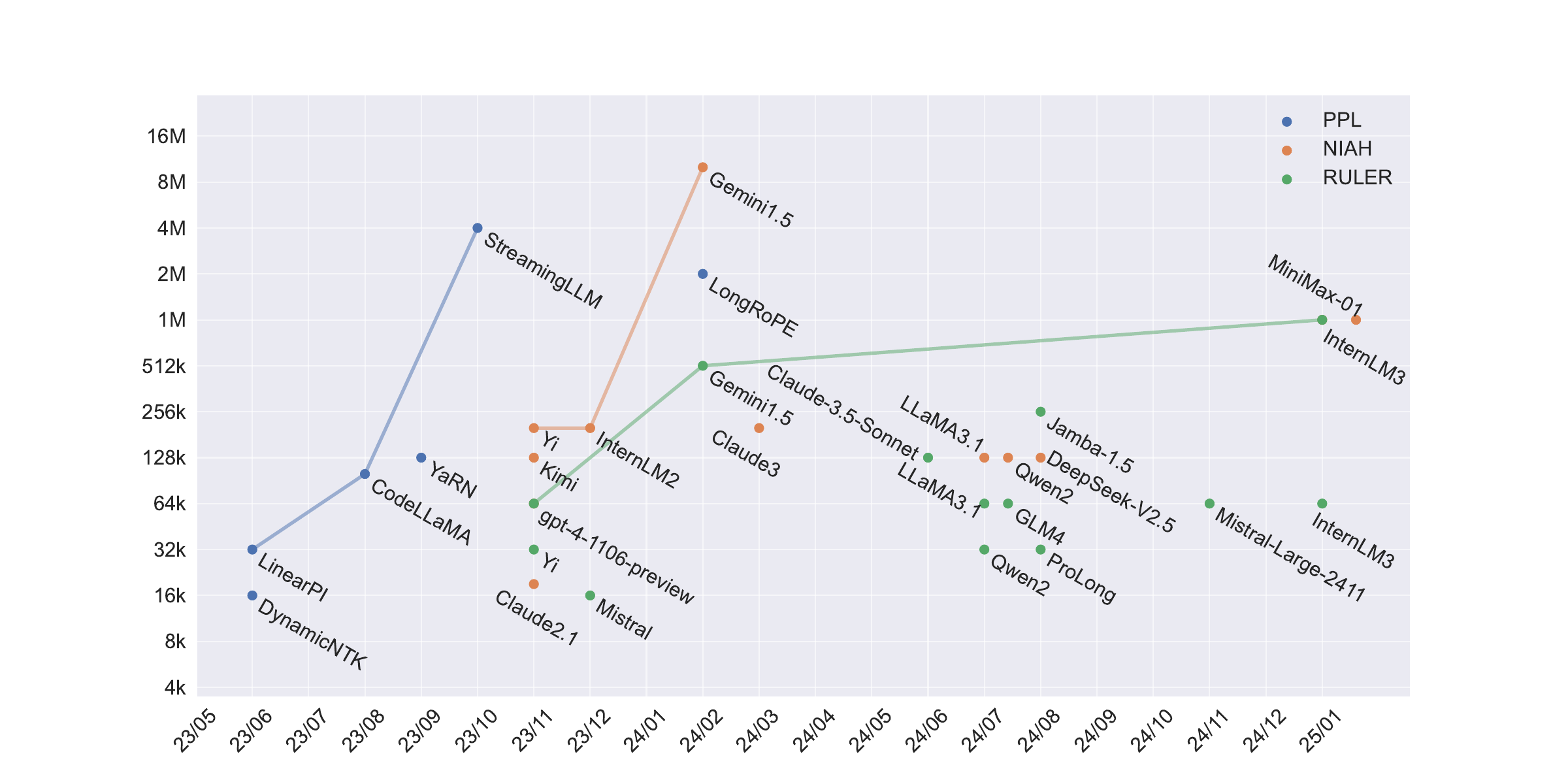}
    \caption{Long-context performance of various LLMs across multiple benchmarks, perplexity (PPL)~\citep{presstrain}, NIAH~\citep{niah}, and RULER~\citep{hsieh2024ruler}. The horizontal axis represents the release time, while the vertical axis indicates the effective context length achieved by the LLMs on the corresponding task. The line associated with each task represents the state-of-the-art performance at a given point in time.}
    \label{fig:eval-ctx}
\end{figure}

However, since the emergence of LLM, long-context capabilities remain one of the most rapidly developing areas and constitute a core competition point~\citep{anthropic2024claude2,cai2024internlm2,meta2024introducing}, as shown in Figure~\ref{fig:eval-ctx}. From April 2023 to February 2024, the context length of open-source LLMs has grown from an initial 2k~\citep{touvron2023llama} to 2M~\citep{ding2024longrope}. In this process, some surveys concentrate on particular aspects~\citep{huang2023advancing,zhao2023length,pawar2024and}, particularly developments in architectural design, while other technical reports focus on summarizing the life-cycle of a specific long-context LLM~\citep{chatglm2024glmlong,gao2024train}, from data construction to context extension and to performance evaluation. Currently, there is a lack of a comprehensive survey that presents the full life-cycle of long-context LLMs from architecture, infrastructure, training, and evaluation, showing the global picture of long-context technology.

Inspired by \textit{Thus Spake Zarathustra}, the symphonic poem of the German composer Richard Strauss, we draw an analogy between the attempts of LLMs to extend their context lengths and the attempts of humans to transcend their mortality. On the journey of extending the context length of LLMs, researchers continuously challenge the boundaries of context through optimizations in architecture, infrastructure, and training, much like \textit{the struggle between man's tremendous need for immortality and his equal need to accept the fact that he is mortal}\footnote{From \textit{Thus Spake Richard Strauss} by Leonard Bernstein in Young People's Concert. \url{https://leonardbernstein.com/lectures/television-scripts/young-peoples-concerts/thus-spake-richard-strauss}}. As shown in Figure~\ref{fig:overall}, this survey comprehensively introduces the life-cycle of long-context LLMs from four perspectives: \textbf{architecture}, \textbf{infrastructure}, \textbf{training}, and \textbf{evaluation}. 
\begin{itemize}
    \item Sections \ref{sec2} to \ref{sec5} focus on the architectural aspect, discussing the enhancement of Transformer in length extrapolation, KV cache optimization, as well as memory management, and the innovation to defeat Transformer by long-context researchers.
    \item Sections \ref{sec6} and \ref{sec7} address infrastructure considerations, detailing optimizations for long context in the training and inference phases of Transformer-based LLMs.
    \item Sections \ref{sec8} to \ref{sec10} introduce the training methods in three corresponding training stages for long-context LLMs, pre-training, post-training, and multi-modal training, particularly for long-context Multi-modal LLM (MLLM).
    \item In Section \ref{sec11}, we will discuss the long-context evaluation. In Section \ref{sec12}, we will outline 10 unanswered questions that long-context LLMs still face as a conclusion.
\end{itemize}

We hope our survey provides a comprehensive technical summary for the long-context research community and serves as an introductory guide for researchers unfamiliar with this area. To present this paper more intuitively, we have made a video that combines the content of this survey with the symphonic poem \textit{Thus Spake Zarathustra}, aiming to raise awareness among more researchers on the importance and entirety of long-context research. The video is available at \url{https://www.bilibili.com/video/BV11h9AYoEYj}.

\section{Length Extrapolation}\label{sec2}

\tikzstyle{my-box}=[
	rectangle,
	draw=black, 
	rounded corners,
	text opacity=1,
	minimum height=1.5em,
	minimum width=5em,
	inner sep=2pt,
	align=center,
	fill opacity=.5,
	line width=0.8pt,
]
\tikzstyle{leaf}=[my-box, minimum height=1.5em,
	 text=black, align=left,font=\normalsize,
	inner xsep=10pt,
	inner ysep=4pt,
	line width=0.8pt,
]

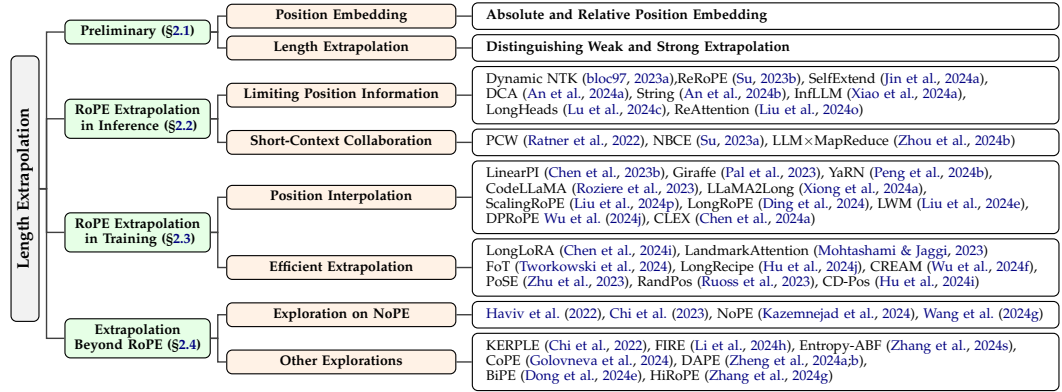
\begin{figure*}[t!]
	\centering
	\resizebox{\textwidth}{!}{
		\begin{forest}
			forked edges,
			for tree={
				grow=east,
				reversed=true,
				anchor=base west,
				parent anchor=east,
				child anchor=west,
				base=center,
				font=\large,
				rectangle,
				draw=black, 
				rounded corners,
				align=left,
				text centered,
				minimum width=4em,
				edge+={darkgray, line width=1pt},
				s sep=3pt,
				inner xsep=2pt,
				inner ysep=3pt,
				line width=0.8pt,
				ver/.style={rotate=90, child anchor=north, parent anchor=south, anchor=center, minimum width=19em, fill=gray!10},
			},
			where level=1{text width=10em, align=center, font=\normalsize,fill=green!10}{},
			where level=2{text width=16em, align=center, font=\normalsize, fill=orange!10}{},
			where level=3{text width=40em, align=left}{},
			where level=4{text width=18em,font=\normalsize,}{},
			where level=5{text width=18em,font=\normalsize,}{},
			[
				\textbf{Length Extrapolation}, ver
				[
                        \textbf{Preliminary~(\S\ref{sec2_1})}
                        [
                            \textbf{Position Embedding}
                            [
                                \textbf{Absolute and Relative Position Embedding}, leaf
                            ]
                        ]
                        [
                            \textbf{Length Extrapolation}
                            [
                                \textbf{Distinguishing Weak and Strong Extrapolation}, leaf
                            ]
                        ]
                    ]
                    [
                        \textbf{RoPE Extrapolation}\\\newline\textbf{in Inference~(\S\ref{sec2_2})}
                        [
                            \textbf{Limiting Position Information}
                            [
                                Dynamic NTK~\citep{dynamicNTK}{,}ReRoPE~\citep{rerope}{,} SelfExtend~\citep{jin2024llm}{,} \\ DCA~\citep{an2024training}{,} 
                                String~\citep{an2024does}{,} InfLLM~\citep{xiao2024infllm}{,} \\ LongHeads~\citep{lu2024longheads}{,} ReAttention~\citep{liu2024reattention}, leaf
                            ]
                        ]
                        [
                            \textbf{Short-Context Collaboration}
                            [
                                PCW~\citep{ratner2022parallel}{,} NBCE~\citep{kexuefm9617}{,} LLM$\times$MapReduce~\citep{zhou2024llm}, leaf
                            ]
                        ]
                    ]
                    [
                        \textbf{RoPE Extrapolation}\\\newline\textbf{in Training~(\S\ref{sec2_3})}
                        [
                            \textbf{Position Interpolation}
                            [
                                LinearPI~\citep{chen2023extending}{,} Giraffe~\citep{pal2023giraffe}{,} YaRN~\citep{pengyarn}{,} \\ CodeLLaMA~\citep{roziere2023code}{,} LLaMA2Long~\citep{xiong2024effective}{,} \\ ScalingRoPE~\citep{liuscaling}{,} LongRoPE~\citep{ding2024longrope}{,} LWM~\citep{liu2024world}{,} \\
                                DPRoPE~\citet{wu2024extending}{,} CLEX~\citep{chenclex}, leaf
                            ]
                        ]
                        [
                            \textbf{Efficient Extrapolation}
                            [
                                LongLoRA~\citep{chenlonglora}{,} LandmarkAttention~\citep{mohtashami2023random} \\ FoT~\citep{tworkowski2024focused}{,} LongRecipe~\citep{hu2024longrecipe}{,} CREAM~\citep{wuefficient}{,} \\
                                PoSE~\citep{zhu2023pose}{,} RandPos~\citep{ruoss2023randomized}{,} CD-Pos~\citep{hu2024cd}, leaf
                            ]
                        ]
                    ]
                    [
                        \textbf{Extrapolation}\\\newline\textbf{Beyond RoPE~(\S\ref{sec2_4})}
                        [
                            \textbf{Exploration on NoPE}
                            [
                                \citet{haviv2022transformer}{,} \citet{chi2023latent}{,} NoPE~\citep{kazemnejad2024impact}{,} \citet{wang2024length}, leaf
                            ]
                        ]
                        [
                            \textbf{Other Explorations}
                            [
                                KERPLE~\citep{chi2022kerple}{,} FIRE~\citep{li2023functional}{,} 
                                Entropy-ABF~\citep{zhang2024extending}{,} \\ 
                                CoPE~\citep{golovneva2024contextual}{,} DAPE~\citep{zheng2024dape, zheng2024dape2}{,} \\
                                BiPE~\citep{dong2024exploring}{,} HiRoPE~\citep{zhang2024hirope}, leaf
                            ]
                        ]
                    ]
			]
		\end{forest}
	}
	\caption{An overview of length extrapolation of long-context LLMs.}
	\label{fig:len_extp}
\end{figure*}

In this section, we start the journey of extending the context length of LLMs with length extrapolation, the foundation of long-context LLMs, as shown in Figure~\ref{fig:len_extp}. 
\begin{itemize}
    \item In \S\ref{sec2_1}, we start with some preliminary knowledge, including \textbf{\textit{position embedding}} and \textbf{\textit{the definition of length extrapolation}}. Then we focus on the length extrapolation based on the widely-used RoPE~\citep{su2024roformer}. 
    \item In the inference stage, as discussed in \S\ref{sec2_2}, the extrapolation is based on \textbf{\textit{limiting position information}} including NTK~\citep{dynamicNTK}, ReRoPE~\citep{rerope} and DCA~\citep{an2024training} or \textbf{\textit{short-context collaboration}} like PCW~\citep{ratner2022parallel}. 
    \item In the training stage, as discussed in \S\ref{sec2_3}, apart from the classical \textbf{\textit{position interpolation}} methods such as LinearPI~\citep{chen2023extending} and YaRN~\citep{pengyarn}, we highlight the discussion of \textbf{\textit{extrapolation mechanism}}~\citep{liuscaling,men2024base} and \textbf{\textit{efficient extrapolation}}~\citep{chenlonglora,zhu2023pose}.
    \item  In \S\ref{sec2_4}, we will add more discussion beyond RoPE, including \textbf{\textit{NoPE}}~\citep{kazemnejad2024impact}, other position embeddings~\citep{golovneva2024contextual,dong2024exploring} and \textbf{\textit{attention entropy}}~\citep{han2024lm,zhang2024extending} 
\end{itemize}

\subsection{Preliminary}\label{sec2_1}

\subsubsection{Position Embedding}

First introduced in \citet{Vaswani2017attention}, position embedding is a key mechanism for encoding positional information in contexts, and remains fundamental to modern LLMs' ability to process long-context. 
The evolution of position embedding begins with absolute position embedding~\citep{Vaswani2017attention}, namely embedding based on the token indices, and follows by the emergence of relative position embedding, namely embedding based on the token distance, such as \citet{shaw2018self}, T5~\citep{raffel2020exploring}, TENER~\citep{yan2019tener} and XLNET~\citep{dai2019transformer}. However, those embeddings face trade-offs between performance and computational efficiency. Later, RoPE~\citep{su2024roformer} is proposed, achieving relative position embedding through absolute position embedding, combining the advantages of both approaches and thus becoming a significant academic interest~\citep{chowdhery2023palm,touvron2023llama,touvron2023llama2,Sun2024MOSS,chen2023extending,dynamicNTK}.

RoPE~\citep{su2024roformer} introduces positional information into self-attention computation through rotary transformations. Given a position index $t$ and an embedding vector $\bm{x}=[x_0,x_1,...x_{d-1}]^T$, where $d$ is the attention head dimension, RoPE defines a complex function:
\begin{equation}\begin{gathered} \bm{A}_{m,n}=\underbrace{\bm{x}_{m}\bm{W}_{Q}\bm{R}_{\Theta,m-n}^d\bm{W}_{K}^{T}\bm{x}_{n}^{T}}_\text{relative position embedding}=\underbrace{\bm{x}_{m}\bm{W}_{Q}\bm{R}_{\Theta,m}^d\left(\bm{x}_{n}\bm{W}_{K}\bm{R}_{\Theta,n}^d\right)^T}_\text{absolute position embedding} \\[0.5ex]  
\end{gathered}\text{,}\end{equation}
where $\theta_j=\beta^{-2j/d}$, with a typical value of rotary base $\beta=10000$.

RoPE has several significant advantages. First, RoPE has solid mathematical foundations with theoretical guarantees~\citep{su2024roformer}. Second, RoPE maintains low computational complexity and eliminates the necessity for storing any position embedding matrices~\citep{su2024roformer,touvron2023llama,chowdhery2023palm}. Third, RoPE can seamlessly integrate with many attention variants, demonstrating excellent compatibility~\citep{su2024roformer}. Finally, through the following improvements, RoPE has shown strong length extrapolation capabilities~\citep{dynamicNTK,liuscaling}. Given these favorable properties of RoPE, many LLMs adopt RoPE as their position embedding~\citep{dubey2024llama,glm2024chatglm,wang2024qwen2,young2024yi,cai2024internlm2}. 

\subsubsection{Length Extrapolation}
In Transformer-XL~\citep{dai2019transformer}, discover the standard Transformer's limitations in handling sequences longer than its training length. ALiBi~\citep{presstrain} later formalizes this as \textbf{length extrapolation} or \textbf{length generalization}, a model's capacity to maintain performance when processing longer sequences during inference than during training.
Before the widespread adoption of RoPE-based linear interpolation, early approaches to length extrapolation include improvements in position embedding and sliding window mechanism~\citep{presstrain,sun2022length,ratner2022parallel}.

For example, ALiBi~\citep{presstrain,yang2023baichuan} introduces fixed attention biases that scale linearly with relative positional information, showing promising results in contexts beyond training length. Later, xPos~\citep{sun2022length,sun2023retentive} addresses the length extrapolation problem by incorporating exponential decay in attention computation and proposing BCA, a windowed attention mechanism similar to a sliding window. After LLM emerges, the sliding window mechanism is first been used for the earliest length extrapolation attempts~\citep{bai2023qwen,jiang2023mistral}. Besides, more sophisticated sliding window variants are proposed. For example, LongNet~\citep{ding2023longnet} achieves length extrapolation to 1B tokens by dilated sliding window attention. Subsequently, LM-Infinite~\citep{han2024lm} and StreamingLLM~\citep{xiaoefficient} introduce $\Lambda$-shaped masks and attention sinks respectively. These two methods preserve information from global initial tokens and local window tokens, implementing attention window truncation to reduce computational complexity while maintaining LLM's performance.

\paragraph{Distinguishing Weak and Strong Extrapolation}
It is essential to distinguish two types of extrapolation capabilities, weak extrapolation and strong extrapolation. \textit{\textbf{Weak extrapolation}} refers to maintaining perplexity across varied context lengths, while \textit{\textbf{strong extrapolation}} indicates the ability to maintain performance on actual long-context understanding and processing tasks. These capabilities can be delineated by examining which tasks maintain consistent performance across different context lengths. For instance, StreamingLLM~\citep{xiaoefficient} demonstrates effective weak extrapolation in perplexity but does not guarantee equivalent performance in practical long-context tasks, as evaluated by benchmarks including NIAH~\citep{niah} and RULER~\citep{hsieh2024ruler}. The conflict of perplexity between its failure to reflect practical context length and its wide application in the long-context research will be further analyzed in \textbf{\nameref{q3_ppl}} in Section\ref{sec12}.

This distinction is crucial, as many length extrapolation works focus only on weak extrapolation~\citep{han2024lm,xiaoefficient,ding2023longnet}. The following discussion focuses on strong extrapolation. Given LLMs' predominant use of RoPE, we first explore the extrapolation of RoPE-based LLMs. Based on implementation stages, these methods can be categorized into inference-time and training-time extrapolation.

\subsection{RoPE Extrapolation in Inference}\label{sec2_2}
At inference time, there are two feasible approaches for enabling LLMs to comprehend longer context lengths. The first approach involves constraining position embeddings during the processing of extended contexts, and the second approach implements segmented understanding where the model processes long contexts in chunks and integrates understanding across these segments.

\subsubsection{Limiting Position Information} 
In RoPE~\citep{su2024roformer}, positional information is represented through trigonometric functions of the product of index and rotary angle. To maintain this product within pre-training bounds as indices increase, approaches including limiting index growth or reducing rotary angles are proposed. Fixed or dynamic NTK methods~\citep{fixedNTK,dynamicNTK} achieve plug-and-play length extrapolation by adjusting RoPE's rotary base and have been widely adopted, while more extrapolation works in inference focus on index limitation.

ReRoPE~\citep{rerope} and SelfExtend~\citep{jin2024llm} explicitly set relative position upper bounds in RoPE to constrain positional information within pre-training ranges. Similarly, InfLLM~\citep{xiao2024infllm} and LongHeads~\citep{lu2024longheads} enable training-free processing of ultra-long sequences through block-level context storage, focusing attention on crucial blocks at the beginning, end, and middle of input text. ReAttention~\citep{liu2024reattention} implements customized operators for fine-grained KV cache retrieval across the full context, enabling plug-and-play context window expansion by at least 100 times. DCA~\citep{an2024training} innovatively decomposes long sequence attention computation into intra-block, adjacent-block, and non-adjacent block components for more efficient long text processing, while String~\citep{an2024does} further simplifies this design and improves performance.

\subsubsection{Short-context Collaboration}
Short-context Collaboration refers to a series of extrapolation methods that process long texts by splitting them into shorter segments and synthesizing the results. PCW~\citep{ratner2022parallel} ensures all processing remains within pre-training length limits by dividing sequences into multiple context segments and one task sequence. NBCE~\citep{kexuefm9617} applies Naive Bayes principles to achieve length extrapolation through independent processing of context segments with prompts. XL3M~\citep{wang2024xl3m} introduces a training-free framework handling long contexts through segmented inference, while LLM×MapReduce~\citep{zhou2024llm} adopts distributed computing concepts, processing text blocks across GPUs with specialized communication structures. Additionally, LongAgent~\citep{zhao2024longagent}, an extrapolation method in training, also employs a similar approach by introducing multi-agent collaboration, where multiple agents cooperate to process long contexts.

\subsection{RoPE Extrapolation in Training}\label{sec2_3}

\subsubsection{Position Interpolation}
Beyond extrapolation methods in inference, researchers propose numerous approaches in training that focus on leveraging short-context positional information for longer contexts through position interpolation~\citep{liuscaling, xiong2024effective}. These methods similarly address either index adjustment or rotary base scaling.

For index adjustment, LinearPI~\citep{chen2023extending} first introduces linear scaling of position indices through a scaling factor to extend context length. However, it remains limited by training length and neglects feature differences across RoPE's query and key vectors' dimensions. YaRN~\citep{pengyarn} subsequently implements dynamic scaling in middle dimensions while maintaining no interpolation in low dimensions and full interpolation in high dimensions, achieving 128k length extrapolation with 64k training. YaRN gains wide adoption in subsequent LLMs like LLaMA3.1~\citep{dubey2024llama}. Similarly, Giraffe~\citep{pal2023giraffe} achieves extrapolation by preserving high-frequency rotations while suppressing low-frequency ones. Additionally, LongRoPE~\citep{ding2024longrope} employs progressive search-based non-uniform interpolation to achieve 2M context length with 256k training.

On the other hand, many models adopt enlarged rotary angles combined with longer training lengths~\citep{roziere2023code, xiong2024effective}. This approach is widely adopted in current LLMs~\citep{cai2024internlm2, young2024yi, chatglm2024glmlong} to achieve long contexts. LWM~\citep{liu2024world} implements multi-stage scaling, gradually increasing both the rotary angle base and fine-tuning length. However, these works make specific attempts on certain context lengths and rotary bases without thoroughly investigating the extrapolation mechanism of RoPE-based LLMs. Apart from the search for mechanism, DPRoPE~\citep{wu2024extending} explores optimizing RoPE rotary angle distributions to enhance extrapolation capabilities and CLEX~\citep{chenclex} introduces neural ordinary differential equations to model continuous scaling of position embedding. 

\subsubsection{Scaling Laws}

As previously discussed, the extrapolation mechanism of RoPE-based LLMs remains a crucial question in length extrapolation research. The keys to this question are the \textbf{\textit{periodicity}} and \textbf{\textit{monotonicity}} of trigonometric functions~\citep{pengyarn,liuscaling,men2024base}. YaRN~\citep{pengyarn} first mentions the relationship between the RoPE-based extrapolation and the periodicity. Furthermore, ScalingRoPE~\citep{liuscaling} identifies a critical dimension $d_\text{extra}$, decided by the pre-training context length $T_\text{train}$ and original rotary base $\beta$, that determines the LLM's extrapolation limit, as shown in Equation \ref{equ:d_extra}. 
\begin{equation}
    d_\text{extra}=2\left\lceil\frac{d}{2}\log_\beta\frac{T_\text{train}}{2\pi}\right\rceil\text{.}
\label{equ:d_extra}\end{equation}
For dimensions before the critical dimension, their position embedding $\sin(\theta t), \cos(\theta t)$ have already experienced a complete period in pre-training and will not be out-of-distribution (OOD) in extrapolation. However, dimensions beyond that will fail to extrapolate when the product of the rotary angle and position index exceeds the range the LLM pre-trained in. Since rotary angles in RoPE are arranged exponentially~\citep{su2024roformer}, the rotary angle at the critical dimension experiences the least shrinkage in base scaling. Consequently, the position embedding at this dimension will first be OOD, making its period serve as the upper bound for extrapolation, $T_\text{extra}$, as shown in Equation \ref{equ:t_extra}.
\begin{equation}
    T_\text{extra}=2\pi\cdot\beta^{\frac{d_\text{extra}}{d}}=2\pi\cdot\beta^{\left\lceil\frac{d}{2}\log_{10000}\frac{T_\text{train}}{2\pi}\right\rceil\cdot\frac{2}{d}}\text{.}
\label{equ:t_extra}\end{equation}

\citet{liuscaling} reveals a part of the extrapolation mechanism in RoPE-based LLM, that RoPE's extrapolation represents position information in a longer context using that previously learned in short-context pre-training. However, forcing LLM to learn more position information in fine-tuning, such as reducing the rotary base, is inappropriate~\citep{men2024base}. \citet{men2024base} proves that reducing rotary bases undermines contextual information modeling because it disrupts the original patterns and overlooks the second feature, monotonicity. The $\cos(\theta t)$ maintains monotonicity locally, reflecting relative distance~\citep{wei2025videorope}. A sufficiently smaller base can prevent position embedding from OOD based on periodicity, but this sacrifices monotonicity, limiting LLMs to perceiving local semantics and performing poorly on generation and ICL tasks~\citep{liuscaling, men2024base}, showing only weak extrapolation. This reveals a contradiction in RoPE, that \textit{\textbf{dimensions with monotonicity perceivable of long dependencies are overfitted to pre-training context and cannot extrapolate, while dimensions capable of extrapolation lose monotonicity and cannot perceive long contexts}}, which will be further analyzed in \textbf{\nameref{q2_rope}} in Section\ref{sec12}.

Although \citet{liuscaling} makes a mistake on the second part, it still has a guiding significance for length extrapolation~\citep{cai2024internlm2,Apple2024AppleIntelligence}. For instance, by finding the inverse function of Equation \ref{equ:t_extra}, we can determine the minimum necessary rotary base for supporting a specific context length $T_\text{extra}$. Compared to the linear relationship between rotary base $\beta$ and $T_\text{extra}$ in Hugging Face's default dynamic NTK implementation
, Equation~\ref{equ_beta} demonstrates a power law which accounts for the extrapolation limit in the NTK approach.
\begin{equation}
    \beta =\left(\frac{T_\text{extra}}{2\pi}\right)^{\frac{d}{d_\text{extra}}}\text{.}
\label{equ_beta}\end{equation}

\subsubsection{Efficient Extrapolation}
Length extrapolation methods in training also consider achieving extrapolation effects with fewer computational resources, known as efficient extrapolation~\citep{chen2023extending, pengyarn}.  Efficient extrapolation methods can be categorized into two types, those focusing on partial contexts and those training on much shorter contexts.

\paragraph{Focusing on Partial Contexts} LongLoRA~\citep{chenlonglora} employs S$^2$-Attn with shift and grouping operations for local sparse attention while using LoRA for long-context scenarios. Zebra~\citep{song2023zebra} introduces local attention with global approximation, combining local attention windows with a global approximation for improved efficiency. LandmarkAttn~\citep{mohtashami2023random} innovatively uses landmark tokens as processing block gates, enabling inference at any context length. CREAM~\citep{wuefficient} alleviates the "middle loss" problem in long context processing through middle sampling optimization. LongRecipe~\citep{hu2024longrecipe} extracts shorter but information-dense segments by identifying tokens with significant impact in long context processing. FoT~\citep{tworkowski2024focused} extends model context length by adding memory attention mechanisms to certain transformer layers and using kNN algorithms for key-value pair retrieval.

\paragraph{Training on Much Shorter Contexts} GrowLength~\citep{jin2023growlength} applies progressive length growth during training, starting with shorter sequences and gradually increasing context length to improve training efficiency while achieving extrapolation. E$^2$-LLM~\citep{liu20242} supports longer context windows during inference by using position index scaling and offset while only requiring training on shorter sequences. FocusLLM~\citep{li2024focusllm} proposes a parallel decoding approach, reducing complexity to $1/n$ of the original by freezing initial parameters and adding minimal training parameters, improving length extrapolation capability through training on short context. PoSE~\citep{zhu2023pose}, RandPos~\citep{ruoss2023randomized}, and CD-Pos~\citep{hu2024cd} enhance model capability in processing varied input lengths by extracting smaller segments and adjusting position embeddings within these windows during training.

\subsection{Extrapolation without RoPE}\label{sec2_4}

\subsubsection{NoPE-based Extrapolation}
Research on NoPE has revealed that causal masking injects sequential constraints into the network, since each token only attends to preceding content which implicitly encodes positional information~\citep{haviv2022transformer, chi2023latent}. This observation motivates NoPE-based LLM. Experiments demonstrate that NoPE-based LLMs achieve comparable performance to traditional position embeddings in certain tasks~\citep{kazemnejad2024impact}.

However, NoPE also struggles with length extrapolation~\citep{kazemnejad2024impact, wang2024length}. Research shows that when context length exceeds the training range, NoPE's attention distribution becomes dispersed, leading to performance degradation. To address this issue, \citet{wang2024length} proposes an optimization method based on attention temperature parameters and improves length generalization capability.

\subsubsection{Other works}
NoPE challenges whether position embedding is necessary for length extrapolation. Besides, there are other discussions regarding position embedding or length extrapolation. 

\paragraph{Other Position Embedding Schema} Several studies have proposed novel position embedding schema to address length extrapolation challenges or model long context better. KERPLE~\citep{chi2022kerple} introduces a kernel-based relative position embedding. FIRE~\citep{li2023functional} improves the Transformer's generalization capability in longer contexts through progressive interpolation. DAPE (data-adaptive position embedding) and DAPE V2~\citep{zheng2024dape, zheng2024dape2} dynamically adjusts positional offset matrices based on input data. CoPE~\citep{golovneva2024contextual} allows positions to depend on context by computing attention through selectively incrementing positions incrementing positions. BiPE~\citep{dong2024exploring} combine intra-segment and inter-segment embeddings, using the former to identify positions within segments and the latter to model relationships between segments. Similarly, HiRoPE~\citep{zhang2024hirope} tries a hierarchical RoPE in long code.

\paragraph{Attention Entropy} Researchers observe that attention entropy increases with context length~\citep{han2024lm,pengyarn}, prompting several innovative solutions. Many researchers introduce scaling factors in attention logits to reduce attention entropy. ReRoPE~\citep{rerope} incorporates a dynamic scale factor $\log_Tt$ (where $T$ is the pre-training sequence length and $t$ is the input token's position index) in attention logits. YaRN~\citep{pengyarn} introduces a scale factor in attention logits. Entropy-ABF~\citep{zhang2024extending} employs a special treatment of scaling factors for the first two attention layers, based on the discovery that the first two attention layers consistently exhibited almost identical attention patterns, with only subsequent layers showing trends of attention concentration.

Beyond these two directions, \citet{dong2024exploring} proposes two training-free methods, positional vector replacement, and attention window extension, to effectively extend context length. From a memory perspective, RMT~\citep{bulatov2023scaling} also extends input context length by adding memory tokens and segment-level recursion to pre-trained LLMs.

\section{KV Cache Optimization}\label{sec3}

Although length extrapolation can theoretically extend the context length of LLMs, it is only the tip of the iceberg of long-context LLMs. In Transformer-based LLMs, the KV cache expands with the increase of context length, resulting in a great computational and memory overhead~\citep{fu2024challenges,luohekeep,xiaoefficient}. Since the size of the KV cache is determined by the product of \textbf{\textit{cached sequence length}}, \textbf{\textit{number of layers}} (\S\ref{sec3_3}), \textbf{\textit{number of KV heads}} (\S\ref{sec3_4}), \textbf{\textit{number of feature dimensions}} (\S\ref{sec3_5}), and \textbf{\textit{storage data type}} (\S\ref{sec3_6})~\citep{fu2024challenges,Venkat2024EssentialMath}, we can optimize the overhead through each of these factors as shown in Figure~\ref{fig:kv_opt}. Particularly, since the optimizations over sequence length are most discussed, we divide them into \textbf{\textit{toke dropping}}  (\S\ref{sec3_1}) and \textbf{\textit{token merging}}  (\S\ref{sec3_2}).



\subsection{Token Dropping}\label{sec3_1}

\tikzstyle{my-box}=[
	rectangle,
	draw=black, 
	rounded corners,
	text opacity=1,
	minimum height=1.5em,
	minimum width=5em,
	inner sep=2pt,
	align=center,
	fill opacity=.5,
	line width=0.8pt,
]
\tikzstyle{leaf}=[my-box, minimum height=1.5em,
	text=black, align=left,font=\normalsize,
	inner xsep=2pt,
	inner ysep=4pt,
	line width=0.8pt,
]

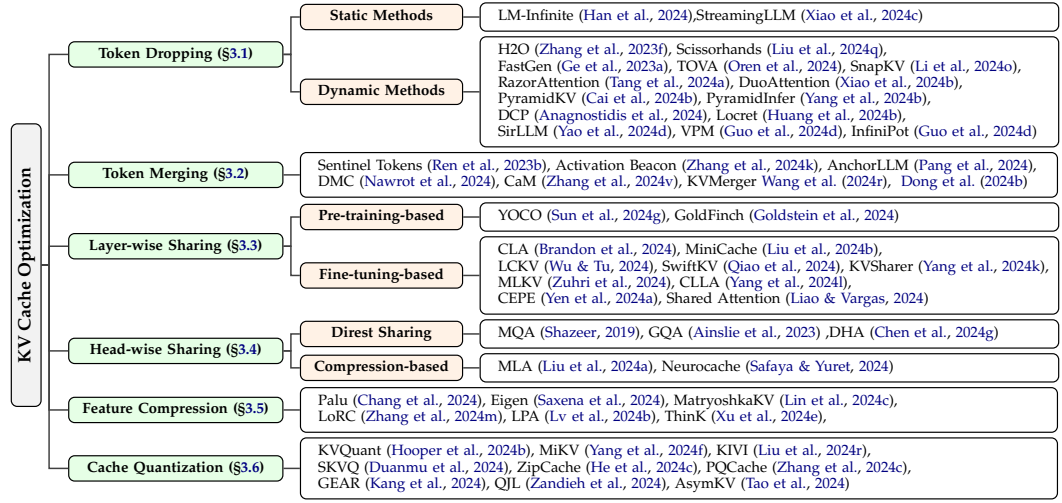
\begin{figure*}[t]
	\centering
	\resizebox{\textwidth}{!}{
		\begin{forest}
			forked edges,
			for tree={
				grow=east,
				reversed=true,
				anchor=base west,
				parent anchor=east,
				child anchor=west,
				base=center,
				font=\large,
				rectangle,
				draw=black, 
				rounded corners,
				align=left,
				text centered,
				minimum width=4em,
				edge+={darkgray, line width=1pt},
				s sep=3pt,
				inner xsep=2pt,
				inner ysep=3pt,
				line width=0.8pt,
				ver/.style={rotate=90, child anchor=north, parent anchor=south, anchor=center, minimum width=19em, fill=gray!10},
			},
			where level=1{text width=14em,font=\normalsize,fill=green!10}{},
			where level=2{text width=10.5em, align=left, font=\normalsize, fill=orange!10}{},
			where level=3{text width=38.5em}{},
			where level=4{text width=18em,font=\normalsize,}{},
			where level=5{text width=18em,font=\normalsize,}{},
			[
				\textbf{KV Cache Optimization}, ver
				[
                        \textbf{Token Dropping~(\S\ref{sec3_1})}
                        [
                            \textbf{Static Methods}
                            [
                                \quad LM-Infinite~\citep{han2024lm}{,}StreamingLLM~\citep{xiaoefficient},leaf
                            ]
                        ]
                        [
                            \textbf{Dynamic Methods}
                            [
                                \quad H2O~\citep{zhang2023h2o}{,} Scissorhands~\citep{liu2024scissorhands}{,} \\ \quad FastGen~\citep{ge2023model}{,} TOVA~\citep{oren2024transformers}{,} SnapKV~\citep{li2024snapkv}{,} \\ \quad RazorAttention~\citep{tang2024razorattention}{,} DuoAttention~\citep{xiao2024duoattention}{,} \\ \quad PyramidKV~\citep{cai2024pyramidkv}{,} PyramidInfer~\citep{yang2024pyramidinfer}{,} \\ \quad DCP~\citep{anagnostidis2024dynamic}{,} Locret~\citep{huang2024locret}{,} \\ \quad SirLLM~\citep{yao2024sirllm}{,} VPM~\citep{guo2024attention}{,} InfiniPot~\citep{guo2024attention}, leaf
                            ]
                        ]
                    ]
                    [
                        \textbf{Token Merging~(\S\ref{sec3_2})}
                        [
                            \quad Sentinel Tokens~\citep{ren2023context}{,} Activation Beacon~\citep{zhang2024long}{,} AnchorLLM~\citep{pang2024anchor}{,} \\ \quad DMC~\citep{nawrot2024dynamic}{,} CaM~\citep{zhangcam}{,} KVMerger~\cite{wang2024model}{,} ~\citet{dong2024get}, leaf, fill=white, text width=50.5em
                        ]
                    ]
                    [
                        \textbf{Layer-wise Sharing~(\S\ref{sec3_3})}
                        [
                            \textbf{Pre-training-based}
                            [
                                \quad YOCO~\citep{sun2024you}{,} GoldFinch~\citep{goldstein2024goldfinch}, leaf
                            ]
                        ]
                        [
                            \textbf{Fine-tuning-based}
                            [
                                \quad CLA~\citep{brandon2024reducing}{,} MiniCache~\citep{liu2024minicache}{,}  \\ \quad LCKV~\citep{wu2024layer}{,} SwiftKV~\citep{qiao2024swiftkv}{,} KVSharer~\citep{yang2024kvsharer}{,}  \\ \quad MLKV~\citep{zuhri2024mlkv}{,} CLLA~\citep{yang2024lossless}{,} \\ \quad CEPE~\citep{yen2024long}{,} Shared Attention~\citep{liao2024beyond},leaf
                            ]
                        ]
                    ]
                    [
                        \textbf{Head-wise Sharing~(\S\ref{sec3_4})}
                        [
                            \textbf{Direst Sharing}
                            [
                               \quad MQA~\citep{shazeer2019fast}{,} GQA~\citep{ainslie2023gqa} {,}DHA~\citep{chen2024dha}, leaf
                            ]
                        ]
                        [
                            \textbf{Compression-based}
                            [
                                \quad MLA~\citep{liu2024deepseek}{,} Neurocache~\citep{safaya2024neurocache}, leaf
                            ]
                        ]
                    ]
                    [
                        \textbf{Feature Compression~(\S\ref{sec3_5})}
                        [
                            \quad Palu~\citep{chang2024palu}{,} Eigen~\citep{saxena2024eigen}{,} MatryoshkaKV~\citep{lin2024matryoshkakv}{,} \\ \quad LoRC~\citep{zhang2024lorc}{,} LPA~\citep{lv2024scalable}{,} ThinK~\citep{xu2024think}{,} , leaf, text width=50.5em, fill=white
                        ]
                    ]
                    [
                        \textbf{Cache Quantization~(\S\ref{sec3_6})}
                        [
                            \quad KVQuant~\citep{hooper2024kvquant}{,} MiKV~\citep{yang2024no}{,} KIVI~\citep{liu2024kivi}{,} \\
                            \quad SKVQ~\citep{duanmu2024skvq}{,} ZipCache~\citep{he2024zipcache}{,} PQCache~\citep{zhang2024pqcache}{,} \\
                            \quad GEAR~\citep{kang2024gear}{,} QJL~\citep{zandieh2024qjl}{,} AsymKV~\citep{tao2024asymkv} ,leaf, text width=50.5em, fill=white
                        ]
                    ]
			]
		\end{forest}
	}
	\caption{An overview of KV cache optimization of long-context LLMs.}
	\label{fig:kv_opt}
\end{figure*}

Token dropping is a technique that identifies \textit{unimportant} tokens and discards them. However, the critical challenge in these methods lies in determining which tokens are \textit{unimportant}. Generally, token classification strategies can be categorized into two main types: static~\citep{xiaoefficient, han2024lm} and dynamic~\citep{zhang2023h2o, li2024snapkv}.

For static strategies, token importance is considered independent of context, with certain tokens at specific positions consistently receiving more attention from the LLMs, and thus being deemed \textit{important}. For instance, sliding window attention~\citep{jiang2023mistral, bai2023qwen} retains the most recent tokens. Building on this, StreamingLLM~\citep{xiaoefficient} and LMInfinite~\citep{han2024lm} observe that the initial tokens also consistently attract more attention from the LLM. Retaining both the most recent and the initial tokens helps mitigate the degradation of LLM performance as context length increases.

In contrast, dynamic strategies adaptively select \textit{important} tokens based on their context. A commonly employed approach involves determining token importance using attention weights. For instance, H2O~\citep{zhang2023h2o} identifies important tokens through cumulative normalized attention scores while prioritizing the retention of the most recent tokens. Scissorhands~\citep{liu2024scissorhands} identifies pivot tokens via attention weights, ensuring that the memory usage of the KV cache remains within a fixed budget.

Due to the inherent flexibility of dynamic approaches, the majority of subsequent work has built upon and extended these methods. In addition to using attention weight as a measure of importance, researchers have identified variations in attention patterns across different attention heads and developed more refined criteria for determining token importance. For example, FastGen~\citep{ge2023model} classifies attention heads into five types and applies distinct token eviction strategies for each. TOVA~\citep{oren2024transformers} removes tokens with the lowest attention scores for each head independently. SnapKV~\citep{li2024snapkv} selects queries within a local window and votes on the importance of previous tokens for each query and head. RazorAttention~\citep{tang2024razorattention} and DuoAttention~\citep{xiao2024duoattention} categorize attention heads into retrieval and non-retrieval heads, prioritizing the retention of initial and recent tokens for non-retrieval heads. \citet{feng2024ada} and \citet{rehg2024kv} take a step further, introducing different eviction rates for each attention head~\citep{fu2024not}.

Other researchers have considered the variability of attention patterns across layers and made corresponding adjustments. PyramidKV~\citep{cai2024pyramidkv} retains more tokens in lower layers, creating a pyramid-like KV cache structure, while PyramidInfer~\citep{yang2024pyramidinfer} extends this by applying token-dropping strategies in deeper layers. SimLayerKV~\citep{zhang2024simlayerkv} focuses on identifying which layers can adopt the StreamingLLM~\citep{xiaoefficient} paradigm and drops intermediate tokens in the corresponding layers. In recent work, SCOPE~\citep{wu2024scope} optimizes KV cache usage separately for the pre-filling and decoding stages.

Beyond attention weights as a measure of token importance, researchers have also explored alternative metrics that may better capture this concept. DCP~\citep{anagnostidis2024dynamic} fine-tunes a low-dimensional QK mapping to determine which tokens to drop. Similarly, Locret~\citep{huang2024locret} fine-tunes a new retention head to prioritize token retention. SirLLM~\citep{yao2024sirllm} utilizes token entropy to decide whether to discard a token. \citet{devoto2024simple} employs the L2 norm of keys to assess token importance. RoCo~\citep{ren2024efficacy} uses the standard deviation of attention scores as a metric for importance. VPM~\citep{guo2024attention} considers not only attention weights but also the values themselves. InfiniPot~\citep{guo2024attention} evaluates token importance based on a combination of future confidence and overlap with past information.

\subsection{Token Merging}\label{sec3_2}

The methods discussed here focus on preserving the information of discarded tokens as much as possible through token merging, which can be seen as an extension of the token-dropping strategies mentioned earlier.

Sentinel Tokens~\citep{ren2023context} introduces sentinel tokens to compress contextual information within segments. Similarly, approaches like Activation Beacon~\citep{zhang2024long} and AnchorLLM~\citep{pang2024anchor} adopt analogous strategies, introducing special tokens to guide LLMs in learning how to effectively compress the KV cache during training, thereby achieving impressive performance. \citet{dong2024get} uses kernel functions to compress preceding contextual information. DMC~\citep{nawrot2024dynamic} fine-tunes decision and weight variables to determine when to expand the KV cache or aggregate weights into the final set of KV caches. \citet{wang2024model} observes the similarity between adjacent keys and employs Gaussian kernel functions to merge neighboring tokens.



\subsection{Layer-wise Sharing}\label{sec3_3}
For optimizations targeting the layer dimension, some approaches involve pre-training LLMs from scratch, while others focus on fine-tuning pre-trained models.

Sharing the KV cache across multiple layers is a common strategy for methods that modify the model architecture during the pre-training stage. YOCO~\citep{sun2024you} divides the decoder into self-decoder and cross-decoder layers. KV cache is generated only in the output layer of the self-decoder, while cross-decoder layers reuse the output from the final self-decoder layer, thereby eliminating the need for additional KV caches. Similarly, GoldFinch~\citep{goldstein2024goldfinch} adopts a related strategy, where the last one-third of the layers utilize a small, compressed global KV cache generated by preceding layers. CEPE~\citep{yen2024long} stores the full KV cache for the main input across all layers, while for additional context, each layer shares a small encoder output cache to perform cross-attention.

For fine-tuning existing LLMs, researchers often adopt straightforward inter-layer cache-sharing strategies. CLA~\citep{brandon2024reducing} uses fine-tuning to enable multiple layers to share the KV cache of a single layer. Additionally, methods such as MiniCache~\citep{liu2024minicache}, LCKV~\citep{wu2024layer}, KVSharer~\citep{yang2024kvsharer}, and SwiftKV~\citep{qiao2024swiftkv} adaptively select inter-layer cache sharing strategies. MLKV~\citep{zuhri2024mlkv} combines layer-wise KV sharing with MQA, integrating adjacent layer sharing with techniques that replace deep-layer KV with shallow-layer KV. CLLA~\citep{yang2024lossless} extends MLA and CLA by incorporating quantization into the shared caching mechanism. In contrast, CEPE~\citep{yen2024long} employs a distinct strategy, storing a single-layer KV cache for all layers by encoding the KV cache with the representation generated by an encoder and integrating it with cross-attention.

Beyond KV cache sharing, researchers have also explored alternative strategies. Shared Attention~\citep{liao2024beyond} directly shares attention weights across different layers to optimize performance along the layer dimension.

\subsection{Head-wise Sharing}\label{sec3_4}

Similar to layer dimension optimizations, reducing the number of heads significantly impacts the representational capacity of LLMs. To preserve performance, head dimension optimizations typically rely on sharing strategies. For instance, GQA~\citep{ainslie2023gqa} and MQA~\citep{shazeer2019fast} reduce memory usage by sharing the KV cache across queries from different heads, a technique now widely adopted in various model architectures. Additionally, fine-tuning existing models can further optimize the size of the head dimension. For example, SHA~\citep{cao2024head} computes the cosine similarity of head weight matrices and groups similar heads to share a single KV cache. DHA~\citep{chen2024dha} employs a centroid alignment method to compute head similarity, linearly fusing the KV caches of similar heads, effectively compressing MHA into GQA.

Beyond KV cache sharing, low-rank compression is frequently used to optimize the head dimension. MLA~\citep{liu2024deepseek} replaces the full KV cache with low-dimensional latent vectors, recovering the KV through a projection matrix and injecting positional information via decoupled RoPE. ECH~\citep{yu2024effectively} applies SVD-based low-rank decomposition to grouped head weight matrices, achieving a KV compression effect similar to GQA, but distinct in its non-averaging fusion. Neurocache~\citep{safaya2024neurocache} applies low-rank compression to head matrices and uses the most similar caches in attention computation.

\subsection{Feature Compression}\label{sec3_5}


Optimization methods targeting feature dimensions primarily focus on low-rank compression, which corresponds to the size per attention head. Palu~\citep{chang2024palu} introduces a medium-grained grouped head low-rank decomposition (G-LRD) method, striking a balance between accuracy and reconstruction efficiency. Eigen Attention~\citep{saxena2024eigen} utilizes a small calibration dataset to select the most significant directions based on SVD. MatryoshkaKV~\citep{lin2024matryoshkakv} addressed the limitations of PCA by fine-tuning the orthogonal projection matrix to align the model outputs as closely as possible with the original outputs. Additionally, it employed a Matryoshka hierarchical strategy to achieve improved compression without sacrificing performance. LoRC~\citep{zhang2024lorc} similarly leveraged SVD, adjusting cumulative condition numbers layer by layer to evaluate and modify compression ratios from deep to shallow layers, effectively preventing error accumulation that could degrade overall performance. In contrast, LPA~\citep{lv2024scalable} focused on incorporating low-rank projection attention structures during pretraining, thereby improving performance on downstream tasks. ThinK~\citep{xu2024think} introduces a dimension-pruning approach for feature compression, evaluating the interaction strength between KV pairs to retain the most significant dimensions.


\subsection{Cache Quantization}\label{sec3_6}
Quantization is one of the most widely used techniques for KV cache compression, commonly adopted in practice for its speed and efficiency~\citep{bai2023qwen, glm2024chatglm}. This optimization focuses on adjusting the size of the KV cache data type, which directly influences the storage size per unit.

Some works adapt traditional quantization methods to the specific characteristics of the KV cache. For example, KVQuant~\citep{hooper2024kvquant} determines quantization parameters through offline data analysis, ensuring that critical information is preserved during the process. In contrast, KIVI~\citep{liu2024kivi} exploits the differing characteristics of keys and values in the model, performing channel-wise quantization for key caches and token-wise quantization for value caches. MiKV~\citep{yang2024no} combines eviction strategies by storing tokens scheduled for eviction at a lower precision. SKVQ~\citep{duanmu2024skvq} rearranges key-value pairs to group outliers together, then trims boundary values within these groups to minimize quantization errors. ZipCache~\citep{he2024zipcache} improves the compression ratio by normalizing attention scores within a channel-separable quantization framework. PQCache~\citep{zhang2024pqcache} integrates embedding retrieval techniques by decomposing the original vector space into Cartesian products of several lower-dimensional vector spaces, which are quantized separately.

Other approaches explore more advanced possibilities in quantization methods. For instance, GEAR~\citep{kang2024gear} further reduces errors compared to full-precision computations by using low-rank and sparse matrices to fit residuals on top of traditional quantization results. QJL~\citep{zandieh2024qjl} introduces a novel KV cache quantization technique optimized specifically for CUDA kernels, enhancing the quantization process's efficiency and making it more suitable for large-scale parallel computing environments. AsymKV~\citep{tao2024asymkv} proposes an asymmetric quantization strategy that enables KV cache operation with extremely low 1-bit precision.

\section{Memory Management}\label{sec4}

While KV cache optimization strives for a longer context practically, essentially, it is a balance between efficiency and performance. Cache optimization does not try to break the ceiling of LLM capabilities, since it does not change the organizing form of contextual information~\citep{fu2024challenges,luohekeep}. Long-context LLMs based on vanilla KV cache mechanism still face limitations including read-only access and the requirement to read all information at once, making them unsuitable for more complex scenarios~\citep{dai2019transformer, bulatov2022recurrent}. This has led to incorporating \textbf{\textit{memory management}} into LLMs, with the KV cache being regarded as a specific memory instance.

Memory management in LLMs can be categorized from two perspectives. One is \textbf{\textit{cache-based memory}} (\S\ref{sec4a}), storing intermediate results that encode contextual information, such as KV cache, or \textbf{\textit{text-based memory}} (\S\ref{sec4b}), storing text directly, which is more convenient and flexible, as it allows the use of external textual data sources. The other is \textbf{\textit{read-only}} or \textbf{\textit{writable}}, based on whether the memory is modifiable during storage. These two aspects divide the memory management methods into four quadrants as shown in Figure~\ref{fig:memory}.


\begin{figure}
    \centering
\begin{tikzpicture}[
    font=\small,
    align=center,
    node distance=1cm and 1.5cm,
    quadrant/.style={draw, minimum width=6cm, minimum height=1.6cm, fill=green!10, anchor=north, inner sep=0pt, text width=5.8cm}, 
]

\node[quadrant] (readonly-cache) at (0, 1.6) {
\begin{minipage}[t][3.2cm][t]{5.8cm}
\centering
\vspace{0.5cm}
\textbf{\S \ref{sec4_1}}\\[1ex]
MemTrans~\citep{wu2022memorizing} \\ AutoCompressor\\~\citep{chevalier2023adapting} \\ ICAE~\citep{ge2023context} \\ PromptCache~\citep{gim2024prompt}
\end{minipage}};

\node[quadrant] (readonly-text) at (6,1.6) {
\begin{minipage}[t][3.2cm][t]{5.8cm}
\centering
\vspace{0.5cm}
\textbf{\S\ref{sec4_3}}\\[1ex]
MemWalker~\citep{chen2023walking} \\ LongRAG~\citep{zhao2024longrag} \\ Self-Route~\citep{li2024retrieval} \\ RAG2.0~\citep{rag2_contextual_ai_2024}
\end{minipage}};

\node[quadrant] (writable-cache) at (0, -1.6) {
\begin{minipage}[t][3.2cm][t]{5.8cm}
\centering
\vspace{0.5cm}
\textbf{\S\ref{sec4_2}}\\[1ex]
Transformer-XL~\citep{dai2019transformer} \\ RMT~\citep{bulatov2022recurrent} \\ MemoryLLM~\citep{wang2024memoryllm} \\ CAMELoT~\citep{he2024camelot} \\ Memory\(^3\)~\citep{yang2024memory3}
\end{minipage}};

\node[quadrant] (writable-text) at (6, -1.6) {
\begin{minipage}[t][3.2cm][t]{5.8cm}
\centering
\vspace{0.5cm}
\textbf{\S\ref{sec4_4}}\\[1ex]
MemGPT~\citep{packer2023memgpt} \\ LongLLMLingua~\citep{jiang2023longllmlingua} \\
RecurrentGPT~\citep{zhou2023recurrentgpt}\\ 
MemoryBank~\citep{zhong2024memorybank}
\end{minipage}};

\node[above=0.25cm of readonly-cache, draw=none, fill=none] {\textbf{Cache-Based Memory}};
\node[above=0.25cm of readonly-text, draw=none, fill=none] {\textbf{Text-Based Memory}};
\node[rotate=90, left=0.5cm of readonly-cache, draw=none, fill=none, anchor=center] {\textbf{Read-Only}};
\node[rotate=90, left=0.5cm of writable-cache, draw=none, fill=none, anchor=center] {\textbf{Writable}};

\end{tikzpicture}

    \caption{An overview of memory management of long-context LLMs.}
    \label{fig:memory}
\end{figure}

\subsection{Cache-Based Memory}\label{sec4a}

In this subsection, memory primarily refers to intermediate computational outputs, including hidden states, KV cache, and compressed textual representations that are irrecoverable.
\subsubsection{Read-Only}\label{sec4_1}

The most intuitive improvement of read-only memory over the KV cache is its more flexible access method, avoiding reading all KV cache at once. MemTrans~\citep{wu2022memorizing} stores the KV cache of pre-training in external memory to provide more relevant information during inference. MemLong~\citep{liu2024memlong} extends this concept to a long context by storing the KV cache of context chunks and retrieving KV pairs based on relevance to guide inference.

Another approach to applying memory to long contexts is to compress the context, ensuring that the LLMs can handle longer sequences. AutoCompressor~\citep{chevalier2023adapting} iteratively processes the context by encoding each segment into a fixed-dimension summary vector and concatenating it with the next part. Later works, such as LLoCO~\citep{tan2024lloco} and E2LLM~\citep{liao2024e2llm}, extend this method with advancements in offline learning and parallel compression, respectively. ICAE~\citep{ge2023context} compresses information by fine-tuning the encoder to encode the entire context into a small number of memory tokens. UIO-LLMs~\citep{li2024uio} further conceptualizes memory-enhanced LLMs as fully connected RNNs, optimized through backpropagation.

\label{prefix_sharing}Additionally, some inference acceleration works have also used memory. PagedAttention~\citep{kwon2023efficient} accelerates inference by reusing the same prefix of KV cache in a single request. Prompt Cache~\citep{gim2024prompt} and SGLang~\citep{zheng2024sglang} speed up inference through structured organization of prompts to enhance performance.

\subsubsection{Writable}\label{sec4_2}

In contrast to read-only memory, writable memory allows dynamic adjustments to stored memories. Transformer-XL~\citep{dai2019transformer}, for example, reuses the hidden states of previous segments to capture long-term dependencies. RMT~\citep{bulatov2022recurrent} improves upon this by introducing special memory tokens to store contextual information, with cross-segment gradient backpropagation to update the memory. \citet{bulatov2023scaling} extends the context length to 1M tokens using RMT. UniMem~\citep{fang2024unimem} further synthesizes previous methods, while MemoryLLM~\citep{wang2024memoryllm} and CAMELoT~\citep{he2024camelot} optimize memory management through more flexible or non-training-based approaches.

As researchers focus on using memory to store contextual or long-term information, Memory$^3$~\citep{yang2024memory3} was the first to introduce knowledge to LLMs and decompose knowledge into abstract knowledge and specific knowledge, formalizing the idea that the LLMs can store only abstract knowledge, while all specific knowledge is stored externally. This external memory is accessed during inference by periodic concatenation of relevant memories, achieving state-of-the-art performance. Titans~\citep{behrouz2024titans} integrated memory with test-time training and further explored the diverse applications of the memory module, thereby pointing out new directions for subsequent research.

\subsection{Text-Based Memory}\label{sec4b}

While cache-based memory has proven effective, it is relatively complex and lacks sufficient interpretability, particularly due to its non-textual nature. Thus, some researchers have turned to text-based memory to enhance LLMs' performance.

\subsubsection{Read-Only}\label{sec4_3}

A common application of text-based memory is the presence of ground truth in text, where providing this text to the LLMs during generation can improve performance. Retrieve Augmented Generation(RAG, \citep{lewis2020retrieval}) utilizes this idea by retrieving external information using a retriever and appending it to the prompt during generation, paving the way for subsequent developments. This idea has been expanded to address long-context problems by retrieving relevant context segments~\citep{chen2023walking}, improving queries~\citep{fei2024retrieval}, combining query and context~\citep{zhao2024longrag}, and improving retrieval methods~\citep{luo2024bge, soh2024you, jiang2024longrag}, effectively addressing long-context challenges.

While RAG-related research has flourished, some studies have questioned the necessity of using RAG. \citet{li2024retrieval} conducted experiments revealing that performance with long-context LLMs outperforms RAG, suggesting an LLM-driven decision of whether to reuse long-context responses after initially employing RAG. The question of whether long-context or RAG is better remains a topic of ongoing discussion, which will be addressed later in \textbf{\nameref{q4_rag}} in Section \ref{sec12}. Some argue that RAG is more suitable for resource-constrained scenarios compared to long-context, and we will also present our perspectives on this matter in Section\ref{sec7}. \citet{rag2_contextual_ai_2024} integrates various RAG components and conducts end-to-end training, achieving state-of-the-art results.

\subsubsection{Writable}\label{sec4_4}

Writable text-based memory can be used to store and update historical information. MemoryBank~\citep{zhong2024memorybank} stores user history and profiles, achieving better user preference. Inspired by LSTM, RecurrentGPT~\citep{zhou2023recurrentgpt} summarizes preceding content during each step, facilitating ultra-long text generation. MemGPT~\citep{packer2023memgpt} designs a multi-layered memory architecture, structuring prompts based on operating system memory access principles. EM$^2$~\citep{yin2024explicit} was the first to recognize that the direction of memory updates is not always optimal, introducing the EM algorithm~\citep{dempster1977maximum} and treating memory as latent variables to estimate the correct update direction.

Some researchers have also used memory to compress long contexts. One approach, which we refer to as text-level compression, involves compressing the context into several complete texts. Researchers have explored content-based compression~\citep{fei2023extending}, relevance-based compression~\citep{yoon2024compact}, and attention-weighted compression~\citep{choi2024reading}, achieving promising results. Another approach, token-level compression, compresses context into tokens that may not form complete sentences. LongLLMLingua~\citep{jiang2023longllmlingua} and Perception Compressor~\citep{tang2024perception} select the most relevant content based on correlations, retaining only the most important tokens to achieve token-level compression. Selection-p~\citep{chung2024selection} retains a proportion of the original context tokens and trains the LLMs to generate responses using this limited set of tokens, resulting in significant improvements.

\section{Architecture Innovation}\label{sec5}

Although KV cache optimization (Section \ref{sec3}) and memory management (Section \ref{sec4}) have improved the long-context capability of Transformer-based LLMs. The inherent shortage of Transformer in computation and memory efficiency still drives researchers to explore innovations in the attention mechanism itself, resulting in more radical architecture innovations~\citep{jiang2024minference,ye2024differential,peng2023rwkv,gu2023mamba}. In this section, we will demonstrate those architectural innovations concerning long-context efficiency or performance from three perspectives as shown in Figure~\ref{fig:arch_inno}.
\begin{itemize}
    \item In \S\ref{sec5_1}, we will analyze \textbf{\textit{efficient attention}}, the attention variant towards better computational efficiency or long-context performance. It can be further divided into two branches. One is \textbf{\textit{attention approximation}}, an efficient approximation for standard attention, such as MInference~\citep{jiang2024minference}, RetrievalAttention~\citep{liu2024retrievalattention} and other sparse attention methods~\citep{yang2024post,zhu2024sampleattention}, while the other is \textbf{\textit{attention alternative}}, which tries a novel attention mechanism like DIFF-Transformer~\citep{ye2024differential}, Lightning Attention~\citep{qin2024various,qin2024lightning} and other linear attentions~\citep{katharopoulos2020transformers}.
    \item As a cache-free architecture, discussion on LSTM~\citep{schmidhuber1997long} is revived for the pursuit of long context. In \S\ref{sec5_2}, we will analyze researches on LSTM in the LLM era, including the \textbf{\textit{module-level Improvements}} like xLSTM~\citep{beck2024xlstm} and HGRN series~\citep{qin2024hierarchically,qin2024hgrn2} and the \textbf{\textit{model-level advancements}}, namely RWKV series~\citep{peng2023rwkv,peng2024eagle,choe2024rwkv}.
    \item In \S\ref{sec5_3}, we will show the developing path of the widely-discussed Mamba series~\citep{gu2023mamba,daotransformers,wang2024mamba}, from the \textbf{\textit{theoretical basis}} such as HiPPO~\citep{gu2020hippo} and S4~\citep{gu2021efficiently} to its improvements~\citep{ben2024decimamba,yuan2024remamba}, then to the \textbf{\textit{hybrid architectures}}~\citep{dong2024hymba,akhauri2024attamba}, including Jamba series~\citep{team2024jamba,lieber2024jamba}
\end{itemize}


\subsection{Efficient Attention}\label{sec5_1}

\tikzstyle{leaf}=[my-box,
	text=black, align=left,font=\normalsize,
	inner xsep=2pt,
	inner ysep=4pt,
	line width=0.8pt,
        minimum height=1cm,
]

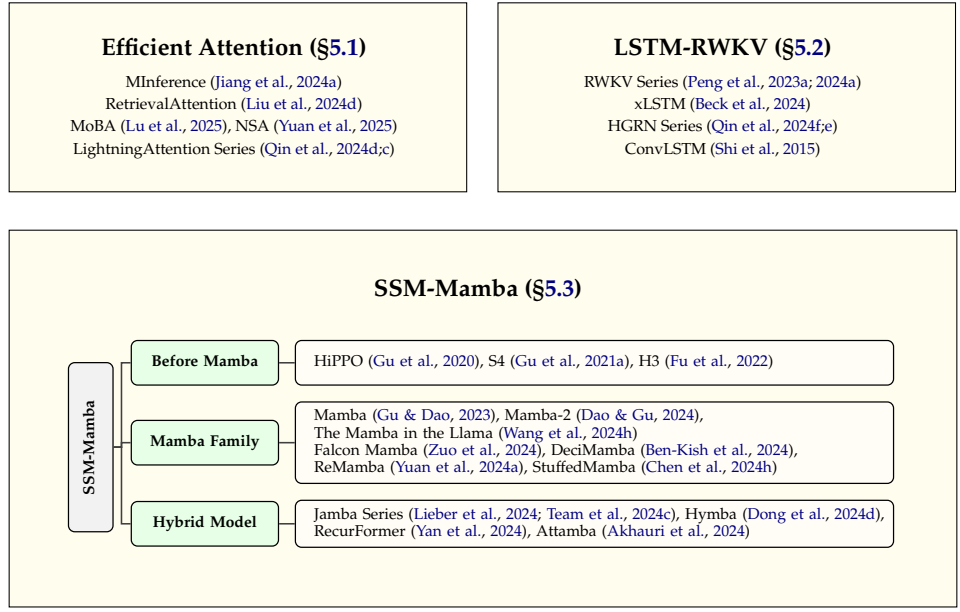
\begin{figure}
    \centering
\begin{tikzpicture}[
    font=\tiny,
    align=center,
    node distance=1cm and 1.5cm,
    quadrant/.style={draw, minimum width=3cm, minimum height=2.5cm, fill=yellow!8},
    central/.style={circle, draw=orange!40, fill=orange!10, minimum size=2.5cm},
    axis/.style={thick}
]

\node[quadrant, inner sep=0pt, anchor=west] (efficient-attention-mechanisms) at (-3, 2.25) {
    \begin{minipage}[t][2.2cm][t]{6cm} 
        \vspace{1.5em} 
        \centering
        {\small\textbf{Efficient Attention~(\S\ref{sec5_1})}} 
        \vspace{0.5em} 

        \renewcommand{\arraystretch}{1.2} 
        {
            \begin{tabular}{c}
                MInference~\citep{jiang2024minference} \\
                RetrievalAttention~\citep{liu2024retrievalattention} \\
                MoBA~\citep{lu2025mobamixtureblockattention}{,} NSA~\citep{yuan2025native} \\
                LightningAttention Series~\citep{qin2024various,qin2024lightning}
            \end{tabular}
        }
    \end{minipage}
};

\node[quadrant, inner sep=0pt] (lstm-rwkv) at (6.5, 2.25) {
    \begin{minipage}[t][2.2cm][t]{6cm} 
        \vspace{1.5em} 
        \centering
        {\small\textbf{LSTM-RWKV~(\S\ref{sec5_2})}} 
        \vspace{0.5em} 

        \renewcommand{\arraystretch}{1.2} 
        {
            \begin{tabular}{c}
                RWKV Series~\citep{peng2023rwkv, peng2024eagle} \\
                xLSTM~\citep{beck2024xlstm} \\ 
                HGRN Series~\citep{qin2024hierarchically,qin2024hgrn2}\\
                ConvLSTM~\citep{shi2015convolutional}
            \end{tabular}
        }
    \end{minipage}
};

\node[draw, fill=yellow!8, align=center, anchor=west] (SSM-Mamba) at (-3, -2) {
    \begin{minipage}[t][4.75cm][t]{12.25cm} 
        \vspace{2.5em} 
        \centering
        \small{\textbf{SSM-Mamba~(\S\ref{sec5_3})}} 
        \vspace{1.5em} 
        
        \resizebox{0.9\textwidth}{!}{ 
            \begin{forest}
                forked edges,
                for tree={
                    grow=east,
                    reversed=true,
                    anchor=base west,
                    parent anchor=east,
                    child anchor=west,
                    base=center,
                    font=\normalsize,
                    rectangle,
                    draw=black, 
                    rounded corners,
                    align=left,
                    text centered,
                    minimum width=4em,
                    minimum height=1cm,
                    edge+={darkgray, line width=1pt},
                    s sep=10pt,
                    inner xsep=2pt,
                    inner ysep=3pt,
                    line width=0.8pt,
                    ver/.style={rotate=90, child anchor=north, parent anchor=south, anchor=center, minimum width=12em, fill=gray!10},
                },
                where level=1{text width=10em, fill=green!10}{},
                where level=2{text width=42em, fill=white}{},
                where level=3{text width=40em, fill=white}{},
                where level=4{text width=18em,}{},
                where level=5{text width=18em,}{},
                [
                    \textbf{SSM-Mamba}, ver
                    [
                        \textbf{Before Mamba}
                        [
                            \quad HiPPO~\citep{gu2020hippo}{,} S4~\citep{gu2021efficiently}{,} H3~\citep{fu2022hungry}, leaf
                        ]
                    ]
                    [
                        \textbf{Mamba Family}
                        [
                            \quad Mamba~\citep{gu2023mamba}{,} Mamba-2~\citep{daotransformers}{,} \\ \quad The Mamba in the Llama~\citep{wang2024mamba} \\ \quad Falcon Mamba~\citep{zuo2024falcon}{,} DeciMamba~\citep{ben2024decimamba}{,} \\ \quad ReMamba~\citep{yuan2024remamba}{,} StuffedMamba~\citep{chen2024stuffed} , leaf
                        ]
                    ]
                    [
                        \textbf{Hybrid Model}
                        [
                            \quad Jamba Series~\citep{lieber2024jamba,team2024jamba}{,} Hymba~\citep{dong2024hymba}{,} \\ \quad RecurFormer~\citep{yan2024recurformer}{,} Attamba~\citep{akhauri2024attamba}, leaf
                        ]
                    ]
                ]
            \end{forest}
        }
    \end{minipage}
};

\end{tikzpicture}

    \caption{An overview of architecture innovation in long-context LLM.}
    \label{fig:arch_inno}
\end{figure}

\subsubsection{Attention Approximation}

Attention approximation is a hot research topic in long-context LLMs. Most attention approximation approaches are achieved with dynamic sparse attention through retrieval-based~\citep{ribar2023sparq,liu2024retrievalattention} or attention pattern observation~\citep{jiang2024minference}. For example, SparQ Attention~\citep{ribar2023sparq} optimizes the attention mechanism through approximate attention computation based on KV cache extraction and interpolation compensation. Similarly, Loki~\citep{singhania2024loki} ranks and selects tokens in the KV-cache based on attention scores computed in low dimensional space. Moreover, SampleAttention~\citep{zhu2024sampleattention} proposes a two-stage sampling filtering mechanism, identifying important attention patterns through query sampling, then combining selected KV cache with sliding windows. DoubleSparse~\citep{yang2024post} uses important feature channels to identify key tokens, thereby reducing access to the KV cache. RetrievalAttention~\citep{liu2024retrievalattention} identifies the inconsistency between query and key vector distribution and resolves it through approximate nearest neighbor search. MagicPIG~\citep{chen2024magicpig} utilizing locality-sensitive hashing (LSH) sampling to estimate attention layer outputs. SqueezedAttention~\citep{hooper2024squeezed} optimizes attention computation by identifying the most important keys through semantic clustering and hierarchical lookup. Recently, MoBA~\citep{lu2025mobamixtureblockattention} combines the concepts of Mixture of Experts (MoE) and sparse attention, allowing each query to selectively focus on a part of the KV pairs, effectively reducing computational costs while maintaining performance.

Other attention approximation approaches are achieved with dynamic sparse attention based on further observation of attention pattern~\citep{jiang2024minference}. For example,
MInference~\citep{jiang2024minference} proposes dynamic sparse attention from the perspective of sparse patterns. StarAttention~\citep{acharya2024star} proposes dividing the input into chunks distributed across different hosts for local attention computation, followed by aggregating global attention results through designated query hosts. And some works improve attention computation efficiency by using full attention and sparse attention in different layers or different heads~\citep{beltagy2020longformer,li2019big,ainslie2020etc}. Additionally, Fourier Transformer~\citep{he2023fourier} removes redundant contextual information from hidden states by discrete cosine transform (DCT) to reduce computational complexity.

\subsubsection{Attention Alternative}
In this part, we will present methods that modify the fundamental mathematics of dot-product attention as attention alternative mechanisms, which require LLM pre-training from scratch but offer theoretical guarantees of improved efficiency~\citep{choromanski2020rethinking}. A representative work is linear attention~\citep{katharopoulos2020transformers}, which reformulates dot-product attention using kernel functions to achieve linear complexity. 
In the LLM era, several recent studies propose novel approaches. SLAB~\citep{guo2024slab} optimizes attention computation efficiency through simplifies linear attention and progressive LayerNorm replacements. Lightning Attention~\citep{qin2024various} achieves efficient computation by blocking and using linear attention between blocks. Its improved version, Lightning Attention-2~\citep{qin2024lightning}, achieves the ability to process infinite-length contexts by introducing an exponential decay mechanism in the KV cache. \citet{minimax2025minimax01scalingfoundationmodels} further takes the advantages of both lightning attention and softmax attention to enhance retrieval performance by substituting lightning attention with softmax attention at intervals of every eight layers. Gated Slot Attention~\citep{zhanggated} enhances ABC~\citep{peng2022abc} by incorporating a gating mechanism, essentially comprising a two-layer GLA~\citep{yanggated} linked via softmax to achieve more efficient memory utilization. What's more, DIFF Transformer~\citep{ye2024differential} calculates attention scores as the difference between two separate softmax attention maps. This subtraction eliminates noise and promotes the emergence of sparse attention patterns. DeepSeek recently release NSA~\citep{yuan2025native}, combining compressed, selected and sliding attention. 

Furthermore, a recent study~\citep{yang2024efficient} reveals an important insight: the efficiency of efficient attention, both sparse and linear attention, is task-dependent, with advantages primarily manifesting in tasks exhibiting locality characteristics. This finding opens new perspectives for research in efficient attention mechanisms.

\subsection{LSTM-RWKV}\label{sec5_2}
Despite numerous advances in Transformer's computational efficiency, significant storage limitations persist~\citep{ribar2023sparq,yang2024post}. This leads researchers to explore \textit{cache-free} architectures, with improvements of LSTM~\citep{graves2012long} emerging as a key direction. Compared to the Transformer's quadratic complexity, LSTM's linear inference complexity demonstrates significant advantages in long context scenarios. The improvements encompass both module-level enhancements to the basic LSTM architecture and large-scale innovations exemplified by RWKV~\citep{peng2023rwkv,peng2024eagle}, which shows exceptional performance in complex reasoning tasks like Sudoku\footnote{https://zeeklog.com/rwkv-tong-guo-ji-wan-token-de-cot-jie-jue-ji-hu-100-de-shu-du-wen-ti-cai-yong-29m-can-shu-de-xiao-mo-xing--2/}.

\subsubsection{Module-level Improvements}
For example, xLSTM~\citep{beck2024xlstm} consists of two parts. sLSTM introduces exponential gating, normalization, and stabilization mechanisms while supporting multi-head processing, significantly enhancing LLM's expressiveness while maintaining parallelism. Meanwhile, mLSTM further expands the cell state from vector to matrix form, giving LLMs stronger memory capabilities. Based on the xLSTM architecture, xLSTM-Mixer~\citep{kraus2024xlstm} further introduces normalization and initial linear prediction mechanisms, enhancing LLM's performance by combining original embeddings and reverse embeddings. HGRN~\citep{qin2024hierarchically} emphasizes the importance of forget gates in recursive layers, achieving hierarchical modeling of long-short term dependencies through learnable, layer-increasing lower bound values. Furthermore, HGRN2~\citep{qin2024hgrn2} innovatively introduces an outer product-based state expansion mechanism, expanding the scale of recursive states without increasing parameters, and addresses increased computational complexity through multi-head variants. Additionally, \citet{feng2024were} simplifies LSTM to enable parallel computation, improving LLM's computational efficiency.

Beyond these works, ConvLSTM~\citep{shi2015convolutional} is an important direction for improvement. ConvLSTM demonstrates the viability and advantages of incorporating convolutional structures into LSTM. By implementing convolutional structures in both input-to-state and state-to-state transitions, ConvLSTM successfully extends LSTM to handle spatiotemporal context data. This innovation provides crucial insights for subsequent improvements of LSTM  improvements~\citep{wang2022predrnn,wang2018predrnn++,wang2019memory,lin2020self}.

\subsubsection{Model-level Advancements}
RWKV series represents a new technical approach, striving to combine the advantages of RNN and Transformer. RWKV4~\citep{peng2023rwkv} introduces token shift, similar to convolutional sliding window processing, and processes context information through the fusion of time dimension (time-mixing) and feature dimension (channel-mixing). Its innovative WKV operator achieves training phase parallelization and linear complexity during inference. Subsequently, RWKV's development reaches new heights with RWKV5 (Eagle) and RWKV6 (Finch)~\citep{peng2024eagle}. RWKV5 introduces multi-head mechanisms similar to Transformer's multi-head attention mechanism and optimizes token shift through linear interpolation. In time-mixing, it enhances LLM's expressiveness by introducing new trainable parameters. RWKV6 further innovates with significant improvements in both token shift and time-mixing, particularly incorporating LoRA's implementation approach and allowing each channel to mix token information rather than relying on fixed trainable parameters. These improvements enable the LLM to demonstrate superior performance and higher efficiency in processing long contexts.

\subsection{SSM-Mamba}\label{sec5_3}
State Space Model (SSM) represent an innovative architecture that delivers several key advances~\citep{gu2023mamba}. Its linear computational complexity significantly outperforms the quadratic complexity of Transformers. It eliminates memory requirement for attention matrices through fixed hidden state storage. Most importantly, SSM supports parallel training and linear generation, offering substantial practical advantages.

SSM originates from modern control system theory. It encodes context information by maintaining hidden states and using linear dynamical systems to describe state evolution:$x'(t)=Ax(t)+Bu(t)$,$y(t)=Cx(t)+Du(t)$, where $x(t)$ represents hidden state, $u(t)$ represents input, $y(t)$ represents output, and $A$,$B$,$C$,$D$ are parameter matrices. 

\subsubsection{Pre-Mamba Works}
Although HiPPO~\citep{gu2020hippo} is initially applied to RNNs, it lays crucial theoretical foundations for the development of Mamba. HiPPO utilizes polynomial approximation and specific probability measures (LegS probability measure) to construct a new matrix structure (HiPPO matrix), effectively modeling context data by encoding historical information into polynomial coefficients. Building on this, LSSL~\citep{gu2021combining} further reveals the connection between RNN, CNN and SSM, discovering that SSM could be represented in both recurrent and convolutional forms. More importantly, LSSL first attempts to use HiPPO Matrix to initialize SSM's parameters, achieving significant performance improvements on multiple tasks. Then, the introduction of S4~\citep{gu2021efficiently} marks a major breakthrough in SSM's computational efficiency. This work represents HiPPO matrix in NPLR (Normal Plus Low-Rank) form and reduces SSM's computational overhead from both recurrent and convolutional perspectives through matrix theory derivations. The subsequent S4D~\citep{gu2022parameterization} proposes a simplified version of S4, further improving computational efficiency while maintaining LLM's performance by restricting the state matrix to a completely diagonal form. Later, H3~\citep{fu2022hungry} focuses on addressing SSM's shortcomings in language modeling tasks. Inspired by linear attention mechanisms, H3 represents the update of SSM's hidden state as $Q\odot SSM_{diag}(SSM_{shift}(K))\odot V$, where the two SSM matrices employ the "hungry hippo" mechanism to enhance efficiency. H3's performance in synthetic language modeling tasks matches attention mechanisms. Additionally, H3 introduces FlashConv to extend context length and improve training efficiency. In the above architecture innovations based on recurrent networks, local information interaction such as token shift often appears. Based on this property, we will further the discussion on new architecture in \textbf{\nameref{q5_na}} in Section\ref{sec12}

\subsubsection{Introduction and Improvements of Mamba}
The introduction of Mamba~\citep{gu2023mamba} represents a significant milestone in SSM's development. It introduces a selective mechanism and enables content-aware capabilities. Specifically, when updating parameter matrices, Mamba incorporates projection information of inputs, allowing each token to have independent parameter matrices. Simultaneously, Mamba proposes a hardware-aware parallel recursive algorithm to improve computational efficiency. Mamba-2~\citep{daotransformers} further improve the architecture, elucidating the dual relationship between Mamba and attention mechanisms through detailed theoretical analysis and providing insights for the integrated use of attention mechanisms and Mamba.

However, as research deepened, researchers discover Mamba's limitations in processing long contexts. Several works propose solutions from different angles. DeciMamba~\citep{ben2024decimamba} proposes a token selection mechanism based on $\Delta_t$ by analyzing Mamba's receptive field. ReMamba~\citep{yuan2024remamba}, inspired by KV cache's compression method, uses architecture's characteristic of aggregating information through hidden states to select the most representative representations using importance score mechanisms. StuffedMamba~\citep{chen2024stuffed} reveals the essence of the state collapse phenomenon, proposing multiple mitigation strategies including increasing state decay amount, reducing input information quantity, normalizing states, and simulating sliding window mechanisms.

Furthermore, researchers are exploring other optimization directions. SMR~\citep{qi2024smr} analyzes SSM's sampling stability issue from a control theory perspective, proposing an event-triggered control (ETC) based solution—introducing learnable memory to adjust current states and resolving Mamba's inability to use convolution, enabling efficient parallel computation. Mamba-PTQ~\citep{pierro2024mamba} discovers the outlier channels problem in Mamba's quantization and uses SmoothQuant technology, balancing weight and activation quantization difficulty through transfer factor $\alpha$. Additionally, The Mamba in the Llama~\citep{wang2024mamba} uses the standard attention parameters to initialize Mamba, combining knowledge distillation and multi-step speculative decoding to improve efficiency.

\subsubsection{Hybrid Architectures}
Recently, researchers have explored hybrid architectures that combine SSM and Transformer. Early Jamba~\citep{lieber2024jamba} adopts a relatively direct approach, stacking Transformer, Mamba, and MoE blocks in combination, aiming to balance memory usage, computational throughput, and LLM's performance. RecurFormer~\citep{yan2024recurformer} then proposes a more targeted hybrid solution, with its core idea being to identify and replace attention heads in Transformer that focus on local perception with Mamba blocks. Subsequently, Hymba~\citep{dong2024hymba} proposes a deeper integration approach, adopting parallel Attention heads and SSM heads structure to avoid potential information bottleneck issues that might arise from serial architecture. And it achieves an organic fusion of the two types of heads through learnable parameters. Additionally, Attamba~\citep{akhauri2024attamba} explores a new compression approach that uses SSM blocks to compress multiple tokens into one chunk token for Transformer processing. And it also combines sliding window concepts to preserve the initial state of local tokens, thereby reducing KV cache.

\paragraph{Other New Architectures}
Beyond the aforementioned work, researchers also propose many other \textit{cache-free} architectures, providing new perspectives for improving LLM's ability to process long contexts. Some works are based on Neural ODE~\citep{chen2018neural}, such as Liquid Time-constant Networks~\citep{hasani2021liquid} introducing a dynamic adjustable liquid time constant mechanism and CfC~\citep{hasani2022closed} avoiding the need for numerical solutions by finding approximate closed-form solutions for LTC. Additionally, MixCon~\citep{xu2024mixcon} proposes a hybrid architecture combining Transformer layers, Conba layers, and MoE and introducing mechanisms such as selective state spaces to enhance LLM's performance. MCSD~\citep{yang2024mcsd} captures local and global features through Slope and Decay components respectively, and adopts a dual-branch design to strengthen feature extraction and fusion.

\section{Training Infrastructure}\label{sec6}

\tikzstyle{my-box}=[
	rectangle,
	draw=black, 
	rounded corners,
	text opacity=1,
	minimum height=1.5em,
	minimum width=5em,
	inner sep=2pt,
	align=center,
	fill opacity=.5,
	line width=0.8pt,
]
\tikzstyle{leaf}=[my-box, minimum height=1.5em,
	text=black, align=left,font=\normalsize,
	inner xsep=10pt,
	inner ysep=4pt,
	line width=0.8pt,
]

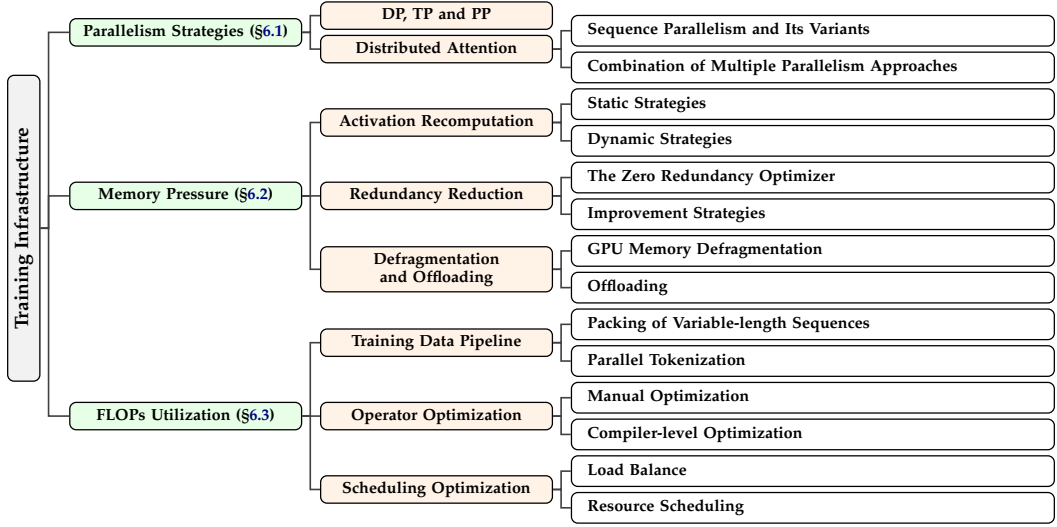
\begin{figure*}[t]
	\centering
	\resizebox{\textwidth}{!}{
		\begin{forest}
			forked edges,
			for tree={
				grow=east,
				reversed=true,
				anchor=base west,
				parent anchor=east,
				child anchor=west,
				base=center,
				font=\large,
				rectangle,
				draw=black, 
				rounded corners,
				align=left,
				text centered,
				minimum width=4em,
				edge+={darkgray, line width=1pt},
				s sep=3pt,
				inner xsep=2pt,
				inner ysep=3pt,
				line width=0.8pt,
				ver/.style={rotate=90, child anchor=north, parent anchor=south, anchor=center, minimum width=19em, fill=gray!10},
			},
			where level=1{text width=14em,font=\normalsize,fill=green!10}{},
			where level=2{text width=14em, align=left, font=\normalsize, fill=orange!10}{},
			where level=3{text width=28em, align=left}{},
			where level=4{text width=18em,font=\normalsize,}{},
			where level=5{text width=18em,font=\normalsize,}{},
			[
				\textbf{Training Infrastructure}, ver
				[
                        \textbf{Parallelism Strategies~(\S\ref{sec6_1})}
                        [
                            \textbf{DP, TP and PP}
                        ]
                        [
                            \textbf{Distributed Attention}
                            [
                                \textbf{Sequence Parallelism and Its Variants}, leaf
                            ]
                            [
                                \textbf{Combination of Multiple Parallelism Approaches}, leaf
                            ]
                        ]
                    ]
                    [
                        \textbf{Memory Pressure~(\S\ref{sec6_2})}
                        [
                            \textbf{Activation Recomputation}
                            [
                                \textbf{Static Strategies}, leaf
                            ]
                            [
                                \textbf{Dynamic Strategies}, leaf
                            ]
                        ]
                        [
                            \textbf{Redundancy Reduction}
                            [
                                \textbf{The Zero Redundancy Optimizer}, leaf
                            ]
                            [
                                \textbf{Improvement Strategies}, leaf
                            ]
                        ]
                        [
                            \textbf{Defragmentation}\\\newline\textbf{and Offloading}
                            [
                                \textbf{GPU Memory Defragmentation}, leaf
                            ]
                            [
                                \textbf{Offloading}, leaf
                            ]
                        ]
                    ]
                    [
                        \textbf{FLOPs Utilization~(\S\ref{sec6_3})}
                        [
                            \textbf{Training Data Pipeline}
                            [
                                \textbf{Packing of Variable-length Sequences}, leaf
                            ]
                            [
                                \textbf{Parallel Tokenization}, leaf
                            ]
                        ]
                        [
                            \textbf{Operator Optimization}
                            [
                                \textbf{Manual Optimization}, leaf
                            ]
                            [
                                \textbf{Compiler-level Optimization}, leaf
                            ]
                        ]
                        [
                            \textbf{Scheduling Optimization}
                            [
                                \textbf{Load Balance}, leaf
                            ]
                            [
                                \textbf{Resource Scheduling}, leaf
                            ]
                        ]
                    ]
			]
		\end{forest}
	}
	\caption{An overview of training infrastructure of long-context LLMs.}
	\label{fig:train_infra}
\end{figure*}

Although architectural innovation has achieved great progress, the mainstream long-context LLMs are still based on Transformer~\citep{dubey2024llama,llama3_3,Deepseek2024DeepSeek-V3} or hybrid architectures~\citep{team2024jamba,minimax2025minimax01scalingfoundationmodels}. Therefore, we need to make long-context training and inference possible while accepting the inherent drawback of the self-attention mechanism. To further the journey of extending context length, we turn our focus to the practical training and inference of long-context LLMs to explore infrastructure improvement. Whether for long-context training discussed in Section~\ref{sec6} or inference infrastructure discussed in Section~\ref{sec7}, the focus of research all involve: computation, storage, and distribution, namely parallelism as shown in Figure~\ref{fig:train_infra} and Figure~\ref{fig:infer_infra}.

For example, for training infrastructure, 
currently, leading LLMs support context lengths exceeding 128k tokens~\citep{meta2024introducing, dubey2024llama, yang2024qwen2technicalreport} and up to 256k tokens during pre-training (e.g., Qwen2.5~\citep{qwen2024qwen25technicalreport}). At such a context length, the distributed parallelism strategies address the basic question of training possibility. Beyond that, handling such long sequences imposes significant memory demands and necessitates enhanced hardware utilization efficiency:

\begin{itemize}[leftmargin=2em]
\item GPU memory overhead scales proportionally with context length through activation values and optimizer states~\citep{guo2024survey, duan2024efficient}. The demand for memory bandwidth intensifies due to larger tensor sizes~\citep{semianalysis2023AICapacity, semianalysis2024trainium2}. The growth in GPU memory capacity and memory bandwidth has consistently fallen behind advances in GPU computational power~\citep{gholami2024ai, semianalysis2023AICapacity}, further exacerbating the aforementioned challenges.
\item Memory-Flops Utilization (MFU) represents the ratio of actual computational use to theoretical hardware performance. Large-scale long-context distributed training introduces considerable computational \& communication overhead~\citep{gu2024loongtrain, sun2024seq1f1befficientsequencelevelpipeline}, reducing  MFU. Accommodating longer contexts typically necessitates smaller batch sizes, thereby decreasing throughput.  
\end{itemize}

We will briefly review mixed-precision training work~\citep{narang2017mixed, kalamkar2019study, sun2019hybrid, peng2023fp8, dubey2024llama, Deepseek2024DeepSeek-V3} at the end of this section, as it reduces GPU memory requirements and increases MFU, however expanding the supported context length of current training systems only indirectly.

\subsection{Distributed Parallelism Strategies}\label{sec6_1}

Training modern AI models with extensive context windows has become increasingly complex, pushing beyond what single GPUs can handle~\citep{semianalysis2023BlackwellInferenceTraining}. This challenge has led to the development of sophisticated distributed training approaches, particularly when dealing with long context.

\subsubsection{Data, Tensor, and Pipeline Parallelism}

The foundational and most widely adopted approaches in the distributed parallelism are~\citep{semianalysis2023BlackwellInferenceTraining}: Data Parallelism (DP), which distributes input data across multiple GPUs~\citep{li2020pytorch, zhao2023pytorch, Zhang2024SimpleFSDPSF, sun2024co2}; Tensor Parallelism (TP), which splits model parameters matrices across devices; and Pipeline Parallelism (PP), which distributes model layers across GPUs. While each approach offers distinct advantages, they also present unique challenges. TP, for instance, effectively manages memory constraints but typically requires high-bandwidth communication between devices~\citep{dong2024lowbitcommunicationtensorparallel}. Similarly, PP often encounters efficiency losses due to pipeline bubble, and efforts are being made to eliminate this problem~\citep{Li2021TeraPipeTP, Qi2024ZeroB, arfeen2024pipefillusinggpusbubbles}. \citet{sun2024seq1f1befficientsequencelevelpipeline} schedules the pipeline of training LLMs at the sequence level on sequences up to 64k, reducing pipeline bubbles and memory footprint.

\subsubsection{Distributed Attention}   

Sequence Parallelism (SP), specifically designed for long-context training, partitions input and output tensors along the sequence dimension at the Transformer layer level. It facilitates distributed processing of attention computations~\citep{li2021sequence} and other operations ~\citep{Shoeybi2019MegatronLMTM}. \citet{Bian2021ColossalAIAU} introduced a sequence dimension partitioning and parallelization scheme. Ring Attention~\citep{li2021sequence} then employs block-wise attention computation combined with a ring communication pattern to partition QKV tensors along the sequence dimension, distributing computation across devices. Ring Attention can be integrated with FlashAttention~\citep{dao2022flashattention, daoflashattention}, preserving IO-awareness and memory efficiency. Ring attention with block-wise transformers~\citep{liu2023ring} further enhances the overlap between communication and computation, enabling the training of sequences exceeding 100 million tokens. Varlen Ring Attention~\citep{minimax2025minimax01scalingfoundationmodels} avoids the excessive padding and subsequent computational waste associated with traditional methods by applying the ring attention algorithm directly to the entire sequence after data-packing. To address Ring Attention's load imbalance in causal attention mask scenarios, several optimization~\citep{Brandon2023StripedAF, li2024distflashattn, fang2024uspunifiedsequenceparallelism, gu2024loongtrain, minimax2025minimax01scalingfoundationmodels} solutions have emerged. Alternatively, Megatron-LM~\citep{Shoeybi2019MegatronLMTM} achieves load balancing through input token reordering.

Ulysses-Attention~\citep{jacobs2023deepspeed} introduces head-parallel stratification atop sequence dimension partitioning, enabling parallel attention head processing across GPU devices. The 2D-Attention mechanism~\citep{gu2024loongtrain} resolves head-parallel strategy scalability limitations while addressing efficiency constraints present in previous context-parallel approaches such as \citet{Brandon2023StripedAF} and \citet{li2024distflashattn}. \citet{sun2024linearattentionsequenceparallelism} tailored to linear attention-based language models, scales sequence length up to 4096k.

In practical implementations, ultra-long context(eg. longer than 256k)~\citep{qwen2024qwen25technicalreport} training typically requires a strategic combination of multiple parallelism approaches. For example, common configurations integrate tensor and sequence parallelism within individual nodes while implementing data parallelism across machines. This hybrid parallelism methodology~\citep{Shoeybi2019MegatronLMTM, Narayanan2021EfficientLL, jacobs2023deepspeed, chen2024internevo, singh20234d, fujii2024acceleratinglargelanguagemodel, dubey2024llama} enables effective scaling to larger computing clusters, substantially enhancing pre-training and fine-tuning efficiency.  In particular, Varlen Ring Attention~\citep{minimax2025minimax01scalingfoundationmodels} can avoid excessive padding by applying the ring attention algorithm directly to the entire sequence after (varlen-like) data-packing. This flexable integration improves computational efficiency in ultra-long context scenarios up to 1024k tokens. However, existing automatic parallelism tools require further optimization for the unique computation and communication patterns characteristic of ultra-long context scenarios.

\subsection{Alleviating GPU Memory Pressure}\label{sec6_2}

GPU memory constraints have emerged as a critical bottleneck in model training as context windows expand. This pressure stems primarily from~\citep{gholami2024ai, guo2024survey, duan2024efficient}: 

\begin{itemize}[leftmargin=2em]
\item Model parameters themselves
\item activation values and optimizer states
\item inter-device communications
\item temporary space allocations and GPU memory fragmentation
\end{itemize}

While not specifically designed for long-context processing, current solutions offer valuable insights for training such models. We will provide a concise overview.

\subsubsection{Activation Recomputation}

GPU memory usage scales with sequence length. Activation recomputation~\citep{chen2016training, chen2024optimizing} trades computational power for memory space, addressing memory constraints while potentially improving the compute-to-memory ratio and helping resolve memory bottlenecks.

Selective checkpointing~\citep{korthikanti2023reducing, torch2024selective} methods preserve outputs from critical layers, such as attention modules~\citep{li2024distflashattn}, while recomputing other intermediate results as needed. Selective-Checkpoint++~\citep{gu2024loongtrain} significantly reduces memory usage while maintaining performance by adding attention modules to a whitelist and preserving their softmax outputs.

In contrast to static strategies, dynamic recomputation approaches determine which activation values to discard and recompute at runtime. \citet{Kirisame2020DynamicTR} and \citet{Hu2022MegTaiChiDT} employs heuristic methods for runtime tensor eviction and recomputation, while \citet{zhao2024efficientlytraining7bllm} uses a token-wise activation recomputation and swapping mechanism with linear programming to optimize, like, activation value recomputation.

\subsubsection{Redundancy Reduction}

The Zero Redundancy Optimizer (ZeRO) introduces a progressive sharding scheme to minimize memory redundancy~\citep{rajbhandari2020zero}. ZeRO-1 distributes optimizer states across GPUs, ZeRO-2 extends this to gradients, and ZeRO-3 further shards model parameters, effectively dividing the total memory overhead by the parallel dimension. While this comprehensive sharding minimizes redundancy, it increases communication overhead. Numerous other works~\citep{wu2023rethinking, luo2023rtp, chen2024lins} have tackled communication efficiency and mitigated communication costs. ZeRO++~\citep{wang2023zero++} redundantly stores an additional set of secondary parameters on each node, enhancing communication efficiency through parameter prefetching. MiCS~\citep{Zhang2022MiCSNS} and Fully Sharded Data Parallel (FSDP)~\citep{zhao2023pytorch} shard all model state components within subgroups and replicate them between subgroups to reduce communication scale. 

\subsubsection{GPU Memory Defragmentation \& Offloading}

Device memory limits affect manageable sequence length, requiring techniques like fragmentation elimination and offloading to expand capacity.

GPU memory defragmentation falls into two categories: tensor-based method~\citep{Kirisame2020DynamicTR, Hu2022MegTaiChiDT, shu2023roam, zhao2024efficientlytraining7bllm, zhang2024coop} and Virtual Memory Management (VMM). For tensor-based approaches, ROAM~\citep{shu2023roam} optimizes operator execution order and tensor allocation strategies using efficient tree-structured algorithms to identify optimal execution plans. MEMO~\citep{zhao2024efficientlytraining7bllm} and Coop~\citep{zhang2024coop} also address memory fragmentation while reducing overall memory consumption. VMM-based solutions, such as GMLake~\citep{guo2024gmlake} and PyTorch Expandable Segments~\citep{PyTorch2024Expandable_Segments}, utilize low-level CUDA driver APIs~\citep{CUDA2020virtual} to consolidate non-contiguous memory blocks into larger, contiguous segments through virtual memory address mapping.

Offloading technologies include CPU and SSD approaches. CPU offloading encompasses Static Offloading~\citep{pudipeddi2020training, ren2021zero} and Dynamic Offloading~\citep{sun2022stronghold, li2022harmony}. SSD Offloading solutions~\citep{rajbhandari2021zero, jang2024smart, liao2024adding} enable training of trillion-parameter models beyond CPU offloading capabilities.
Recent advancements have proposed comprehensive solutions for managing high activation value occupancy and memory fragmentation during training. \citet{zhao2024efficientlytraining7bllm} employs token-level decisions to determine which activation values to recompute and which to transfer to CPU memory, utilizing integer programming for memory allocation and space reuse by leveraging the uniform structure of Transformer layers. Ulysses-Offload~\citep{yao2024training} achieves substantial GPU memory reductions through its novel Distributed Attention with Fetching and Offloading mechanism, and leverages a dedicated double buffer design to overlap almost all fetching with computation.

\subsection{Enhancing Model FLOPs Utilization}\label{sec6_3}

Despite access to large-scale GPU clusters, LLaMA3.1~\citep{dubey2024llama} achieves a mere 38-41\% Model FLOPs Utilization (MFU), suggesting substantial room for optimization. These inefficiencies~\citep{duan2024efficient} are exacerbated when handling longer context (e.g. longer than 32k).

\begin{itemize}[leftmargin=2em]
\item Data processing operations, including sequence packing and tokenization, encounter significant challenges with extended sequences.
\item Longer sequence length results in quadratic growth in attention computation complexity. The memory bandwidth of current accelerator cards lags behind this computational surge, leading to longer processing times and reduced MFU.
\item Different sequence lengths from 2k to 128k and above complicate load balancing and efficient scheduling.
\end{itemize}

\subsubsection{Training Data Pipeline for Long-Context Models}

Processing longer sequences introduces specific challenges in the training data pipeline, particularly in text sorting, packing, and tokenization. While research in this area remains limited, the training data pipeline for long-context training is a critical challenge that warrants further investigation, as discussed in  \textbf{\nameref{q8_balance}} in Section\ref{sec12}.

Training only on long data hurts models' long-context performance~\citep{gao2024train}. The conventional approach of batch-packing sequences of similar lengths introduces potential training biases through length uniformity, while random long \& short-sequence packing results in GPU underutilization. To address this, GLM-Long~\citep{chatglm2024glmlong} organizes batches based on computational complexity, ensuring uniform computational complexity across packages and significantly reducing GPU idle periods. Furthermore, GLM-Long employs layer accumulation techniques to mitigate sorting-induced biases and utilizes loss reweighting strategies to handle imbalanced data volumes across packages. 

Tokenization inherently allows for parallel processing along the sequence dimension. ParallelTokenizer~\citep{cai2024internlm2, ParallelTokenizer} leverages this by implementing parallel tokenization.

\subsubsection{Operator Optimization}

Optimizing operators primarily involves enhancing the Transformer's core computation—the attention mechanism. FlashAttention~\citep{dao2022flashattention, daoflashattention} represents a significant advancement in this domain by optimizing memory access patterns through block-wise computations, enabling efficient use of on-chip fast memory. This approach reduces latency without compromising attention accuracy and eliminates quadratic memory complexity, thereby supporting long-context training. FlashAttention-3~\citep{shah2024flashattention} further optimizes for H100 GPUs by fully utilizing hardware features such as asynchronous WGMMA instructions. Simultaneously, normalization, dropout and feed-forward network (FFN) computations have undergone engineering optimizations~\citep{liu2023ring, Ma2024ReducingTC, Shoeybi2019MegatronLMTM}, often through operator fusion. For instance, the JAX implementation of Ring Attention with Blockwise Transformers~\citep{liu2023ring} incorporates operator fusion for FFN, enhancing computational efficiency. Native Sparse Attention (NSA)~\citep{minimax2025minimax01scalingfoundationmodels} introduces a hardware-aligned sparse strategy with dynamic token compression and selection, achieving substantial speedups by writing a triton kernel.

Compiler-level optimizations have also made significant strides, particularly with OpenAI Triton~\citep{tillet2019triton} and other frameworks~\citep{dong2024flex, Spector2024ThunderKittensSF}. Triton offers a Python-based programming language and an MLIR-based~\citep{lattner2020mlir} compiler enriched with built-in optimizations, facilitating the development of high-performance operators through a user-friendly interface. Additionally, compiler-level operator fusion, which often requires comprehensive computation graph information~\citep{chen2018tvm, PyTorch2024torch_compiler, wu2024multi}, automates optimization processes, thereby improving MFU.

\subsubsection{Scheduling Optimization}

Scheduling optimization is critical for enhancing training efficiency in long-context LLMs. As LLMs scale and context window size increases, factors such as computation-communication overlap~\citep{wang2024hiding}, load balancing, and CPU time significantly influence training speed (tokens per GPU per second)~\citep{dubey2024llama, Deepseek2024DeepSeek-V3}.  Given the limited research specifically for typical long-context, this section provides a concise overview.

Recent workload-scheduling developments have been tailored to LLMs. \citet{xue2024codesign} optimizes concurrent training efficiency through hybrid parallel strategies and hardware affinity in heterogeneous clusters. Hydro~\citep{hu2023hydro} enhances hardware utilization through model scaling and consolidation, while \citet{hu2024characterization} addresses mixed workload characteristics through solutions such as decoupled evaluation scheduling.

Resource-level improvements have also emerged. For example, SiloD~\citep{zhao2023silod} jointly allocates data caching and remote I/O as first-class resources, significantly improving system throughput. 

\subsubsection*{mixed-precision training}
In addition to the aforementioned methods, there are numerous approaches~\citep{narang2017mixed, kalamkar2019study, sun2019hybrid, dubey2024llama, qwen2024qwen25technicalreport, Deepseek2024DeepSeek-V3} that improve the long context training throughput and MFU from the perspective of mixed-precision training. \citet{wang2023bitnet} explores 1-bit precision training. Recent hardware and framework developments~\citep{xi2024jetfire, xi2023training, jacobs2023deepspeed, Shoeybi2019MegatronLMTM, Bian2021ColossalAIAU, peng2023fp8, torchtitan, meta_lingua, NVIDIA2024Transformer_Engine} have expanded support for lower precision operations (in FP8, FP4, INT4, etc.), offering new avenues for further enhancing MFU.

\section{Inference Infrastructure}\label{sec7}

\tikzstyle{my-box}=[
	rectangle,
	draw=black, 
	rounded corners,
	text opacity=1,
	minimum height=1.5em,
	minimum width=5em,
	inner sep=2pt,
	align=center,
	fill opacity=.5,
	line width=0.8pt,
]
\tikzstyle{leaf}=[my-box, minimum height=1.5em,
	text=black, align=left,font=\normalsize,
	inner xsep=10pt,
	inner ysep=4pt,
	line width=0.8pt,
]

\begin{figure*}[t]
	\centering
	\resizebox{\textwidth}{!}{
		\begin{forest}
			forked edges,
			for tree={
				grow=east,
				reversed=true,
				anchor=base west,
				parent anchor=east,
				child anchor=west,
				base=center,
				font=\large,
				rectangle,
				draw=black, 
				rounded corners,
				align=left,
				text centered,
				minimum width=4em,
				edge+={darkgray, line width=1pt},
				s sep=3pt,
				inner xsep=2pt,
				inner ysep=3pt,
				line width=0.8pt,
				ver/.style={rotate=90, child anchor=north, parent anchor=south, anchor=center, minimum width=16em, fill=gray!10},
			},
			where level=1{text width=15em,font=\normalsize,fill=green!10}{},
			where level=2{text width=15em, align=left, font=\normalsize, fill=orange!10}{},
			where level=3{text width=30em}{},
			where level=4{text width=18em,font=\normalsize,}{},
			where level=5{text width=18em,font=\normalsize,}{},
			[
				\textbf{Inference Infrastructure}, ver
                    [
                        \textbf{Memory Optimization~(\S\ref{sec7_1})}
                        [
                            \textbf{Defragmentation}
                            [
                                \textbf{Virtual Memory on Device \& Paging Mechanisms}, leaf
                            ]
                        ]
                        [
                            \textbf{Footprint Reduction}
                            [
                                \textbf{Traditional Methods}, leaf
                            ]
                            [
                                \textbf{Fine-grained Memory Management}, leaf
                            ]
                        ]
                    ]
                    [
                        \textbf{Computation Optimization~(\S\ref{sec7_2})}
                        [
                            \textbf{System-level Optimization}
                        ]
                        [
                            \textbf{Redundancy Elimination}
                        ]
                        [
                            \textbf{KV Cache Reuse}
                            [
                                \textbf{Prefix Sharing}, leaf
                            ]
                            [
                                \textbf{Approximation Method}, leaf
                            ]
                        ]
                    ]
                    [
                        \textbf{Distributed Processing~(\S\ref{sec7_3})}
                        [
                            \textbf{Distributed Attention}
                        ]
                        [
                            \textbf{Scheduling Strategies}\\ \newline\textbf{of Inference Service}
                            [
                                \textbf{Disaggregated Inference}, leaf
                            ]
                            [
                                \textbf{Other Resource Partitioning \& Scheduling Methods}, leaf
                            ]
                        ]
                    ]
			]
		\end{forest}
	}
	\caption{An overview of inference infrastructure of long-context LLMs.}
	\label{fig:infer_infra}
\end{figure*}
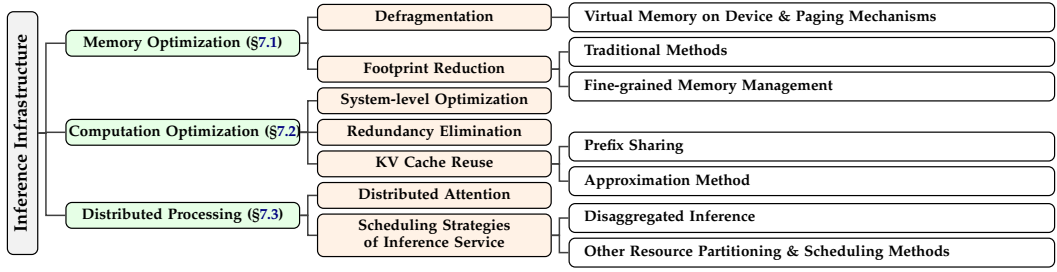

Developing effective strategies for long-context inference represents a strategic imperative for both industry and academia. In today's business landscape where API sales and Agent products dominate, efficient handling of longer contexts is essential~\citep{Google2024why-do-they-matter, koh2024visualwebarena}. Meanwhile, researchers have noted inherent limits on current pretrain paradigms~\citep{reuters2024Ilya}, especially as the growth of high-quality training data slows. 

As context lengths extend to tens of thousands or even millions of tokens~\citep{anthropic2024claude2, reid2024gemini}, inference encounters bottlenecks including quadratic complexity of attention mechanism, KV cache storage demands, communication overhead and other challenges~\citep{li2024llm, yuan2024llm}. These technical barriers directly impact inference systems' \textbf{throughput} and \textbf{latency}.

Researchers have tried to improve throughput by refining memory utilization~\citep{sheng2023flexgen}, optimizing batching techniques to maximize parallelism~\citep{anyscale2024selective}, etc. At the same time, efforts to reduce latency include but are not limited to, minimizing redundant attention calculations~\citep{jiang2024minference}, reusing KV cache~\citep{zheng2024sglang} and making the prefill and decode phases disaggregated~\citep{jin2024p}. Lastly, for contexts of hundreds of thousands or millions of tokens, there are scalable distributed solutions~\citep{fang2024uspunifiedsequenceparallelism, lin2024infinite, wu2024loongserve}.

This section ends with a curated overview of popular inference frameworks~\citep{kwon2023efficient, 2023lmdeploy, zheng2024sglang, huggingface2024huggingfaceTGI, NVIDIA2024tensorrt-llm} to guide readers in their research and deployment decisions, reflecting how today's inference engines have matured into sophisticated platforms that integrate recent findings~\citep{Gyeong280922Orca, daoflashattention, dao2022flashattention, agrawal2024taming, jin2024p} with best engineering practices.

\subsection{Memory Optimization}\label{sec7_1}

The pursuit of higher throughput has led us to optimize GPU memory usage in LLM inference systems~\citep{semianalysis2023BlackwellInferenceTraining, kwon2023efficient, zheng2024sglang}, as the growing demands of processing long sequences pose some challenges for GPU memory, which we will discuss in the following.

\subsubsection{GPU Memory Defragmentation}

PagedAttention~\citep{kwon2023efficient} leverages virtual memory paging mechanisms similar to those operating systems use to manage the KV cache on fixed-size pages. TokenAttention~\citep{LightLLM2024token_attention, hu2024lightllm} manages the KV cache at the token level, achieving zero GPU memory waste. vAttention~\citep{prabhu2024vattention, xu2024vtensor}, leverages CUDA's native virtual memory management capabilities~\citep{CUDA2020virtual}, eliminates PagedAttention-like lookup tables, resulting in reduced latency.

\subsubsection{Memory Footprint Reduction}

\paragraph{Traditional Methods}
Chunk prefill~\citep{agrawal2024taming, holmes2024deepspeed, zeng2024memorize} divides long sequences into smaller blocks for gradual processing to reduce GPU memory pressure or batch them together with decoding requests to improve overall throughput. Approximate attention mechanisms, cache-free and other non-attention architectures shown in Section \ref{sec5} can significantly reduce GPU memory costs for long-sequence computations and KV cache. Cache optimization techniques shown in Section \ref{sec3} can substantially reduce deployment memory overhead while improving processing speed through low-precision advantages.

\paragraph{Fine-grained Memory Management}
Extended sequence length has necessitated more sophisticated memory management approaches. Researchers have introduced fine-grained memory management techniques~\citep{sheng2023flexgen, he2024fastdecode, jiang2024neo, gao2024attentionstore, lee2024infinigen}. FlexGen~\citep{sheng2023flexgen} uses linear programming to select optimal storage formats and access patterns for weights and attention cache. 

The CPU memory and disk offloading~\citep{liu2023deja} need management too. Frameworks like DeepSpeed-inference~\citep{aminabadi2022deepspeed} and Huggingface Accelerate~\citep{Huggingface2022accelerate} offload the weights of large models to CPU memory. \citet{alizadeh2023llm} enables models up to twice the size of available DRAM to run by the combination of a low-rank predictor for selective neuron loading, a dynamic sliding window technique for caching activated neurons, and a row-column bundling mechanism to optimize data transfers between flash storage and DRAM.

\subsection{Computation Optimization}\label{sec7_2}

Attention computation costs grow quadratically with sequence length, creating significant latency challenges for long-context inference~\citep{beltagy2020longformer, liu2024reattention}. Recent Studies address this by optimizing system-level implementation, reducing unnecessary calculations in attention and reusing existing results.

\paragraph{System-level Implementation Optimization}
this type of work focuses purely on engineering and implementation optimizations without modifying the underlying algorithms~\citep{daoflashattention, dao2022flashattention, shah2024flashattention, cascade-inference, FlashInfer0.2, ye2025flashinfer, llama.cpp}. For example, FlashDecoding++~\citep{hong2023flashdecoding++} accelerates flat GEMM~\citep{FastGEMV, ibrahim2024balanced} with double buffering that overlaps computation and data transfer, hiding the memory latency in loading input matrices. Continuous batching~\citep{Gyeong280922Orca, anyscale2024selective, kwon2023efficient} allows new sequences to be inserted into a batch whenever existing sequences complete their generation, yielding higher GPU utilization compared to static batching. In a similar vein, Lightning Attention~\citep{minimax2025minimax01scalingfoundationmodels} introduces several system-level optimizations, such as batched kernel fusion and the separation of prefill and decoding execution. These innovations improve memory access efficiency and reduce latency in long-context inference, particularly for heterogeneous batch inputs.

\paragraph{Computational Redundancy Elimination}
Research has revealed that attention patterns are notably sparse~\citep{xiaoefficient, jiang2024minference}, with only a small subset of tokens significantly impacting next-token prediction. This insight has led to many optimization strategies that are discussed in Section \ref{sec5}.

\paragraph{KV Cache Reuse}
In practical applications, context often contains repetitive segments, while recalculating these increases latency with longer contexts~\citep{gim2024prompt}. Early approaches used simple prefix matching for cache reuse or prefix sharing in decoding~\citep{juravsky2024hydragen, cascade-inference}, integrated into deployment frameworks~\citep{NVIDIA2024tensorrt-llm, FlashInfer0.1, FlashInfer0.2, lin2024parrot} rather than published as standalone work. RadixAttention~\citep{zheng2024sglang} later improved this by organizing contexts in a radix tree structure, enabling efficient reuse with minimal CPU overhead and across requests. Another research direction employs approximation methods~\citep{hu2024epic, yao2024cacheblend} to reuse KV cache across requests with partially matching prefixes, where identical segments are not contiguous. This requires careful handling of internal attention and position embedding while approximating cross-segment attention. For instance, EPIC~\citep{hu2024epic} introduced position-independent context caching, enabling flexible cache reuse across positions without affecting model accuracy. 

\subsection{Distributed Processing}\label{sec7_3}

When context length extends to hundreds of thousands or even millions of tokens~\citep{yang2024qwen2technicalreport, qwen2024qwen25technicalreport, reid2024gemini, InternLM25}, the memory and computational capabilities of a single machine with a single GPU can no longer meet the demands. This section briefly discusses existing distributed solutions for enhancing long-context processing capabilities, focusing on Distributed Attention, scheduling strategies, and the increasingly popular Prefill-Decode (PD) disaggregation architecture.

\subsubsection{Distributed Attention}

Ring Attention~\citep{li2021sequence} enables efficient processing of long sequences by splitting them across devices. Each device stores a portion of the KV cache, reducing GPU memory usage. Since the data transfer and computation can be fully overlapped through optimization~\citep{liu2023ring, fang2024uspunifiedsequenceparallelism}, the additional communication overhead does not impact throughput. When combined with Context Parallel~\citep{Shoeybi2019MegatronLMTM}, this method could enable a longer context.

\citet{yang2024context} demonstrates near-linear scaling in long-context prefill latency through two approaches: Pass-KV, which transfers Key and Value matrices between GPUs for KV cache reuse, and Pass-Q, which transfers only Query matrices to reduce bandwidth and latency during decoding. For further exploration of how recent research has enhanced the efficiency of distributed attention, please refer to Section \ref{sec5} and \ref{sec6}.

\subsubsection{Scheduling Strategies of Inference Service}

Currently, inference service providers face two key challenges~\citep{sun2024llumnix}: the unpredictable nature of input lengths and the lack of effective scheduling strategies. As the demand for processing long texts continues to grow, the variability in input lengths has expanded, further complicating the situation~\citep{semianalysis2024ScalingLawO1Pro}. Without proper scheduling strategies, inference systems using traditional tensor, pipeline, and data parallelism alone would be less efficient at large cluster scales~\citep{guo2024survey}.

\paragraph{Disaggregated Inference}
The prefill and decoding stage of LLM inference have fundamentally different characteristics and resource requirements~\citep{Venkat2024EssentialMath, patel2024splitwise, qin2024mooncake}:
\begin{itemize}[leftmargin=2em]
    \item prefill is computationally intensive with its superlinear scaling with batch size and sequence length. Time to First Token (TTFT) is an important metric for this stage.
    \item decoding is memory(bandwidth)-constrained with its sublinear scaling with batch size. Time Between Tokens (TBT) and end-to-end latency are key metrics.
\end{itemize}

Given these differences, disaggregating the two stages~\citep{patel2024splitwise, zhong2024distserve, qin2024mooncake, hu2024inference, jin2024p} enables targeted optimization of tasks with two distinct computational characteristics, balancing computational efficiency, memory utilization, and latency requirements through independent resource pools and scheduling strategies, improving both latency and throughput.

\paragraph{Other Resource Partitioning \& Scheduling Methods}

Several innovative approaches have been proposed for resource partitioning and scheduling in LLM inference~\citep{lin2024parrot, hu2024memserve, lin2024infinite, srivatsa2024preble, wu2024loongserve}
. Infinite-LLM~\citep{lin2024infinite} allows the independent scheduling and resource allocation for non-attention layers and improves system scalability through a two-tier global and local scheduling strategy. Co-optimizing KV state reuse and computation load-balancing, Preble~\citep{srivatsa2024preble} is the first distributed LLM serving platform that targets prompt sharing. Elastic Sequence Parallelism~\citep{wu2024loongserve} dynamically adjusts to resource usage variations for prefill and decode stages, reducing KV cache migration overhead and fragmentation.

Multi-level cache management has emerged as another key optimization strategy, with several studies~\citep{jiang2024neo, qin2024mooncake, song2024powerinfer, Deepseek2024DeepSeek-V3} utilizing hierarchical distributed caches across GPUs, CPUs, DRAM, and SSDs. These studies implement load-aware scheduling and pre-estimate input/output lengths to optimize resource utilization.

\subsection*{Open Source Frameworks}

Open-source frameworks have proven effective for handling context lengths of up to 100k tokens. More recent frameworks have optimized sequence processing through structured output~\citep{zheng2024sglang} and cache reuse while maintaining high throughput~\citep{kwon2023efficient, zheng2024sglang}. Organizations and famous enterprises have also released open-source inference frameworks~\citep{qin2024mooncake, NVIDIA2024tensorrt-llm, zhihu2024ZhiLight, 2023lmdeploy, huggingface2024huggingfaceTGI}, each offering unique features. The accumulated engineering expertise from these projects has enriched technical options and advanced the field toward maturity.


\paragraph{vLLM} 
Developed by the University of California, Berkeley, vLLM~\citep{kwon2023efficient} is renowned for its PagedAttention mechanism and strong open-source community support. It supports a wide range of models, including multimodal and non-Transformer architectures, and is compatible with diverse hardware. The upcoming version 1.0 will address previous limitations such as reliance on serial scheduling, limited graph optimization, and complex codebases that hinder further development.

\paragraph{SGLang}
Also from UC Berkeley, SGLang~\citep{zheng2024sglang} is primarily written in Python and optimized with the torch.compile tool. It features optimizations like Radix Attention, structured output enhancements, and multi-process GMP transmission, which significantly reduce CPU overhead.

\paragraph{LMDeploy}
Developed by SenseTime and the Shanghai AI Laboratory, LMDeploy~\citep{2023lmdeploy} provides implementations based on both CUDA and Triton acceleration. This framework supports multimodal tasks effectively and includes several commonly used pre-trained models.

\paragraph{Huggingface's Text Generation Inference}
Huggingface's Text Generation Inference (TGI)~\citep{huggingface2024huggingfaceTGI} also utilizes the PagedAttention mechanism and employs Rust for low-level functions and Python (70\%) for higher-level layers. Despite this, its throughput performance is average, particularly with larger batch sizes, due to decreased GPU memory management efficiency. Additionally, its CPU-GPU serial scheduling design limits GPU resource utilization.

\paragraph{TensorRT-LLM}
TensorRT-LLM~\citep{NVIDIA2024tensorrt-llm} is NVIDIA's open-source framework on their GPUs. The framework stands out for its comprehensive optimization of popular LLMs, multiple NVIDIA hardware platforms(H100, L40, A100, V100, T4, etc.), flexible customization of plugins and kernels, and seamless multi-GPU/multi-node deployment capabilities. 

\section{Long-Context Pre-training}\label{sec8}

The development of deployment and training infrastructure has enabled the training and inference of LLMs with longer contexts. In this background, the pre-training length of LLMs has evolved from the initial 2k tokens~\citep{touvron2023llama} to 4k~\citep{touvron2023llama2}, 32k~\citep{xiaoefficient,cai2024internlm2}, over 128k~\citep{meta2024introducing,InternLM25}, and even 1M~\citep{liu2024world}. To expand the context length of LLMs effectively, more training strategies specialized for long-context LLMs are necessary. We begin our analysis from the long-context pre-training. Compared to the preceding short-context pre-training, long-context pre-training is featured with requiring fewer tokens, generally 1B-10B, and facing both challenges of quality and quantity~\citep{fudata,lv2024longwanjuan}.

\subsection{Long-Context Data Quality}

In the earliest works, researchers often focused on the length of pre-training~\citep{chen2023extending,roziere2023code,pengyarn}, with little discussion of other factors. Subsequently, ScalingRoPE first discovers that continual pre-training at the original pre-training context length could extrapolate the context length of LLMs~\citep{liuscaling}. LLaMA2Long~\citep{xiong2024effective} further points out that in long-context pre-training, data quality is more crucial than data length and provides detailed discussions on the mixing ratio and training cycles between long and short data.

Following this, \citet{fudata} first raises the concept of long-context data engineering and suggests that the data required for long-context training is much less than that for short-context pre-training. Only 0.5B to 5B tokens are enough. Instead of relying solely on long book and long paper data, \citet{fudata} also emphasizes that, besides length up-sampling, it is essential to maintain balance across domains, which has gained widespread acceptance~\citep{zhang2406long,young2024yi,chatglm2024glmlong,gao2024train}. Recently, \citet{gao2024train} conducts an in-depth investigation into long-context training, finding that mixing code repositories and long books with high-quality short-context data is crucial for both long-context performance and retaining the short-context capabilities. The exploration of long-short-mixing training inspires thinking about training long-context LLMs from scratch, which will be discussed in \textbf{\nameref{q8_balance}} in Section\ref{sec12} 

Regarding the quality of a single long data sample, LongWanjuan~\citep{lv2024longwanjuan} is the first to propose that using LLM-based or rule-based metrics could reflect whether a long text exhibits long-context dependency characteristics from the perspective of coherence, cohesion, and diversity. It then categorizes long texts into holistic, aggregated, and chaotic types and conducts data mixing to achieve optimal long-context training results. ProLong~\citep{chen2024long} goes deeper into long-context dependencies, designing scores for dependency strength, dependency distance, and dependency specificity to measure long-distance dependencies between different segments in a long text, for data filtering.

\subsection{Long-Context Data Curation} 

Discussions on long-context data quality remain very limited, primarily because long-context data itself is extremely scarce, leading to a greater focus on data synthesis~\citep{chatglm2024glmlong}. In early long-context training, researchers employ the simplest splicing methods to obtain sufficient long-context data~\citep{chenlonglora,tworkowski2024focused,chenclex,li2023functional}. Notably, CodeLLaMA utilized the feature of code data to concatenate code from the same project, resulting in ultra-long code datasets~\citep{roziere2023code}.

Subsequent efforts begin to stitch similar short texts into a long context through similarity matching. For instance, ICLM~\citep{shicontext} constructs a graph of documents with embeddings from an encoder-only model and applies the traveling salesman algorithm to extract efficiently. SPLiCe~\citep{staniszewski2023structured} replaces selection criteria with BM25 retrieval or attribute label matching and extends the splicing length to 32k. BM25Chunk~\citep{zhao2024analysing} provides in-depth analysis for training on concatenated long-context data, while later work explored retrieval methods using LLM embeddings~\citep{chatglm2024glmlong} and keyword matching~\citep{gao2024quest}. DataSculpt attempted to optimize the synthesis of spliced data through multi-objective combinatorial optimization~\citep{lu2024datasculpt}.

In addition to sequential splicing, a few works have attempted to achieve extended length through interleaved splicing of short texts~\citep{zhao2024longskywork,tian2024untie}. LongSkywork proposes CIP~\citep{zhao2024longskywork}, which splits, shuffles, and splices short texts, allowing LLMs to identify relevant segments within seemingly chaotic contexts through self-attention adaptively, thus enhancing long-context modeling capabilities. Following this, UTK~\citep{tian2024untie} introduces knot tokens pushing LLMs to untie these knots and gain long-context capabilities more effectively. These methods could significantly improve the performance of synthetic tasks such as RULER~\citep{hsieh2024ruler}.

Additionally, a few pieces of research concern loss design specialized for long-context training~\citep{fang2024wrong}. Discussions regarding long-context pre-training work are still limited, which we will highlight and summarize in \textbf{\nameref{q7_scarce}} in Section\ref{sec12}, and much of the discourse is dispersed across various technical reports of LLMs. We have compiled these technical reports of long-context LLMs, listing the information related to long-context pre-training, post-training, and evaluation, for the reader's reference.

\begin{table}[!ht]
\renewcommand{\arraystretch}{1.35}
\tabcolsep=0.1cm
\centering
\small
    \resizebox{\textwidth}{!}{
    \begin{tabular}{cccccc}
    \toprule
    \textbf{Model} & \textbf{Organization} & \textbf{Time} & \textbf{Version} & \textbf{Context Length} & \textbf{Benchmark} \\ 
    \midrule
    \multirow{3}{*}{ChatGPT~\citeyearpar{chatgpt2022}} & \multirow{3}{*}{OpenAI} & \multirow{3}{*}{22.11} & gpt-3.5-turbo & 4K & \multirow{3}{*}{-} \\ 
     &  &  & gpt-3.5-turbo-instruct & 4K & ~ \\ 
     &  &  & gpt-3.5-turbo-0125 & 16K & ~ \\ 
    \multirow{2}{*}{GPT-4~\citeyearpar{gpt4}} & \multirow{2}{*}{OpenAI} & \multirow{2}{*}{23.03} & (default) & \multirow{2}{*}{128K} & \multirow{2}{*}{-} \\
     &  &  & turbo & ~ & ~ \\ 
    \multirow{2}{*}{GPT-4o~\citeyearpar{gpt4}} & \multirow{2}{*}{OpenAI} & \multirow{2}{*}{24.05} & (default) & \multirow{2}{*}{128K} & \multirow{2}{*}{-} \\ 
     &  &  & mini &  & ~ \\ 
    \multirow{2}{*}{OpenAI-o1~\citeyearpar{OpenAI2024o1}} & \multirow{2}{*}{OpenAI} & \multirow{2}{*}{24.09} & (default) & 200K & \multirow{2}{*}{-} \\ 
     &  &  & mini & 128K & ~ \\ 
     \midrule

    Claude~\citeyearpar{anthropic2023claude} & Anthropic & 23.03 & (default) & - & - \\ 
    \multirow{2}{*}{Claude2~\citeyearpar{anthropic2024claude2}} & \multirow{2}{*}{Anthropic} & \multirow{2}{*}{23.07} & (default) & 100K & \multirow{2}{*}{-} \\ 
     &  &  & 2.1 & 200K & ~ \\ 
    \multirow{3}{*}{Claude3~\citeyearpar{anthropic2024claude3}} & \multirow{3}{*}{Anthropic} & \multirow{3}{*}{24.03} & Haiku & \multirow{3}{*}{200K} & \multirow{3}{*}{NIAH} \\ 
     &  &  & Sonnet & ~ & ~ \\ 
     &  &  & Opus & ~ & ~ \\ 
    \multirow{3}{*}{Claude3.5~\citeyearpar{anthropic2024claude3}} & \multirow{3}{*}{Anthropic} & \multirow{3}{*}{24.06} & Haiku & \multirow{3}{*}{200K} & \multirow{3}{*}{-} \\ 
     &  &  & Sonnet & ~ & ~ \\ 
     &  &  & Opus & ~ & ~ \\ 
     \midrule
    \multirow{3}{*}{Gemini~\citeyearpar{team2023gemini}} & \multirow{3}{*}{Google} & \multirow{3}{*}{23.12} & Ultra & \multirow{3}{*}{32K} & \multirow{3}{*}{SCROLLS} \\ 
     &  &  & Pro &  &  \\ 
     &  & & Nano & ~ & ~ \\ 
    \multirow{2}{*}{Gemini-1.5~\citeyearpar{reid2024gemini}} & \multirow{2}{*}{Google} & \multirow{2}{*}{24.02} & Pro & \multirow{2}{*}{1M} & \multirow{2}{*}{NIAH, LQA, LICL} \\ 
     &  &  & Flash &  & ~ \\ 
    \multirow{2}{*}{Gemini-2.0~\citeyearpar{google2024gemini2}} & \multirow{2}{*}{Google} & \multirow{2}{*}{24.12} & Pro & \multirow{2}{*}{1M} & \multirow{2}{*}{LQA} \\ 
     &  &  & Flash &  &  \\
     
     \midrule
     Kimi-chat~\citeyearpar{kimi} & MoonshotAI & 23.11 & (default) & 2M & NIAH\\ 
     Kimi-K1.5~\citeyearpar{team2025kimi} & MoonshotAI & 25.01 & (default) & 2M & -\\
     \midrule
     
     AFM~\citeyearpar{gunter2024apple} & Apple & 24.07 & (default) & 32k & LQA\\ \midrule

     abab~\citeyearpar{minimax2024} & MiniMax & 24.04 & \makecell[c]{6.5s\\7} & 240k & NIAH\\ \midrule  

     Step-1~\citeyearpar{step12024} & Step & 24.03 & (default) & 256k & - \\ 
     Step-2~\citeyearpar{step12024} & Step & 24.07 & (default) & 16k & - \\ 
     \bottomrule
    \end{tabular}
    }
    \label{table:close_source_model}
    \caption{Comparison of mainstream close-source long-context LLMs. The symbol “-” indicates that no relevant information was found. \textit{Benchmark} refers to the long-context benchmarks used in the evaluation. Specifically, \textit{PPL} stands for perplexity, \textit{LQA} for Long QA, \textit{LC} for Long Code, and \textit{LICL} for Long In-Context Learning.}
\end{table}
\begin{table}[!ht]
\renewcommand{\arraystretch}{1.35}
\tabcolsep=0.05cm
    \centering
    \resizebox{\textwidth}{!}{
    \rotatebox{90}{
\begin{tabular}{>{\centering\arraybackslash}m{2.75cm}>{\centering\arraybackslash}m{2cm}>{\centering\arraybackslash}m{1cm}>{\centering\arraybackslash}m{2.25cm}>{\centering\arraybackslash}m{3cm}>{\centering\arraybackslash}m{1.25cm}m{3.5cm}m{3.5cm}>{\centering\arraybackslash}m{3.5cm}}
    \toprule
\textbf{Model} & \textbf{Organization} & \textbf{Time} & \textbf{Version} & \makecell[c]{\textbf{Architecture Detail}\\(Base-Q-KV)} & \makecell[c]{\textbf{Context}\\ \textbf{Length}} & \makecell[c]{\textbf{Pre-Training}\\ \textbf{Strategy}} & \makecell[c]{\textbf{Post-Training}\\ \textbf{Strategy}} & \textbf{Benchmark} \\ \midrule

LLaMA~\citeyearpar{touvron2023llama} & Meta & 23.03 & \makecell[c]{7B\\13B\\33B\\65B} & \makecell[c]{1e4-32Q-32KV\\1e4-40Q-40KV\\1e4-52Q-52KV\\1e4-64Q-64KV} & 2k & len=2k & - & - \\ 

LLaMA2~\citeyearpar{touvron2023llama2} & Meta & 23.07 & \makecell[c]{7B\\13B\\70B} & \makecell[c]{1e4-32Q-32KV\\1e4-40Q-40KV\\1e4-64Q-8KV} & 4k & len=4k & - & SCROLLS \\ 

LLaMA3~\citeyearpar{meta2024introducing} & Meta & 24.04 & \makecell[c]{8B\\70B} & \makecell[c]{5e5-32Q-32KV\\5e5-64Q-8KV} & 8k  & len=8k & - & - \\ 

LLaMA3.1$^\diamond$~\citeyearpar{dubey2024llama} & Meta & 24.07 & \makecell[c]{8B\\\\70B\\\\405B} & \makecell[c]{\makecell[c]{5e5-32Q-8KV\\freq 1,4; factor 8}\\ \makecell[c]{5e5-64Q-8KV\\freq 1,4; factor 8}\\ \makecell[c]{5e5-128Q-8KV\\freq 1,4; factor 8}} & 128k & len=8k$\to$128k; context parallelism & Iterative training; synthetic data  & LQA, LICL, ZeroSCROLLS, NIAH, InfiniteBench \\ 

LLaMA3.2$^\diamond$~\citeyearpar{meta2024llama} & Meta & 24.09 & \makecell[c]{1B\\\\3B\\\\11B} & \makecell[c]{\makecell[c]{5e5-32Q-8KV\\freq 1,4; factor 32}\\\makecell[c]{5e5-32Q-8KV\\freq 1,4; factor 32}\\\makecell[c]{5e5-32Q-8KV\\freq 1,4; factor 8}} & 128k & - & - & - \\ 

LLaMA3.3$^\diamond$~\citeyearpar{llama3_3} & Meta & 24.12 & 70B & \makecell[c]{5e5-64Q-8KV\\freq 1,4; factor 8} & 128k & -& -& -\\

\midrule
Gemma~\citeyearpar{team2024gemma} & Google & 24.03 & \makecell[c]{2B\\7B} & \makecell[c]{1e4-8Q-1KV\\1e4-16Q-16KV} & 8k & len=8k & - &- \\ 

Gemma2$^\flat$~\citeyearpar{team2024gemma2} & Google & 24.06 & \makecell[c]{3B\\\\9B\\\\27B} & \makecell[c]{\makecell[c]{1e4-8Q-4KV\\Sliding Window=4096}\\\makecell[c]{1e4-16Q-8KV\\Sliding Window=4096}\\\makecell[c]{1e4-32Q-16KV\\Sliding Window=4096}} & 8k & len=8k & - & - \\
\midrule
Mistral-v0.1$^\flat$~\citeyearpar{jiang2023mistral} & MistralAI & 23.1 & 7B & \makecell[c]{1e4-32Q-8KV\\Sliding Window=4096} & 8k & - & - & - \\ 
Mistral-v0.2~\citeyearpar{jiang2023mistral} & MistralAI & 23.11 & 7B & 1e6-32Q-8KV & 32k & - & - & - \\
Mistral-v0.3~\citeyearpar{jiang2023mistral} & MistralAI & 24.1 & 7B & 1e6-32Q-8KV & 32k & - & - & - \\ 
\bottomrule

    \end{tabular}
    }
    }
    \caption{Comparison of mainstream open-source long-context LLMs. The symbol "-" indicates that no relevant information was found. \textit{Architecture Details} is composed of \textit{Base-Q-KV}, which respectively represent the RoPE Base, num\_attention\_heads and num\_kv\_heads. If RoPE is not used, the type of positional encoding employed will be specified in the \textit{RoPE Base} field. The symbol "$\diamond$" indicates that Scaling RoPE is used and we provide the scaling frequency and scaling factor below the \textit{Base-Q-KV}. The symbol "$\flat$" indicates that Sliding Window Attention is used and we provide the sliding window below the \textit{Base-Q-KV}. \textit{Context Length} refers to the maximum length of context that the model can process. \textit{Pre-Training Strategy} and \textit{Post-Training Strategy} refer to the strategies employed by the model for handling long contexts during the respective pre-training and post-training phases. Additionally, we provide the context lengths (denoted as \textit{len}) used during long-context training, as specified in the technical reports. \textit{Benchmark} refers to the long-context benchmarks used in the evaluation. Specifically, \textit{PPL} stands for perplexity, \textit{LQA} for Long QA, \textit{LC} for Long Code, and \textit{LICL} for Long In-Context Learning.}
    \label{table:open_source_model_p1}
\end{table}
\clearpage

\begin{table}[!ht]
\renewcommand{\arraystretch}{1.35}
\tabcolsep=0.1cm
    \centering
    \resizebox{\textwidth}{!}{
    \rotatebox{90}{
    \begin{tabular}{>{\centering\arraybackslash}m{2.75cm}>{\centering\arraybackslash}m{2cm}>{\centering\arraybackslash}m{1cm}>{\centering\arraybackslash}m{2.25cm}>{\centering\arraybackslash}m{3cm}>{\centering\arraybackslash}m{1.25cm}m{3.5cm}m{3.5cm}>{\centering\arraybackslash}m{3.5cm}}
    \toprule
\textbf{Model} & \textbf{Organization} & \textbf{Time} & \textbf{Version} & \makecell[c]{\textbf{Architecture Detail}\\(Base-Q-KV)} & \makecell[c]{\textbf{Context}\\ \textbf{Length}} & \makecell[c]{\textbf{Pre-Training}\\ \textbf{Strategy}} & \makecell[c]{\textbf{Post-Training}\\ \textbf{Strategy}} & \textbf{Benchmark} \\ \midrule

phi-3~\citeyearpar{abdin2024phi3} & Microsoft & 24.04 & \makecell[c]{Phi-3.5-MoE\\Phi-3.5-Mini} & \makecell[c]{1e4-32Q-8KV\\1e4-32Q-32KV} & 128k & Long-RoPE & - & RULER, LC \\ 
phi-4~\citeyearpar{abdin2024phi4} & Microsoft & 24.12 & Phi-4-14B & 2.5e5-40Q-10KV & 16k & len=4k; Mix long and short context & - & HELMET \\ 
\midrule 

Falcon~\citeyearpar{almazrouei2023falcon} & TII & 23.11 & \makecell[c]{7B\\40B\\180B} & \makecell[c]{1e4-71Q-1KV\\1e4-128Q-8KV\\1e4-232Q-8KV} & 2k & len=2k & - & - \\ 
Falcon2~\citeyearpar{malartic2024falcon2} & TII & 24.07 & 11B & 5e5$^\star$-32Q-8KV & 8k & len=2k$\to$8k & - & - \\
Falcon3~\citeyearpar{Falcon3} & TII & 23.12 & \makecell[c]{1B\\3B\\7B\\10B} & \makecell[c]{1e6$^\star$-8Q-4KV\\1e6$^\star$-12Q-4KV\\1e6$^\star$-12Q-4KV\\1e6$^\star$-12Q-4KV} & \makecell[c]{4k\\8k\\32k\\32k} & - & - & -\\
\midrule

Qwen~\citeyearpar{bai2023qwen} & Alibaba & 23.09 & \makecell[c]{1.8B\\7B\\14B\\72B} & \makecell[c]{1e4-16Q-16KV\\1e4-32Q-32KV\\1e4-40Q-40KV\\1e6-64Q-64KV} & \makecell[c]{8k\\32k\\32k\\32k} & len=2k; NTK & - & PPL \\ 

Qwen1.5~\citeyearpar{bai2023qwen} & Alibaba & 24.02 & \makecell[c]{0.5B\\1.8B\\4B\\7B\\14B\\32B\\72B} & \makecell[c]{1e6-16Q-16KV\\1e6-16Q-16KV\\5e6-20Q-20KV\\1e6-32Q-32KV\\1e6-40Q-40KV\\1e6-40Q-8KV\\1e6-64Q-64KV} & 32k & len=32k & - & L-Eval \\ 

Qwen2~\citeyearpar{yang2024qwen2technicalreport} & Alibaba & 24.07 & \makecell[c]{0.5B\\1.5B\\7B\\72B} & \makecell[c]{1e6-14Q-2KV\\1e6-12Q-2KV\\1e6-28Q-4KV\\1e6-64Q-8KV} & 128k & len=4k$\to$32k; RoPE base=1e4$\to$1e6; YaRN, DCA & - & NIAH, NeedleBench, LV-Eval \\

Qwen2.5~\citeyearpar{qwen2024qwen25technicalreport} & Alibaba & 24.09 & \makecell[c]{0.5B\\1.5B\\3B\\7B\\14B\\32B\\72B} & \makecell[c]{1e6-14Q-2KV\\1e6-12Q-2KV\\1e6-16Q-2KV\\1e6-28Q-4KV\\1e6-40Q-8KV\\1e6-40Q-8KV\\1e6-64Q-8KV} & \makecell[c]{128k\\128k\\128k\\128k\\128k\\128k\\128k} & len=32k$\to$256k; RoPE base=1e4$\to$1e6; YaRN, DCA & len=32k$\to$256k & RULER, LV-Eval, LongBench-chat \\ 
QwQ~\citeyearpar{qwq-32b-preview} & Alibaba & 24.11 & 32B-preview & 1e6-40Q-8KV & 32k & - & - & - \\\midrule

Index~\citeyearpar{Index} & Bilibili & 24.10 & 1.9B & 3.2e6-16Q-16KV & 32k & len=32k; Doc Packing & len=32k; Long SFT; Doc Packing & NIAH, LongBench, LEval \\ \midrule

MiniMax-01$^\natural$~\citeyearpar{minimax2025minimax01scalingfoundationmodels} & MiniMax & 25.01 & Text-01 & 1e7-64Q-8KV & 4M & len=32k$\to$1M & len=8k$\to$1M; Long SFT and Long RL & NIAH, RULER, LongBench-v2, MTOB\\

    \bottomrule
    \end{tabular}
    }
    }
    \caption{Continued table of Table \ref{table:open_source_model_p1}. The symbol $^\star$ indicates that the actual RoPE Base is the annotated value plus 42. The symbol $^\natural$ indicates that lightning attention is used. }
\label{table:open_source_model_p2}
\end{table}
\clearpage

\begin{table}[!ht]
\renewcommand{\arraystretch}{1.35}
\tabcolsep=0.1cm
    \label{table:open_source_model}
    \centering
    \resizebox{\textwidth}{!}{
    \rotatebox{90}{
\begin{tabular}{>{\centering\arraybackslash}m{2.75cm}>{\centering\arraybackslash}m{2cm}>{\centering\arraybackslash}m{1cm}>{\centering\arraybackslash}m{2.25cm}>{\centering\arraybackslash}m{3cm}>{\centering\arraybackslash}m{1.25cm}m{3.5cm}m{3.5cm}>{\centering\arraybackslash}m{3.5cm}}
    \toprule
\textbf{Model} & \textbf{Organization} & \textbf{Time} & \textbf{Version} & \makecell[c]{\textbf{Architecture Detail}\\(Base-Q-KV)} & \makecell[c]{\textbf{Context}\\ \textbf{Length}} & \makecell[c]{\textbf{Pre-Training}\\ \textbf{Strategy}} & \makecell[c]{\textbf{Post-Training}\\ \textbf{Strategy}} & \textbf{Benchmark} \\ \midrule

DeepSeek-V2 ~\citeyearpar{liu2024deepseek} & DeepSeek-AI & 24.05 & \makecell[c]{(default)\\Lite} & \makecell[c]{1e4-128Q-128KV$^\dagger$\\1e4-16Q-16KV$^\dagger$} & 128k & len=32k; YaRN & - & NIAH \\ 
DeepSeek-V2.5~\citeyearpar{liu2024deepseek} & DeepSeek-AI & 24.08 & (default) & 1e4-128Q-128KV$^\dagger$ & 128k & len=32k; YaRN & - & NIAH \\
DeepSeek-V3~\citeyearpar{Deepseek2024DeepSeek-V3} & DeepSeek-AI & 24.12 & (default) & 1e4-128Q-128KV$^\dagger$ & 128k & len=32k$\to$128k; YaRN & Distill long-CoT capacity from DeepSeek-R1 & LongBench-v2, LQA \\
DeepSeek-R1~\citeyearpar{guo2025deepseekr1} & DeepSeek-AI & 25.01 & \makecell[c]{(default)\\Zero} & \makecell[c]{1e4-128Q-128KV$^\dagger$\\1e4-128Q-128KV$^\dagger$} & 128K & YaRN & LongCoT; Long RL & -\\
\midrule

ChatGLM~\citeyearpar{glm2024chatglm} & Zhipu, THU & 23.05 & 6B & 1e4-32Q-32KV & 2k & len=2k & - & - \\ 
ChatGLM2~\citeyearpar{glm2024chatglm} & Zhipu, THU & 23.06 & 6B & 1e4-32Q-16KV & 32k & - & len=32k; long SFT; Positional Interpolation & ~ \\ 
ChatGLM3~\citeyearpar{glm2024chatglm} & Zhipu, THU & 23.1 & 6B & 1e4-32Q-16KV & 32k & - & - & - \\ 
GLM-4~\citeyearpar{glm2024chatglm} & Zhipu, THU & 24.06 & \makecell[c]{9B\\9B-chat\\9B-chat-1M} & 1e4-32Q-16KV & \makecell[c]{8k\\128k\\1M} & len=8k$\to$1M & LongAlign; multi-task long SFT & LongBench-chat \\ \midrule

InternLM2~\citeyearpar{cai2024internlm2} & InternLM & 23.12 & \makecell[c]{1.8B\\7B\\20B} & \makecell[c]{1e6-16Q-8KV\\1e6-32Q-8KV\\1e6-48Q-8KV} & 200k & len=4k$\to$32k; NTK; RoPE base=5e4$\to$1e6; & len=32k; long SFT; book and code data & L-Eval, LongBench, NIAH \\ 

InternLM2.5~\citeyearpar{InternLM25} & InternLM & 24.08 & \makecell[c]{1.8B\\7B\\20B} & \makecell[c]{1e6-16Q-8KV\\5e7-32Q-8KV\\5e7-48Q-8KV} & 1M & len=1M & - & - \\ 

InternLM3$^\diamond$~\citeyearpar{InternLM3} & InternLM & 25.01 & 8B & \makecell[c]{5e7-32Q-2KV\\factor 6} & 1M & - & - & RULER \\
\midrule

Yi~\citeyearpar{young2024yi} & 01.AI & 23.11 & \makecell[c]{6B\\9B\\34B} & \makecell[c]{5e6-32Q-4KV\\1e7-32Q-4KV\\1e7-56Q-8KV} & \makecell[c]{200k\\200k\\200k} & len=4k; NTK from 4k to 200k; book and synthetic data & Long SFT; synthetic data & NIAH \\ 
Yi-1.5~\citeyearpar{young2024yi} & 01.AI & 24.05 & \makecell[c]{6B\\9B\\34B} & \makecell[c]{5e6-32Q-4KV\\5e6-32Q-4KV\\5e6-56Q-8KV} & \makecell[c]{4k\\32k\\32k} & - & - & - \\ \midrule

Baichuan~\citeyearpar{baichuan7b2023} & Baichuan-Inc & 23.06 & \makecell[c]{7B\\13B} & \makecell[c]{1e4-32Q-32KV\\1e4-40Q-40KV} & 4k & - & - & - \\ 
Baichuan2~\citeyearpar{yang2023baichuan}
 & Baichuan-Inc & 23.09 & \makecell[c]{7B\\13B} & \makecell[c]{1e4-32Q-32KV\\ALiBi-40Q-40KV} & 4k & - & - & - \\

\midrule

MiniCPM~\citeyearpar{hu2024minicpm} & OpenBMB & 24.02 & 2B & 1e5-36Q-36KV & 4k & - & - & - \\
MiniCPM2~\citeyearpar{hu2024minicpm} & OpenBMB & 24.04 & \makecell[c]{1B\\2B} & \makecell[c]{1e5-24Q-24KV\\1e6-36Q-36KV} & \makecell[c]{4k\\128k} & len=4k$\to$128k & Long SFT; synthetic long QA data & InfiniteBench \\
MiniCPM3~\citeyearpar{hu2024minicpm} & OpenBMB & 24.08 & 4B & 1e5-40Q-40KV & 32k & Long-RoPE & - & -\\ 
    \bottomrule
    \end{tabular}
    }
    }
    \caption{Continued table of Table \ref{table:open_source_model_p1}. The symbol $^\dagger$ indicates that MLA is used in this model.}
\label{table:open_source_model_p3}
\end{table}
\clearpage

\section{Long-Context Post-training}\label{sec9}

Based on the above long-context pre-training strategy, long-context LLMs are trained to understand the long context well. Subsequently, the post-training is introduced to ensure the LLMs can follow human instructions and preferences are problems that need to be addressed during post-training~\citep{dubey2024llama, bai2024longalign}. Long-context post-training methods can be classified into three categories based on the length of input and output: \textbf{\textit{Long-In-Short-Out}}, \textbf{\textit{Short-In-Long-Out}}, and \textbf{\textit{Long-In-Long-Out}}. Currently, there is a lack of research on Long-In-Long-Out, which is an important direction for future studies. Therefore, we will focus the following discussion on the Long-In-Short-Out and Short-In-Long-Out scenarios and add something beyond post-training later.

\subsection{Long-In-Short-Out}

In the post-training process of LLMs, task-specific data is typically constructed to enhance the LLM’s performance on particular tasks, with the data construction type determined by the method (Supervised Fine-Tuning, SFT or Reinforce Learning RL). In the Long-In-Short-Out scenario, due to the length of the input context, manual annotation is difficult, and thus, synthetic data is often used. This section will introduce common data construction methods for various tasks.
\begin{itemize}
    \item{\textbf{Instruction Following}}
Provided with long-context data, instructions are given to generate relevant responses~\citep{chenlonglora}, or prompts are used to guide the LLMs to generate corresponding instructions and responses~\citep{koksal2023longform, bai2024longalign}.
    \item{\textbf{DocQA}}
Given a long document, relevant questions and answers are generated using LLMs. These can be based on the entire document~\citep{kaili2024mdcure}, or on specific context segment~\citep{an2024make, xiong2024effective, dubey2024llama}. In some cases, questions are generated and information is retrieved to ensure the quality of the generated answers~\citep{anonymous2024longpo, yu2023training}. Some researchers use shorter context segments to construct QA pairs and then concatenate many short pieces to form a long document~\citep{li2024longsyntheticdata, young2024yi}. To ensure the quality of responses, LLMs are often asked to provide citations~\citep{zhang2024longcite}.
    \item{\textbf{Multi-Hop QA}}
Long-context multi-hop QA data can usually be formed by combining multiple single-hop QA data~\citep{trivedi2022musique}. When combining single-hop QAs, similarity or question relevance can be considered~\citep{chen2024essential} to ensure coherence in question generation. Some studies require LLMs to generate responses using methods such as CoT~\citep{wei2022chain} or citation to improve data quality~\citep{li2024making}.
    \item{\textbf{Summarization}}
Besides using manually written documents and summary data from the Internet, LLMs are also often used to summarize long contexts. One method is to split the long context into chunks and summarize them individually, then provide a final summary of the summaries~\citep{dubey2024llama, chatglm2024glmlong}. Another method is to summarize short contents first, then concatenate them into a longer document and summarize that~\citep{li2024longsyntheticdata}.
    \item{\textbf{Retrieve}}
Information is inserted into the long context, and questions are posed about the inserted information~\citep{niah}. Alternatively, several pieces of information are combined to create a long context, and a question is asked about specific information~\citep{xiong2024artificialneedlesrealhaystacks}.
\end{itemize}

Researchers have also studied how to filter data. LOGO~\citep{tang2024logo} scores the contribution of answers from different chunks to determine the quality of data samples. LongReward~\citep{zhang2024longreward} uses predefined rules to score the responses. GATEAU~\citep{si2024selecting} filters data based on the relevance of the final reply to the long document and the importance of certain parts in the response, giving high attention weights to crucial parts. This method has shown significant effects with only a small amount of data.

Some researchers have explored methods to improve long-context capabilities without constructing long-context data. SkipAlign~\citep{wu2024skipalign} modifies the position embedding indices in short-text data, training LLMs on short texts to give it the ability to handle long texts. ProLong~\citep{gao2024train} adjusts the data sources and proportions of long and short texts to find efficient long-context LLM training methods. It has been found that using only short-text instruction data can also help the LLMs perform well on long-text tasks.

\subsection{Short-In-Long-Out}

When the task is more complex or requires detailed steps, longer output is necessary to express thoughts and details~\citep{wei2022chain, yao2024tree}. Therefore, long output is also a key capability for long-context LLMs. Data construction in this field is challenging, and there is still insufficient research. Current data construction methods can be classified into three categories: backtranslation, planning, and iterative training.
\paragraph{Backtranslation}
In backtranslation, given the context and a response, the LLMs generates instruction data in reverse. This method leverages the long-context LLMs' strong ability to handle long inputs~\citep{pham2024suri}.
\paragraph{Planning}
Another method is planning, which involves breaking the task down into sub-tasks to reduce complexity, eventually solving the original task. Some researchers apply planning by breaking down writing tasks, first generating an outline and then using the outline to create segments that combine into the final text~\citep{bai2024longwriter, liang2024integrating}. \citet{li2024large} also uses planning to guide reasoning, improving LLMs' reasoning.
\paragraph{Iterative Training}
Iterative training is also a commonly used method for enhancing LLMs' capabilities in the post-training stage. Self-Lengthen~\citep{quan2024language} uses two LLMs, a Generator and an Extender. The Generator generates responses within a specified length range, and the Extender extends the content to the target length. The concatenated extended data is then used to train the next generation of Generator and Extender.
\paragraph{Long Thought}
Long-output tasks, especially long thought, have attracted particular attention. The success of generation strategies like CoT~\citep{wei2022chain} and ToT~\citep{yao2024tree} has shown that LLMs can fully utilize their reasoning capabilities to generate better results. OpenAI o1~\citep{OpenAI2024o1} further enhances reasoning ability with CoT~\citep{wei2022chain}, achieving impressive results. More and more research is focused on how to achieve o1 or even better performance in long-context reasoning~\citep{zeng2024scaling, team2025kimi, guo2025deepseekr1}. \citet{qin2024o1, huang2024o1} improves the LLMs' reasoning ability using tree search and multi-agent strategies. ConTReGen~\citep{roy2024contregen} applies planning strategies in document QA by first generating sub-tasks from top-down and then retrieving relevant documents to solve the sub-tasks until the entire task is completed. K1.5~\citep{team2025kimi} and DeepSeek-R1~\citep{guo2025deepseekr1} significantly improves LLMs' reasoning ability through RL scaling. Long thought is an important task and we will discuss it in \textbf{\nameref{q9_output}} in Section\ref{sec12}

\subsection{Beyond Post-Training}

Besides post-training, many methods are being explored to enhance long-context LLMs. \textbf{\textit{Test Time Training}} (TTT) utilizes self-supervised learning during inference to further train LLMs using input test data~\citep{sun2020test, liang2024comprehensive}. Temp-Lora~\citep{wang2024greater} applies TTT in long-context scenarios by fine-tuning temporary Lora modules using contextual information during inference, guiding generation. Some works achieve alignment by providing examples or guidance during inference~\citep{sun2024principle, zhang2024metaalign, xie2023defending}, and long-context LLMs facilitate the effectiveness of these methods. Some researchers have effectively improved the performance of LLMs through \textbf{\textit{Test-Time Scaling}}~\citep{liao2024beyond, snell2024scaling}, proposing a new direction and further emphasizing the importance of long context. Additionally, LUQ~\citep{zhang2024luq} focuses on calibration for long-context LLMs, using NLI classifiers to determine the confidence of generated results and reducing uncertainty through model ensembling.

\section{Long-Context MLLM}\label{sec10}

\tikzstyle{leaf}=[my-box,
	text=black, align=left,font=\normalsize,
	inner xsep=2pt,
	inner ysep=4pt,
	line width=0.8pt,
        minimum height=1cm,
]
\begin{figure}
    \centering
\begin{tikzpicture}[
    font=\tiny,
    align=center,
    node distance=1cm and 1.5cm,
    quadrant/.style={draw, minimum width=3cm, minimum height=1cm, fill=yellow!8},
    empty/.style={draw, minimum width=3cm, minimum height=1cm},
    central/.style={circle, draw=orange!40, fill=orange!10, minimum size=2.5cm},
    axis/.style={thick}
]

\node[quadrant, inner sep=0pt, anchor=west] (long-docvlm) at (-4.5, 2) {
    \begin{minipage}[t][2.5cm][t]{2.5cm} 
        \vspace{1.5em} 
        \centering
        \textbf{Long DocVLM} 
        \vspace{0.5em} 

        \renewcommand{\arraystretch}{1.2} 
        
            \begin{tabular}{c}
            PDF-WuKong\\~\citep{xie2024wukong}\\
            mPLUG-DocOwl2\\~\citep{hu2024mplug}
            \end{tabular}
    \end{minipage}
};

\node[quadrant, inner sep=0pt, anchor=west] (high-resolution) at (-4.5, -1.0) {
    \begin{minipage}[t][2.5cm][t]{2.5cm} 
        \vspace{2em} 
        \centering
        \textbf{High-Resolution} 
        \textbf{ImageLLM}
        \vspace{0.5em} 

        \renewcommand{\arraystretch}{1.2} 
            \begin{tabular}{c}
            Monkey~\citep{li2024monkey}\\
            Sphinx~\citep{lin2023sphinx}
            \end{tabular}
    \end{minipage}
};

\node[quadrant, inner sep=0pt, anchor=west] (high-resolution) at (-4.5, -4) {
    \begin{minipage}[t][2.5cm][t]{2.5cm} 
        \vspace{1.5em} 
        \centering
        \textbf{Long CLIP} 
        \vspace{0.5em} 

        \renewcommand{\arraystretch}{1.2} 
            \begin{tabular}{c}
            LongCLIP\\~\citep{zhang2025long}\\
            VideoCLIP-XL\\~\citep{wang2024videoclip}
            \end{tabular}
    \end{minipage}
};

\node[quadrant, inner sep=0pt, anchor=north] (long-video) at (4, 3.25) {
    \begin{minipage}[t][8.5cm][t]{10cm} 
        \vspace{2.5em} 
        \centering
        \textbf{LongVideo} 
        \vspace{1.5em} 
        
        \resizebox{0.9\textwidth}{!}{ 
            \begin{forest}
                forked edges,
                for tree={
                    grow=east,
                    reversed=true,
                    anchor=base west,
                    parent anchor=east,
                    child anchor=west,
                    base=center,
                    font=\normalsize,
                    rectangle,
                    draw=black, 
                    rounded corners,
                    align=left,
                    text centered,
                    minimum width=4em,
                    minimum height=1cm,
                    edge+={darkgray, line width=1pt},
                    s sep=10pt,
                    inner xsep=2pt,
                    inner ysep=3pt,
                    line width=0.8pt,
                    ver/.style={rotate=90, child anchor=north, parent anchor=south, anchor=center, minimum width=18em, fill=gray!10},
                },
                where level=1{text width=20em, fill=green!10}{},
                where level=2{text width=20em, fill=orange!10}{},
                where level=3{text width=55em, fill=white}{},
                where level=4{text width=18em,}{},
                where level=5{text width=18em,}{},
                [
                    \textbf{LongVideo}, ver
                    [
                        \textbf{Input Adaptation}\\ \newline \textbf{(\S\ref{sec10_1})}
                        [
                            \textbf{Text-Only}
                            [
                                \quad {zhang2023simple}{,} LangRepo~\citep{kahatapitiya2024language}, leaf
                            ]
                        ]
                        [
                            \textbf{Image-Only}
                            [
                                \quad IG-VLM~\citep{kim2024image}{,} LongVA~\citep{zhang2406long}{,} \\ \quad IXC-2.5~\citep{zhang2024internlm}{,} FreeVA~\citep{wu2024freeva}, leaf
                            ]
                        ]
                        [
                            \textbf{Q-Former-based}
                            [
                                \quad MovieChat~\citep{song2024moviechat}{,} MA-LMM~\citep{he2024ma}{,} \\ \quad VidCompress~\citep{lan2024vidcompress}{,} TCR~\citep{korbar2025text}{,} \\ \quad Vista-LLaMA~\citep{ma2023vista}{,}  TimeChat~\citep{ren2024timechat} {,} \\ \quad LVChat~\citep{wang2024lvchat}{,}  Momentor~\citep{qianmomentor}, leaf
                            ]
                        ]
                        [
                            \textbf{Q-Former-free}
                            [
                                \quad LVNet~\citep{park2024too}{,} KeyVideoLLM~\citep{liang2024keyvideollm} \\ \quad Frame-Voyager~\citep{yu2024frame}{,} VideoStreaming~\citep{qian2024streaming}{,} \\ \quad SlowFast-LLaVA~\citep{xu2024slowfast}{,} TESTA~\citep{ren2023testa}{,} \\ \quad Videollama2~\citep{cheng2024videollama}, leaf
                            ]
                        ]
                    ]
                    [
                        \textbf{Architecture Adaptation}\\ \newline \textbf{(\S\ref{sec10_2})}
                        [
                            \textbf{Position Embedding}
                            [
                                \quad TC-LLaVA~\citep{gao2024tc}{,} V2PE~\citep{ge2024v2pe}{,} \\ \quad RoPE-Tie~\citep{kexuefm10040}{,} M-RoPE~\citep{wang2024qwen2,li2024giraffe}, leaf
                            ]
                        ]
                        [
                            \textbf{Cache Optimization}
                            [
                                \quad FastV~\citep{chen2025image}{,} PyramidDrop~\citep{xing2024pyramiddrop}\\ \quad ZipVL~\citep{he2024zipvl}{,} VL-Cache~\citep{tu2024vl}{,} \\ \quad Look-M~\citep{wan2024look}{,} ElasticCache~\citep{liu2025efficient}, leaf
                            ]
                        ]
                        [
                            \textbf{Architecture Innovation}
                            [
                                \quad S4ND~\citep{nguyen2022s4nd}{,} ViS4mer~\citep{islam2022vis4mer}{,} \\ \quad S5~\citep{wang2023s5}{,} VideoMambaSuite~\citep{chen2024video}{,} \\ \quad VideoMamba~\citep{li2025videomamba}{,} LongLLaVA~\citep{wang2024longllava}, leaf
                            ]
                        ]
                    ]
                    [
                        \textbf{Training and Evaluation}\\ \newline \textbf{(unfinished)} 
                        [
                            \textbf{Infrastructure}
                            [
                                \quad LWM~\citep{liu2024world}{,} LongVILA~\citep{xue2024longvila}{,}\\\quad RLT~\citep{choudhurydon}, leaf
                            ]
                        ]
                        [
                            \textbf{Training}
                            [
                                \quad TimeIT-125k~\citep{ren2024timechat}{,} Moment-10M~\citep{qianmomentor}{,}\\ \quad T2Vid~\citep{yin2024t2vid}{,} VISTA~\citep{ren2024vista}{,}\\ \quad Kangaroo~\citep{liu2024kangaroo}{,} Video-T3~\citep{li2024temporal}, leaf
                            ]
                        ]
                        [
                            \textbf{Evaluation}
                            [
                                \quad Egoschema~\citep{mangalam2023egoschema}{,} Video-MME~\citep{fu2024video}{,} \\ \quad MLVU~\citep{zhou2024mlvu}{,} MMBench-Video~\citep{fang2024mmbench}{,} \\ \quad LongVideoBench~\citep{wulongvideobench}{,} V-NIAH~\citep{zhang2406long} , leaf
                            ]
                        ]
                    ]
                ]
            \end{forest}
        }
    \end{minipage}
};

\end{tikzpicture}

	\caption{An overview of the long context in Multi-modal LLMs.}
    \label{fig:mllm}
\end{figure}
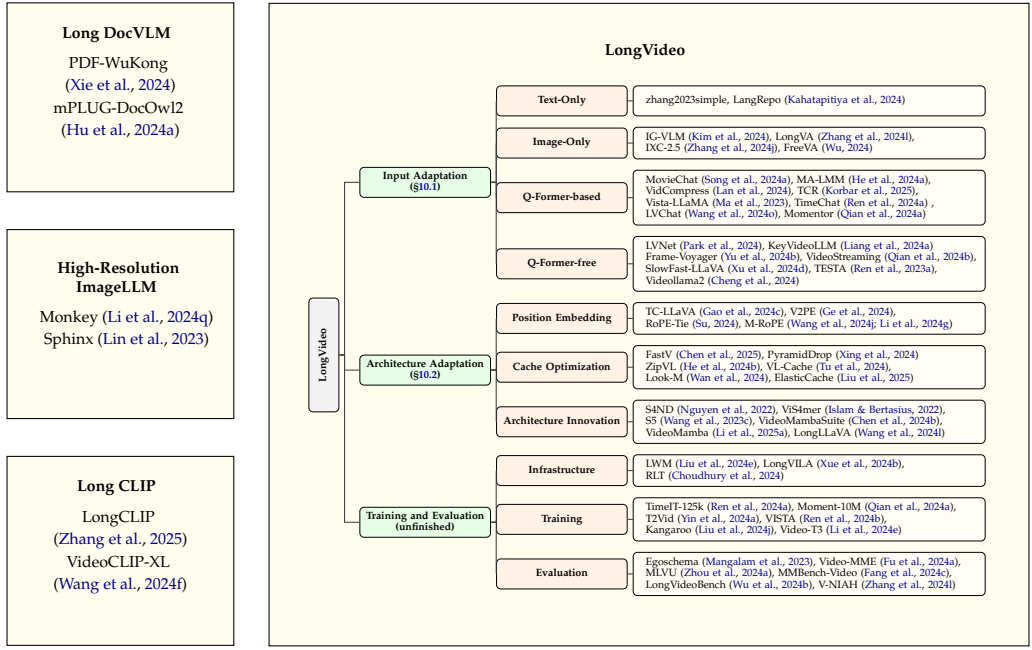

Based on the above technique, now we have numerous LLMs with strong long-context capabilities. But that is not the end. Long context is crucial for LLM focused on textual information and holds even greater significance for Multi-modal LLM (MLLM). In this section, we will change our focus on Long-context MLLMs. Long-context MLLMs involve various scenarios, including DocVLM for long documents with images~\citep{xie2024wukong,hu2024mplug}, ImageLLM for high-resolution images~\citep{li2024monkey,lin2023sphinx}, VideoLLM for long videos~\citep{zhang2406long,wang2024qwen2}, CLIP with long descriptions~\citep{zhang2025long,wulotlip}, as well as long speech models~\citep{reid2024gemini} and even world models\citep{liu2024world,zhan2024anygpt}. In this version, we only discuss long VideoLLM in detail. We omit speech for it is 1D like text and can be viewed as a new language~\citep{zhang2023speechgpt,zhang2024speechgpt}. Regarding long DocVLM, while it matters practically~\citep{jaisankar2024postdoc,zhang2024mgte,ma2024mmlongbench,chia2024m}, related discussions are relatively limited~\citep{xie2024wukong,hu2024mplug,blau2024gram,liu2024focus}.

Long CLIP and ImageLLM emphasize the extension of descriptive texts~\citep{zheng2025dreamlip,zhang2025long,wang2024videoclip} and images~\citep{li2024monkey,lin2023sphinx}, respectively. On one hand, the study of long CLIP could generally follow a length extrapolation discussion in text domain~\citep{zhang2025long,najdenkoska2024tulip}. On the other hand, while currently viewed as a long-context issue, high-resolution ImageLLM faces the backbone in the design of the vision encoder and may not necessarily remain a long-context problem in the future. In contrast, Long VideoLLM presents the most discussed long-context challenges in the MLLM due to its rich fine-grained spatiotemporal details as well as long-term dependencies in long videos~\citep{zou2024seconds,li2024giraffe}. Therefore, the following content will focus on the issues related to Long Video in detail.

Generally, long VideoLLMs and other long-context MLLMs training are after the textual pre-training and fine-tuning~\citep{liu2024world,zhang2406long}. Accordingly, this section focuses on how to obtain a long-context MLLM, especially a long VideoLLM, from a trained long-context LLM through input and architecture adaptation as well as multi-modal training. We will also discuss long video evaluations at the end of this section. This section can be viewed as a microcosm of the entire survey, emphasizing the differences and extensions of long-context in videos compared to text. Unlike text, videos have lower information density and are characterized by sample extraction~\citep{zou2024seconds}. Many studies concentrate on the extraction and compression of video information~\citep{song2024moviechat,ren2024timechat,qianmomentor,yu2024frame}. However, discussions on fine-grained alignment from long texts to long videos, such as the generalization of position embedding\citep{kexuefm10040,wang2024qwen2} and reasoning capabilities\citep{li2024temporal}, are relatively scarce.

\subsection{Input Adaptation}\label{sec10_1} 

The adaptation of long-context LLMs to long videos begins with input processing. Unlike text, which can be directly tokenized, video requires frame sampling, patch segmentation, and vision encoding and connector processing before entering the LLM~\citep{zou2024seconds}. For example, a one-hour video, sampled at 2 frames per second (fps) with 66 tokens per frame~\citep{wang2024qwen2}, results in over 400k tokens, while sampling at 1 fps with 2 tokens per frame~\citep{li2025llama} yields fewer than 8k tokens. Thus, the adaptation of video input affects the MLLM's context length, processing efficiency, and downstream performance. The input adaptation of long VideoLLMs has benefited from the redundancy of video~\citep{zou2024seconds}, utilizing both text-only~\citep{zhang2023simple} and image-only~\citep{kim2024image,zhang2406long} methods to replace original video inputs, as well as employing modules such as Q-Former~\citep{li2023blip,ma2023vista} to compress video data.

\paragraph{Text-Only} A relatively simple method for long video input is to truncate the long video into several short segments, convert them into corresponding text descriptions, concatenate these to form a complete video description and input it to the LLMs~\citep{zhang2023simple}. This approach was first proposed in LLoVi~\citep{zhang2023simple} and later improved by LangRepo~\citep{kahatapitiya2024language}, which iteratively processes video segments along with previous descriptions, eventually to generate a description for the entire video. Similarly, MVU~\citep{ranasinghe2024understanding} further simplifies the video into a combination of key information about objects. These methods do not input the original long video into the backbone LLM, thus reducing the adaptation and processing costs. Subsequent work has combined this approach with agents, enabling LLMs to collaborate with VLMs~\citep{wang2025videoagent} or interactively invoke tools~\citep{fan2025videoagenttool} for long video understanding.

\paragraph{Image-Only} Another class of methods treats videos as a comic strip, allowing LLMs to adapt to image features without additional long-video training. This line of work can be traced back to IG-VLM~\citep{kim2024image}, which achieves video understanding by transforming a video into a single image by arranging multiple frames in a grid. InternLM2-XComposer2.5~\citep{zhang2024internlm} (IXC-2.5 for simplicity) inherits this method and exhibits strong performance on various video benchmarks. Meanwhile, FreeVA~\citep{wu2024freeva} also demonstrates that using a similar approach without video training conditions enables ImageLLMs to process video. LongVA~\citep{zhang2406long} further proposes that the image-only method can transform an ImageLLM with long-context capabilities into a long VideoLLM capable of handling 2000 frames or over 200k visual tokens.

\paragraph{Q-Former-based} Besides the tricks mentioned above, early VideoLLMs tend to use cross-attention-based Q-Former~\citep{li2023blip,zhang2023video,song2024moviechat} to compress multi-modal information into fixed-length inputs, due to the significant redundancy in video representations~\citep{zou2024seconds}. However, in the scenario of long videos, Q-Former faces greater challenges in processing capacity, which introduce techniques including memory~\citep{song2024moviechat}, keyframe selection~\citep{korbar2025text}, Q-Former variants~\citep{ma2023vista,ren2024timechat}, and timestamp enbedding~\citep{ren2024timechat,qianmomentor}.

MovieChat~\citep{song2024moviechat} first introduces the memory mechanism into long VideoLLM, using a queue-based short-term memory and long-term memory based on adjacent fusion with higher frame similarity. Similarly, the concepts of compression and memory are also referenced in MA-LMM~\citep{he2024ma} and VidCompress~\citep{lan2024vidcompress}. Besides compression and memory, keyframe extraction is also an intuitive approach. For instance, TCR~\citep{korbar2025text} locates relevant information based on text and feeds it to Q-Former. TGB~\citep{wang2024efficient} uses the RoPE-involved product between optical flow and text embeddings to identify the ranges of multiple key content segments. Furthermore, some works attempt to overcome the fixed-length output constraint of Q-Former. For example, Vista-LLaMA~\citep{ma2023vista} proposes a SeqQ-Former similar to Transformer-XL~\citep{dai2019transformer}, and TimeChat~\citep{ren2024timechat} introduces a sliding window-based Q-Former. Notably, in LVChat~\citep{wang2024lvchat}, long videos are interleaved into multiple groups, encoded separately, and interleaved reversely to achieve complete encoding within a limit.

Although Q-Former faces limitations in processing long videos, the attention in Q-Former still facilitates the injection of information beyond images, particularly temporal and spatial embeddings~\citep{ren2024timechat,qianmomentor}. For example, TimeChat~\citep{ren2024timechat} employs a timestamp-aware frame encoder that explicitly binds visual content with corresponding timestamps in Q-Former with prompts. Similarly, TCR~\citep{korbar2025text} injects temporal information into visual representations through prompts based on special tokens after keyframe extraction. Momentor~\citep{qianmomentor} achieves the same goal with temporal embeddings in a continuous temporal token space. Some Q-Former-free models also adopt recurrent compression~\citep{wang2024videollamb} and temporal encoding. For instance, VideoStreaming~\citep{qian2024streaming} uses structures similar to RMT~\citep{bulatov2022recurrent} and LandmarkAttention~\citep{mohtashami2023random} to recurrently encode images, injecting corresponding temporal information with text prompts during the encoding process and recalling only relevant vision encoding segments during inference.

\paragraph{Q-Former-Free} In MLLMs, the LLaVA series~\citep{liu2024llava,liu2024llava1_5,liu2024llavanext,zhang2024llavanextvideo} first propose feeding the visual tokens from the vision encoder directly to the LLM backbone, avoiding the information bottleneck caused by Q-Former, which is inherited by many subsequent researches~\citep{li2024llava,wang2024qwen2,zhang2024internlm}. However, due to the high redundancy of video information, long VideoLLMs also introduce keyframe extraction and token compression to balance the compression of redundant information with the retention of key information~\citep{zou2024seconds,yu2024frame,cheng2024videollama}.

Regarding keyframe extraction, early attempts are often limited in uniform sampling of long videos~\citep{zhang2024llavanextvideo,cheng2024videollama}, resulting in low efficiency and high information loss~\citep{shen2024longvu}, prompting subsequent improvements. For instance, LVNet~\citep{park2024too} enhances keyframe extraction efficiency through a hierarchical keyframe selector, while KeyVideoLLM~\citep{liang2024keyvideollm} employs frame clustering to find central frames and integrates keyframe extraction in instruction fine-tuning. Frame-Voyager~\citep{yu2024frame} trains a frame extraction module by enumerating all possible frame extractions, discovering that extracting only 8 keyframes can achieve good understanding.

Regarding token compression, unlike text-only LLMs, which are limited to uni-modal, uni-dimensional, and unreadable compression, VideoLLM compression explores the differences in information density between image and text information, as well as how to integrate spatiotemporal information better. For the former, LLaMA-VID~\citep{li2025llama} leverages cross-attention between textual queries and visual features, arguing that one frame is worth two tokens in VideoLLM. SlowFast-LLaVA~\citep{xu2024slowfast} combines fine-grained slow features and coarse-grained fast features to achieve effective and efficient representation for detailed video understanding. For the latter, VideoLLM employs not only intuitive methods such as adjacent token merging in LLaVA-NeXT-Video~\citep{zhang2024llavanextvideo} and LongLLaVA~\citep{wang2024longllava}, hierarchical token merging~\citep{weng2025longvlm}, and average pooling~\citep{cai2024matryoshka}, but also more exquisite techniques like adaptive pooling in PLLaVA~\citep{xu2024pllava}, temporal-spatial aggregation in TESTA~\citep{ren2023testa} and LongVU~\citep{shen2024longvu}, and 3D convolution in Videollama2~\citep{cheng2024videollama}, Kangaroo~\citep{liu2024kangaroo}, Qwen2-VL~\citep{wang2024qwen2}.

\subsection{Model Adaptation}\label{sec10_2} 

\paragraph{Position Embedding} After inputting visual tokens into LLM, other problems arise, how to encode the relationship between visual tokens and textual tokens, and how to handle the extrapolation of visual tokens in the context of long videos~\citep{kexuefm10040,wang2024qwen2,li2024giraffe}. There are two schools of research regarding this. One ignores these questions or believes that VideoLLM can inherently perceive the spatiotemporal relationships of visual tokens without explicit representation in position embeddings~\citep{liu2024world,zhang2024llavanextvideo,chen2024internvl}. For long videos, in addition to directly applying existing text extrapolation methods~\citep{liu2024world,zhang2024llavanextvideo,shang2024intp} has also made some attempts. To avoid hallucinations caused by the increasing gap between video and text during generation, Vista-LLaMA~\citep{ma2023vista} does not apply RoPE to image tokens. TC-LLaVA~\citep{gao2024tc} improves downstream performance by varying the growth steps of image and text token indices and applying full attention to the same frame images. V2PE~\citep{ge2024v2pe} uses a similar approach to E$^2$-LLM~\citep{liu20242}, employing variable and smaller increments for visual tokens to enable the model to handle 1M long sequences under a 256k training setting.

The other group of research argues that images and videos possess additional spatiotemporal features compared with text, necessitating a more sophisticated position embedding schema for explicit representation~\citep{kexuefm10040,wang2024qwen2,li2024giraffe}. For instance, \citet{kexuefm10040} first proposes RoPE-Tie and conducts a comprehensive mathematical analysis. Subsequently, Qwen2-VL~\citep{wang2024qwen2} introduced M-RoPE, which unifies the position embedding of text, image, and video by decomposing the feature dimensions of text from low to high dimensions to represent time, height, and width. Regarding extrapolation, Giraffe~\citep{li2024giraffe} proposes M-RoPE++ by combining the three split intervals with YaRN~\citep{pengyarn} interpolation, achieving improved results. Recently, \citet{wei2025videorope} gives a depth-in analysis of what makes for good video rotary position embedding and proposes a new RoPE designed for video input.

\paragraph{Cache Optimization} The redundancy of multi-modal information is reflected not only in the sampling or compressing of input content but also in the sparsity of attention distribution~\citep{wan2024look,ma2023vista,tu2024vl}, which leads to the emergence of multi-modal cache optimization. However, since the compression of multi-modal information is more dominant in the input adaptation~\citep{song2024moviechat,ren2024timechat,qianmomentor,yu2024frame}, the exploration of KV cache optimization left for long VideoLLMs is relatively limited. Considering that KV cache optimization in the text domain has been discussed in Section~\ref{sec3}, here, we primarily analyze the work on long video KV cache optimization by cache dropping and merging~\citep{chen2025image,wan2024look}.

Regarding cache dropping, discussions in MLLMs are more centered on layer adaptation. For example, FastV~\citep{chen2025image} is the first to utilize LLMs' signal to guide the cache optimization, dropping visual tokens starting from the second layer of the MLLM in the inference phase. Similarly, PyramidDrop~\citep{xing2024pyramiddrop} emphasizes pruning more unimportant visual tokens as the layer goes up, and ZipVL~\citep{he2024zipvl} presents a layer-wise adaptive dropping ratio that boosts the overall compression ratio and accuracy compared to a fixed ratio. Notably, VL-Cache~\citep{tu2024vl} discovers that the attention patterns of MLLMs vary by modality, and therefore designing different sparsity levels for these patterns, adaptively adjusting the sparsity degree across layers. 

Regarding cache merging, related explorations focus more on the input side~\citep{li2025llama,shen2024longvu,lan2024vidcompress}, while there is less work on merging tokens within the attention block~\citep{wan2024look,liu2025efficient}. Interestingly, \citet{wan2024look} finds that the attention score for textual tokens is very dense, whereas the attention score for visual tokens is sparse. However, this finding contradicts the results from \citet{ma2023vista} and \citet{tu2024vl}, which show that the visual components are more attended.

\paragraph{Architecture Innovation} In the text domain, the emergence of RWKV~\citep{peng2023rwkv,choe2024rwkv} and SSM-Mamba~\citep{gu2020hippo,gu2023mamba,daotransformers} has led to new architectural innovations for LLMs, and similar research exists in the multi-modal field as well. Before the introduction of Mamba~\citep{gu2023mamba}, S4ND~\citep{nguyen2022s4nd}, ViS4mer~\citep{islam2022vis4mer}, and S5~\citep{wang2023s5} utilized S4~\citep{gu2021efficiently} blocks to capture long-context dependencies in video. After the introduction of Mamba~\citep{gu2023mamba}, there have been efforts to model video using Mamba~\citep{li2025videomamba,chen2024video}. Besides, there are also long video hybrid architectures, including LongLLaVA~\citep{wang2024longllava}, and efficient attention approaches for long videos, such as VideoTree~\citep{wang2024videotree}.





\section{Long-Context Evaluation}\label{sec11}

We finally come to the part of the long-context evaluation, which is an important technique of long-context LLM~\citep{an2023eval,bai2023longbench,zhang2024bench,niah,hsieh2024ruler,yen2024helmet}. Before the mainstream length extrapolation methods emerged, long-context evaluation primarily includes four assessment methods. The first is language modeling perplexity, typically on datasets like WikiText~\citep{merity2022pointer} or PG19~\citep{rae2019compressive}. The second is Long-Range Arean (LRA)~\citep{taylong}, testing whether models can capture the underlying structure through artificially constructed sequences. Furthermore, LongEval~\citep{longchat} assesses the retrieval ability of LLM across different context lengths through coarse-grained topic retrieval and fine-grained line retrieval. The only benchmark based on natural long texts to reflect the actual downstream performance is Scrolls~\citep{shaham2022scrolls}, along with its upgraded version ZeroScrolls~\citep{shaham2023zeroscrolls}, which enrich the longer data samples in existing QA and summarization tasks.

With the development of long-context LLMs, researchers construct more benchmarks, as shown in Table \ref{eval_benchmark_feat} and \ref{eval_task_type}. In this section, we will introduce the development of long-context evaluation from two perspectives, \textbf{\textit{type of tasks}} and \textbf{\textit{benchmark features}}. In this process, we will reveal the pain point of long-context evaluation that persists from early explorations. If real texts are used to construct tasks, while they can reflect long-context scenarios more authentically, the evaluation length cannot scale automatically and a careful metric design is necessary~\citep{zhang2023movqa,xu2024detectiveqa,yen2024helmet}. In contrast, if synthetic data are used, although lengths and metrics can be easily controlled, it is challenging to ensure that they are consistent with real-world scenarios~\citep{hsieh2024ruler,li2024needlebench}.

\begin{table}[!ht]
\renewcommand{\arraystretch}{1.35}
\tabcolsep=0.1cm
\centering
\small
\begin{tabular}{lcccccccc}
\toprule
\multirow{2}{*}{\textbf{Name}}
 & \multirow{2}{*}{\textbf{Time}} & \multicolumn{7}{c}{\textbf{Benchmark Feature}} \\ \cmidrule(lr){3-9}
& & \textbf{Len.} & \textbf{Lang.} & \textbf{Flexible} & \textbf{Stable} & \textbf{D.C.} & \textbf{Align.} & \textbf{L.O.} \\
\midrule
Scroll~\citep{shaham2022scrolls} & 22.01 & $\sim$8k & En & \xmark & \xmark & \xmark & \xmark & \xmark \\
ZeroScrolls~\citep{shaham2023zeroscrolls} & 23.05 & $\sim$8k & En & \xmark & \xmark & \xmark & \xmark & \xmark \\
LEval~\citep{an2023eval} & 23.07 & 4k-60k & En & \xmark & \cmark & \xmark & \xmark & \xmark \\
LongBench~\citep{bai2023longbench} & 23.08 & $\sim$10k & En, Zh & \xmark & \xmark & \xmark & \xmark & \xmark \\
BAMBOO~\citep{dong2024bamboo} & 23.09 & 4k-16k & En & \omark & \cmark & \cmark & \cmark & \xmark \\
M4LE~\citep{kwan2023m4le} & 23.10 & 1k-128k & En, Zh & \omark & \xmark & \xmark & \xmark & \xmark \\
LooGLE~\citep{li2023loogle} & 23.11 & $\sim$20k & En & \omark & \xmark & \cmark & \xmark & \xmark \\
Marathon~\citep{zhang2023marathon} & 23.12 & $\sim$80k & En & \xmark & \cmark & \xmark & \xmark & \xmark \\
\makecell[l]{Needle-In-A-Haystack\\~\citep{niah}} & 23.11 & 1k-128k & En & \cmark & \cmark & \xmark & \xmark & \xmark \\
InfiniteBench~\citep{zhang2024bench} & 24.02 & $\sim$200k & En, Zh & \xmark & \xmark & \xmark & \xmark & \cmark \\
LV-Eval~\citep{yuan2024lv} & 24.02 & 16k-56k & En & \cmark & \cmark & \cmark & \xmark & \xmark \\
Multi-NIHA~\citep{reid2024gemini} & 24.03 & 1k-1M & En & \cmark & \cmark & \xmark & \cmark & \xmark \\
CLongEval~\citep{qiu2024clongeval} & 24.03 & 1k-100k & Zh & \omark & \xmark & \xmark & \xmark & \xmark \\
LongICLBench~\citep{li2024long} & 24.04 & 2k-50k & En & \cmark & \cmark & \xmark & \xmark & \xmark \\
XL2Bench~\citep{ni2024xl} & 24.04 & $\sim$200k & En, Zh & \xmark & \xmark & \cmark & \xmark & \xmark \\
RULER~\citep{hsieh2024ruler} & 24.04 & 4k-1M & En & \cmark & \cmark & \xmark & \xmark & \xmark \\
Ada-LEval~\citep{wang2024ada} & 24.04 & 2k-128k & En & \omark & \cmark & \xmark & \xmark & \xmark \\
LoFT~\citep{lee2024can} & 24.06 & 32k-1M & \makecell{En, Es, Fr,\\Hi, Zh} & \omark & \cmark & \xmark & \xmark & \xmark \\
Loong~\citep{wang2024leave} & 24.06 & 10k-250k & En, Zh & \omark & \cmark & \cmark & \xmark & \xmark \\
BABILong~\citep{kuratov2024babilong} & 24.06 & 4k$\sim$10M & En & \cmark & \cmark & \cmark & \xmark & \xmark \\
LongIns~\citep{gavin2024longins} & 24.06 & 256-16k & En & \cmark & \cmark & \xmark & \cmark & \xmark \\
NeedleBench~\citep{li2024needlebench} & 24.07 & 20k-1M & En, Zh & \cmark & \cmark & \xmark & \cmark & \xmark \\ 
HelloBench~\citep{que2024hellobench} & 24.09 & $\sim$2k & En & \xmark & \cmark & \xmark & \cmark & \cmark \\
LongGenBench$_1$~\citep{wu2024longgenbench} & 24.09 & $\sim$20k & En & \omark & \cmark & \xmark & \cmark & \cmark \\
LongGenBench$_2$~\citep{liu2024longgenbench} & 24.10 & 4k-128k & En & \cmark & \cmark & \xmark & \cmark & \cmark \\
HELMET~\citep{yen2024helmet} & 24.10 & 8k-128k & En & \omark & \cmark & \xmark & \xmark & \xmark \\
\makecell[l]{LongSafetyBench\\~\citep{huang2024longsafetybench}} & 24.11 & $\sim$40k & En & \xmark & \cmark & \xmark & \cmark & \xmark \\
LIFBench~\citep{wu2024lifbench} & 24.11 & 4k-128k & En & \cmark & \cmark & \xmark & \cmark & \xmark \\
LongBench v2~\citep{bai2024longbench} & 24.12 & 32k-128k & En, Zh & \omark & \cmark & \xmark & \xmark & \xmark \\ 
LongProc~\citep{ye2025longproc} & 25.01 & 500~8k & En & \omark & \cmark & \xmark & \cmark & \cmark \\ 
\bottomrule
\end{tabular}
\caption{Comparison of the mainstream or comprehensive long-context benchmarks at present. The comparison includes the benchmark features such as average length, language, etc., and type of tasks including QA, summary, and retrieval in the continued table. In this table, Flexible stands for whether the length of evaluating data is flexible. Stable stands for stable evaluation. D.C. stands for data contamination. Align. stands for containing alignment tasks. L.O. stands for long output. \cmark~means yes, while \xmark~means no, and \omark~means the data in the benchmark are grouped into subsets by different length ranges. \label{eval_benchmark_feat}}
\end{table}

\begin{table}[!ht]
\renewcommand{\arraystretch}{1.35}
\tabcolsep=0.1cm
\centering
\small
\begin{tabular}{lcccccccc}
\toprule
\multirow{2}{*}{\textbf{Name}} & \multicolumn{8}{c}{\textbf{Type of tasks}} \\ \cmidrule(lr){2-9}
& \textbf{QA} & \textbf{Summ.} & \textbf{Retrieval} & \textbf{Code} & \textbf{Math} & \textbf{Agg.} & \textbf{ICL} & \textbf{Reasoning} \\
\midrule
Scroll~\citep{shaham2022scrolls} & \cmark & \cmark & \xmark & \xmark & \xmark & \xmark & \xmark & \xmark \\ 
ZeroScrolls~\citep{shaham2023zeroscrolls} & \cmark & \cmark & \xmark & \xmark & \xmark & \cmark & \xmark & \xmark \\ 
LEval~\citep{an2023eval} & \cmark & \cmark & \cmark & \cmark & \cmark & \xmark & \cmark & \xmark \\ 
LongBench~\citep{bai2023longbench} & \cmark & \cmark & \cmark & \cmark & \xmark & \xmark & \cmark & \xmark \\ 
BAMBOO~\citep{dong2024bamboo} & \cmark & \xmark & \xmark & \cmark & \xmark & \cmark & \xmark & \xmark \\ 
M4LE~\citep{kwan2023m4le} & \cmark & \cmark & \cmark & \xmark & \xmark & \xmark & \xmark & \xmark \\ 
LooGLE~\citep{li2023loogle} & \cmark & \cmark & \cmark & \xmark & \xmark & \cmark & \xmark & \cmark \\ 
Marathon~\citep{zhang2023marathon} & \cmark & \xmark & \cmark & \xmark & \cmark & \cmark & \xmark & \cmark \\
\makecell[l]{Needle-In-A-Haystack\\~\citep{niah}} & \xmark & \xmark & \cmark & \xmark & \xmark& \xmark & \xmark & \xmark \\
InfiniteBench~\citep{zhang2024bench} & \cmark & \cmark & \cmark & \cmark & \cmark & \cmark & \xmark & \xmark \\ 
LV-Eval~\citep{yuan2024lv} & \cmark & \xmark & \cmark & \xmark & \xmark & \xmark & \xmark & \xmark \\ 
Multi-NIHA~\citep{reid2024gemini} & \xmark & \xmark & \cmark & \xmark & \xmark& \xmark & \xmark & \xmark \\
CLongEval~\citep{qiu2024clongeval} & \cmark & \cmark & \cmark & \xmark & \xmark & \xmark & \xmark & \xmark \\ 
LongICLBench~\citep{li2024long} & \cmark & \xmark & \xmark & \xmark & \xmark & \xmark & \cmark & \xmark \\ 
XL2Bench~\citep{ni2024xl} & \cmark & \cmark & \cmark & \xmark & \xmark & \xmark & \xmark & \xmark \\ 
RULER~\citep{hsieh2024ruler} & \cmark & \xmark & \cmark & \xmark & \xmark & \cmark & \xmark & \xmark \\ 
Ada-LEval~\citep{wang2024ada} & \xmark & \xmark & \xmark & \xmark & \xmark & \cmark & \xmark & \xmark \\ 
LoFT~\citep{lee2024can} & \cmark & \xmark & \cmark & \cmark & \xmark & \xmark & \cmark & \xmark \\ 
Loong~\citep{wang2024leave} & \cmark & \xmark & \cmark & \xmark & \xmark & \cmark & \xmark & \cmark \\ 
BABILong~\citep{kuratov2024babilong} & \cmark & \xmark & \cmark & \xmark & \xmark & \xmark & \xmark & \cmark \\ 
LongIns~\citep{gavin2024longins} & \cmark & \xmark & \cmark & \xmark & \xmark & \xmark & \xmark & \xmark \\ 
NeedleBench~\citep{li2024needlebench} & \cmark & \xmark & \cmark & \xmark & \xmark & \xmark & \xmark & \cmark \\
HelloBench~\citep{que2024hellobench} & \cmark & \cmark & \xmark & \xmark & \xmark & \xmark & \xmark & \xmark \\ 
LongGenBench$_1$~\citep{wu2024longgenbench} & \cmark & \xmark & \xmark & \xmark & \cmark & \xmark & \xmark & \cmark \\ 
LongGenBench$_2$~\citep{liu2024longgenbench} & \cmark & \xmark & \xmark & \xmark & \cmark & \xmark & \xmark & \xmark \\ 
HELMET~\citep{yen2024helmet} & \cmark & \cmark & \cmark & \xmark & \xmark & \cmark & \cmark & \xmark \\ 
\makecell[l]{LongSafetyBench\\~\citep{huang2024longsafetybench}} & \cmark & \xmark & \cmark & \xmark & \xmark & \xmark & \cmark & \xmark \\ 
LIFBench~\citep{wu2024lifbench} & \cmark & \xmark & \cmark & \xmark & \xmark & \cmark & \xmark & \xmark \\ 
LongBench v2~\citep{bai2024longbench} & \cmark & \xmark & \xmark & \cmark & \xmark & \xmark & \cmark & \cmark \\ LongProc~\citep{ye2025longproc} & \cmark & \xmark & \xmark & \cmark & \cmark & \cmark & \xmark & \cmark \\ \bottomrule
\end{tabular}
\caption{The continued table of Table \ref{eval_benchmark_feat} comparing the type of tasks in the mainstream or comprehensive long-context benchmarks at present. QA stands for question-answer tasks. Summ. stands for summarization tasks. Retrieval stands for retrieval task. Code stands for code tasks. Math stands for math tasks. Agg. stands for aggregation tasks. ICL stands for long in-context learning tasks. Reasoning stands for reasoning tasks. \label{eval_task_type}}
\end{table}

\subsection{Type of Tasks} 

\paragraph{Long QA and Summary} The evaluation of long-context LLMs originated from long QA and summarization. Early long-context benchmarks including Scrolls~\citep{shaham2022scrolls}, ZeroScrolls~\citep{shaham2023zeroscrolls}, LEval~\citep{an2023eval}, and LongBench~\citep{bai2023longbench}, enrich the long-context data from QA (NarrativeQA~\citep{kovcisky2018narrativeqa}, QuALITY~\citep{pang2022quality}, Qasper~\citep{dasigi2021dataset}) and summarization (GovReport~\citep{huang2021efficient}, QMSum~\citep{zhong2021qmsum}) datasets as the main components of the evaluation. Based on this, different evaluations impose varying requirements on the tasks. LEval~\citep{an2023eval} and CLongEval~\citep{qiu2024clongeval} emphasize high-quality evaluation data, obtaining reliable long-context evaluation data through manual screening or annotation. M4LE~\citep{kwan2023m4le} highlights the diversity of data sources and categorizes long-context evaluation into five scenarios based on the distribution of answers in the text: explicit single-span, semantic single-span, explicit multiple-span, semantic multiple-span, and global context understanding. LooGLE~\citep{li2023loogle} proposes evaluating long-context and short-context dependencies simultaneously. LV-Eval~\citep{yuan2024lv} focuses on QA tasks by introducing confusing facts in the context to increase the difficulty.

\paragraph{Long-Context Retrieval} Retrieval is also a classic task in long-context evaluation, with early benchmarks such as LongEval\cite{longchat} emphasizing it. Retrieval tasks offer better flexibility than QA and summarization based on naturally long texts. Needle-In-A-Haystack (NIAH)~\citep{niah} is the first to propose reflecting LLM's recall performance in varying depths at varying context lengths. This sparks a surge in research on long-context retrieval tasks~\citep{young2024yi,cai2024internlm2,wang2024qwen2}, significantly altering the trajectory of long-context evaluation development. Notably, Gemini-1.5~\citep{reid2024gemini} expands the single-NIAH to multi-NIAH, achieving impressive results. Moreover, \citet{hsieh2024ruler} proposes various variants such as multikey and multivalue NIAH, creating an entirely synthetic long-context evaluation, RULER, which has become a new competitive focus among long-context LLMs~\citep{team2024jamba,liu2024retrievalattention,LFM}.

Furthermore, there are also domain-specific retrievals, such as DocFinQA~\citep{reddy2024docfinqa} and \citet{gupta2024systematic}, and structured data retrievals, namely enabling long-context LLMs to simulate SQL execution or database manipulation, including S3eval~\citep{lei2024s3eval}, BIRD~\citep{li2024can}, Spider 2.0~\citep{lei2024spider}, HoloBench~\citep{maekawa2024holistic}. To improve recall accuracy and assess whether LLM truly understands the context~\citep{gao2023enabling,hilgert2024evaluating,zhang2024longcite}, researchers also want LLM to provide relevant citations for the retrieval content~\citep{buchmann2024attribute,tang2024citeeval}. Such tasks have now been integrated into emerging long-context evaluation benchmarks, such as LoFT~\citep{lee2024can}, HELMET~\citep{yen2024helmet} and SCBench~\citep{li2024scbench}.

Due to the popularity of retrieval tasks, discussions on retrieval have also emerged. For example, \citet{liu2024lost} and \citet{an2023eval} highlight that LLMs tend to recall topics at the beginning and end of a context more easily and make mistakes with topics in the middle, thus demonstrating the Lost-In-the-Middle phenomenon. Furthermore, \citet{koo2024large} separates QA and evidence selection within retrieval from the perspective of task alignment. \citet{yu2024hyper} divides retrieval into matching and logical retrieval, exploring the corresponding improving methods. \citet{goldman2024really} analyzes long-context evaluation from the recall perspective and proposes two orthogonal dimensions, dispersion, and scope, to identify potential directions for more challenging long-context evaluations.

\paragraph{Code, Math, and Aggregation} In addition to tasks focusing on long natural language text, there are also long-context tasks centered on logical languages such as code and mathematics. Regarding code, LEval~\citep{an2023eval} and LongBench~\citep{bai2023longbench} are the first to incorporate code into long-context evaluation, which has been inherited by subsequent benchmarks~\citep{zhang2024bench,bai2024longbench}. Additionally, there are tasks specifically aimed at repository-level long code, such as RepoQA~\citep{liu2024repoqa}. Regarding math, LEval~\citep{an2023eval} and LongGenBench$_2$~\citep{liu2024longgenbench} expand the short-context task GSM8k~\citep{cobbe2021training} into a long-context task using many-shot ICL and question concatenation respectively. In contrast, Marathon~\citep{zhang2023marathon} and InfiniteBench~\citep{zhang2024bench} introduced more complex long-context mathematical computation tasks, while LongGenBench$_1$~\citep{wu2024longgenbench} examined the LLM's spatial-temporal understanding in long context.

Besides, there is also a category of long-context evaluation that includes sorting and statistics, generally referred to as aggregation tasks ~\citep{shaham2023zeroscrolls,hsieh2024ruler}. Aggregation tasks are first mentioned in LRA~\citep{taylong} and introduced into text evaluation in ZeroScrolls~\citep{shaham2023zeroscrolls}, which includes positive review statistics and summary sorting. After that, sorting tasks still exists in ~\citep{dong2024bamboo,li2023loogle,zhang2023marathon,zhang2024bench,wang2024ada}, while the recent HELMET evaluation suite also includes re-ranking tasks~\citep{yen2024helmet}. Regarding statistics, finding the maximum number and identifying~\citep{zhang2024bench} the most frequent words~\citep{hsieh2024ruler} are also proposed. Although aggregation tasks often occur in long-context benchmarks, they are less emphasized due to the deviation from natural long texts~\citep{hsieh2024ruler}.

\paragraph{Long In-Context Learning} Regarding LLMs, the two most notable capabilities are ICL~\citep{brown2020language,pan2023context} and reasoning~\citep{wei2022chain}, and long-context provides a deeper exploration of both. For ICL, a longer context enables more demonstrations, offering greater potential to stimulate LLM. Long ICL is first introduced in long-context evaluation in LEval~\citep{an2023eval} and LongBench~\citep{bai2023longbench}, primarily to extend the context of short-context tasks. After Gemini-1.5~\citep{reid2024gemini} prompts LLM to learn new languages with the grammar book and dictionary, long ICL has become a new focus for long-context evaluation~\citep{li2024long,agarwal2024many}. 

For example, LongICLBench~\citep{li2024long} evaluates a wide range of long-context LLMs and finds that most can benefit from extensive demonstrations when the length is within a certain range. As the input grows longer, it will lead to a performance fluctuation or decline~\citep{li2024long}. \citet{bertsch2024context} further points out that long ICL is sensitive to the distribution of demonstrations, and when there are enough demonstrations, the effect of the sampling method diminishes. Other studies indicate that long ICL is also influenced by the quality of the demonstrations~\citep{li2024demonstrations,agarwal2024many}, precision~\citep{wang2024precision}, retrieval~\citep{zou2024retrieval}, reasoning~\citep{kai2025mirbench} and other factors~\citep{agarwal2024many}. Additionally, \citet{wang2024benchmarking} propose General Purpose In-Context Learning, covering more domains including decision-making and world modeling through continuous generation and interaction. Long ICL has become a significant sub-item in emerging long document evaluation standards, such as LoFT~\citep{lee2024can} and HELMET~\citep{yen2024helmet}, and further discussions on long ICL will be present in \textbf{\nameref{q10_icl}} in Section\ref{sec12}. 


\paragraph{Long-Context Reasoning} The discussion of long reasoning can be traced back to early multi-hop reasoning tasks, such as HotpotQA~\citep{yang2018hotpotqa} and MuSiQue~\citep{trivedi2022musique}. The emergence of long context provides more exploration space for multi-hop reasoning. For example, Variable Tracing in RULER, CountingStars~\citep{song2024counting}, Loong~\citep{wang2024leave}, BABILong~\citep{kuratov2024babilong}, and Needlebench~\citep{li2024needlebench} ask models to aggregate multi-hop evidence inserted in the long-context when answering the final question. However, these evaluations still tend to focus on synthetic texts, lacking assessments of reasoning capabilities in real-world scenarios.

Apart from explicit multi-hop reasoning, some benchmarks~\citep{li2023loogle,zhang2023marathon,bai2024longbench}, also regard a deeper understanding of context as a type of reasoning. Recently, NovelQA~\citep{wang2024novelqa}, NoCha~\citep{karpinska2024one}, and DetectiveQA~\citep{xu2024detectiveqa} design reasoning evaluations for native long texts, leveraging the complex reasoning chains present in long novels. Moreover, NovelQA and DetectiveQA require LLM to output its reasoning process and conduct a process-centered evaluation~\citep{wang2024novelqa,xu2024detectiveqa}, which offers a more realistic and challenging evaluation. LongProc~\citep{ye2025longproc} however, uses long procedure generation to assess the model’s ability to handle long outputs and complex reasoning. More discussion on long-context reasoning and long output will be shown \textbf{\nameref{q9_output}} in Section\ref{sec12}.

In addition to the aforementioned evaluation tasks, there are long-context evaluations on other traditional NLP tasks. For example, some tasks in M4LE~\citep{kwan2023m4le} involve text classification. StNLab in CLongEval~\citep{qiu2024clongeval} explores the annotation issues in long Chinese texts. \citet{manikantan2024identifyme} and \citet{vodrahalli2024michelangelo} focuses on referential understanding within long texts.

\subsection{Benchmark Features}

\paragraph{Length} Length is an important feature for long-context evaluations. Before retrieval tasks like NIAH~\citep{niah,multi_niah} mark a turning point, the length of long-context evaluation benchmarks lags behind the lengths reported by long-context LLMs~\citep{pengyarn,young2024yi,cai2024internlm2}. This is primarily due to the limited native long-context corpora, making it difficult to enrich long-context evaluation~\citep{an2023eval,li2023loogle}. After this point, the situation reverses. On one hand, the length of synthetic tasks is flexible, and any length for evaluation is allowed ~\citep{liu2024reattention,lieber2024jamba}. On the other hand, researchers begin proposing more challenging evaluations~\citep{levy2024same,li2024long,hsieh2024ruler,gavin2024longins}, discovering that long-context LLMs fail to achieve acceptable performance within the claimed context lengths.

In addition to length itself, flexibility is a key feature of long-context evaluations~\citep{niah,yen2024helmet}. As mentioned earlier, traditional long-context benchmarks~\citep{an2023eval,bai2023longbench,zhang2024bench} are not scalable and only able to measure performance at different context lengths by truncating to various lengths~\citep{bai2023longbench}. Subsequent synthetic task evaluations~\citep{niah,levy2024same,hsieh2024ruler,liu2024longgenbench}, generally allow for customized context length. Additionally, some evaluation benchmarks~\citep{kwan2023m4le,lee2024can,yen2024helmet} group the evaluation data to different subsets by different length ranges, representing a trade-off.

\paragraph{Stability} Another important feature for long-context evaluation is stability, a persistent pain point in long-context evaluation~\citep{novikova2017we,an2023eval,yen2024helmet}. Specifically, it is difficult to provide a reliable evaluation metric for generative tasks such as long QA, summarization, and open-ended generation~\citep{an2023eval,novikova2017we} which are common in long-context benchmarks. In response, different long-context evaluation benchmarks have proposed various solutions. First, benchmarks like \citet{dong2024bamboo} avoid this issue by directly discarding long output tasks. Next, benchmarks like \citet{zhang2023marathon,lee2024can,bai2024longbench} address the problem by transforming generative answers into multiple-choice questions or constraining evaluation metrics.

Furthermore, some long-context research has delved more deeply into the stability of long output evaluations. LEval~\citep{an2023eval} is the first to propose using LLMs to compute reference-free, pairwise win rates to measure the quality of long outputs. After that, LV-Eval\cite{yuan2024lv} improves the stability of output measurement through a keyword-recall-based metric design without the aid of LLMs. In contrast, DetectiveQA~\citep{xu2024detectiveqa} introduced a step-wise reasoning metric that compares the reasoning chains of the model outputs to reference steps, measuring long reasoning based on the recall of reasoning steps. Similarly, HELMET~\citep{yen2024helmet} breaks down conventional summarization references into atomic claims and then has LLMs evaluate their recall. HelloBench~\citep{que2024hellobench}, based on the ordinary LLM-as-a-Judge~\citep{zheng2023judging}, decomposes answer quality into a linear combination of multiple scoring items from LLMs in a checklist, thereby reducing bias in LLM judges. Additionally, there are other solutions involving task formats, such as ProxyQA~\citep{tan2024proxyqa}, which evaluates a model's performance based on the outputs of the model under meta-questions to reflect the long generation capability of the model being tested, as well as \citet{liu2024longgenbench}.

\paragraph{Data Contamination} Evaluation benchmarks always face the issue of data contamination~\citep{golchintime}, and avoiding it is an important topic. In response, BAMBOO~\citep{dong2024bamboo} and LooGLE~\citep{li2023loogle} are the first to propose constructing evaluation sets using newly crawled data to mitigate this problem. Besides, LV-Eval~\citep{yuan2024lv} and XL2Bench~\citep{ni2024xl} employed keyword, phrase, and text replacement methods to address the issue. Additionally, DetectiveQA~\citep{xu2024detectiveqa} suggests comparing model performance under context-free scenarios to determine whether LLM relies on internal knowledge rather than context to answer questions. Finally, some studies~\citep{wang2024leave,kuratov2024babilong} claim that the data contamination may not exist for particularly long or very general texts.

\paragraph{Alignment Evaluation} Finally, some long-context evaluations also discuss the alignment performance of long-context LLMs. On one hand, LongIns~\citep{gavin2024longins} and LIFBench~\citep{wu2024lifbench} examine the instruction-following performance of long-context LLMs, with LongInsc\citep{gavin2024longins} reporting that the effective context length for instruction following is significantly shorter than the claimed context length of long-context LLMs. On the other hand, Many-shot Jailbreaking~\citep{anil2024many} focuses on the long-context safety performance of long-context LLMs, finding that numerous demonstrations can disrupt model alignment under long-context attacks. Subsequently, \citet{huang2024longsafetybench} and \citet{roberts2024needle} offer a broader discussion of long-context safety, exploring safety issues in various scenarios. Besides, some long-context evaluations investigate the memory capabilities of LLMs in real-world interactions~\citep{thonet2024elitr,wu2024longmemeval}. 

Additionally, there are some domain-specific long-context benchmarks such as \citet{hosseini2024benchmark} in the medical domain and \citet{reddy2024docfinqa} in the financial domain.

\section{Unanswered Questions}\label{sec12}

In the 10 sections above, we have illustrated the development trajectory of long-context from the extensive literature in different aspects. In this section, we can make a more comprehensive conclusion in the final section, unlike the previous survey focused on particular domains~\citep{huang2023advancing,zhao2023length,pawar2024and,luohekeep}. However, instead of listing some take-home messages or definitive claims, inspired by the masterpiece of Richard Strauss, we are more willing to end our journey with a longer context with 10 unanswered questions, to stimulate more in-depth thoughts and research on long-context LLMs from these perspectives. Whatever the answers may be, we believe we come out of reading this survey, wiser and better people than before. 

\paragraph{Q1}\label{q1_bias}\textbf{Position Bias}\quad
While considerable efforts have been devoted to augmenting the context window length of LLMs~\citep{chen2023extending, dynamicNTK, pengyarn}, position bias persists within these models. Position bias refers to LLMs' propensity to favor certain positions over others~\citep{wang2023large, zheng2023judging}.
A notable manifestation of this bias is the phenomenon known as \textbf{\textit{lost in the middle}}, where LLMs tend to allocate anomalously higher attention to the beginning and end of context, while the middle part receives relatively less focus~\citep{liu2024lost}. This tendency is further exacerbated by what has been termed the \textbf{\textit{attention sink}} effect, wherein the majority of attention scores are concentrated on the initial tokens of the context~\citep{xiaoefficient}. Surprisingly, such bias is observed even in NoPE-based LLM, where no position information is explicitly injected, but the performance of NIAH still declines from the middle~\citep{wang2024length}. On one hand, this bias has benefited research in streaming processing~\citep{xiaoefficient, yang2024seed} and KV cache optimization~\citep{tang2024razorattention, xiao2024duoattention}. On the other hand, many empirical efforts have also been devoted to addressing this bias~\citep{zhang2024attention, mcilroyorder,hsieh2024found}, such as fill-in-the-middle~\citep{an2024make}. However, minor studies try to answer why this bias exist~\citep{gu2024attention}. The theoretical understanding of related mechanisms is still an unanswered question.

In parallel, \citet{levy2024same} examines the impact of input length on the inference performance of LLMs, observing a significant performance decline though the input length is still shorter than the maximum context length. Leveraging this effect, some evaluation datasets have increased the length of evaluation texts to obscure relevant information and enhance the evaluation difficulty~\citep{yuan2024lv,hsieh2024ruler,li2024long}. However, questions regarding this aspect remain relatively unsolved. Though we can easily extrapolate the LLMs to a longer context, we often struggle to guarantee an exhaustive short-to-long generalization in downstream tasks~\citep{li2024long,anil2024many}.

\paragraph{Q2}\label{q2_rope}\textbf{RoPE Design}\quad
RoPE\citep{su2024roformer} has emerged as the mainstream position embedding for LLMs due to its superior performance\citep{dubey2024llama, bai2023qwen, liu2024deepseek}. However, regarding length extrapolation, the current RoPE scaling methods~\citep{roziere2023code,xiong2024effective}, can only achieve weak extrapolation for an infinite context length or strong extrapolation for a finite one. In strong extrapolation, RoPE-based LLMs rely on the high-dimensional, low-frequency features to represent long-context dependencies at greater distances~\citep{barbero2024round, hong2024token, zhong2024understanding}. However, these dimensions present OOD in extrapolation. Besides, even when position information is not OOD, the increased attention entropy also harms the long-context performance~\citep{pengyarn,han2024lm}. The conflicts between periodicity and monotonicity and between full attention and attention entropy are the inherent drawbacks of scaling RoPE-based LLMs to an infinite context~\citep{liuscaling,men2024base,han2024lm}. 

Given these limitations of RoPE, researchers have explored additional approaches based on alternative position embedding design~\citep{kazemnejad2024impact,wang2024length}, or cache operation~\citep{xiao2024infllm, liu2024reattention}, to address these challenges. Regarding RoPE itself, the modification for length extrapolation also simulates modifications for other perspectives, such as the selection of rotary angles~\citep{wu2024extending}, the number of dimensions for RoPE~\citep{glm2024chatglm,biderman2023pythia}, the index schema for RoPE~\citep{golovneva2024contextual}, whether there are better design alternatives for RoPE~\citep{sun2022length,chi2022kerple}, how the scaling laws change under these alternative designs, and even how to design RoPE for multi-modal information~\citep{kexuefm10040,wang2024qwen2,li2024giraffe,wei2025videorope}, all remain open questions await deeper investigation.

\paragraph{Q3}\label{q3_ppl}\textbf{Dilemma of Perplexity}\quad
For a long time, perplexity has been a primary indicator for determining the upper bound of the length extrapolation~\citep{presstrain,liuscaling}. However, subsequent research has found that perplexity does not truly reflect the performance of LLMs in the downstream tasks of long context~\citep{men2024base,hu2024can,fang2024wrong,gao2024train,xiaoefficient}. Despite this, there are still works that define long-context quality based on perplexity, such as LongWanjuan~\citep{lv2024longwanjuan} and ProLong~\citep{chen2024long} with perplexity-based metrics to compare the information gain of long contexts with short ones. Recently, LongPPL~\citep{fang2024wrong} based on the comparison between sliding window perplexity and standard perplexity, is proposed to reflect LLM's real downstream performance more accurately. 

However, both definitions of short-context perplexity have flaws: chunking breaks long-context dependencies while sliding windows imply that the receptive field increases with model depth. Additionally, the perplexity of different LLMs may vary due to differences in their training data distributions. Therefore, there is much space for improving perplexity in assessing LLMs performance and data quality in long-context scenarios.

\paragraph{Q4}\label{q4_rag}\textbf{Long Context v.s. RAG}\quad
The choice between long-context LLMs and RAG has been a topic of debate. \citet{xu2023retrieval} suggests that retrieval-augmented approaches allow LLMs with smaller context windows to perform on par with larger context window LLMs, and even improve the performance of long-context LLMs. However, \citet{li2024retrieval} has reached the opposite conclusion, indicating that under their experimental setup, long-context LLMs generally outperforms RAG. Moreover, \citet{leng2024long} indicates that using longer context does not uniformly increase RAG performance while \citet{jiang2024longrag} holds an opposite opinion. ~\citet{li2024long_vs_rag} conducts a more in-depth investigation into this issue. This raises two intriguing question: which paradigm represents the better approach for generation? Should these two paradigm be combined?

To begin with, long-context LLMs offer more complete contextual information compared to RAG, but they also come with challenges such as lower information density and high computational resource consumption.  KV cache is position-sensitive while RAG is position-independent. Whether the positional relationships within long-context LLMs play a significant role in generation remains an important topic for exploration~\citep{bertsch2024context}. In contrast, RAG is more lightweight and better suited for edge devices, but it is unable to handle special scenarios, such as long outputs. Many attention acceleration or approximation methods utilize retrieval~\citep{zhang2023h2o, li2024snapkv}, raising the question of whether long-context generation can be unified with RAG. We believe that a more flexible memory-based approach, which can leverage both text and KV cache, may represent a promising and potentially superior generation paradigm in the future~\citep{yang2024memory3}.

\paragraph{Q5}\label{q5_na}\textbf{Discussion on New Architecture}\quad
Recent advances in LLM's architectures have revealed an intriguing pattern: the incorporation of local interaction, such as the token shift in RWKV~\citep{peng2023rwkv,peng2024eagle} or convolution in Mamba~\citep{gu2023mamba}. While these architectural choices designed for a long context are different, they coincidentally introduce similar mechanisms to modeling local interaction. This raises the question of whether traditional RNN, LSTM, or SSM can achieve long-context capabilities comparable to Transformer by incorporating local interaction mechanisms. Furthermore, does the standard self-attention mechanism equal the combination of local interaction based on CNN or token shift and long-context dependency captured with RNN or SSM, and why or why not?

A potential explanation lies in the mechanisms of information processing. In attention-based architecture, information from different positions is processed in parallel before fusion~\citep{Vaswani2017attention}, whereas RNN, LSTM, and SSM architectures process information from various distances (both short and long-range) simultaneously~\citep{beck2024xlstm,gu2023mamba}. This mixed processing can lead to mutual interference, where short-context information may disrupt the modeling of long-context dependencies, and vice versa. The introduction of convolution or token shift represents an attempt to decouple information interaction across different scales. Moreover, such assumptions also need further validation.

\paragraph{Q6}\label{q6_slm}\textbf{On-Device Long Context}\quad 
The future of long-context LLMs also involves edge-based multi-modal applications, which require locally deployed models as a foundation or important support. Although major AI companies are integrating their models into local software~\citep{wu2024first, yin2024llm}, these solutions still rely on LLMs in the cloud. The future interaction paradigm will be fundamentally multi-modal~\citep{yao2024minicpm}, processing and generating across multiple modalities such as speech, images, text, and action sequences~\citep{Google2024long-context-usecase, Google2024why-do-they-matter, Apple2024AppleIntelligence, Google2024PixelGemini}. Meanwhile, to reduce latency, ensure privacy, balance server loads, and enable personalization, a substantial portion of computation and storage tasks of long-context LLMs will migrate closer to users, specifically to edge devices~\citep{wu2024first, xu2024device}.

Although the direction of development is clear, many challenges remain in delivering a smooth and natural long-context interaction experience to users. These challenges span multiple domains~\citep{xu2024device}: How can algorithms, hardware, and software be further optimized to reduce the resource footprint of long-context operations and improve inference speed~\citep{mlc-llm, lu2024bluelm, xue2024powerinfer, choe2024rwkv}? What technologies are needed for more seamless integration~\citep{yao2024minicpm, yin2024llm}? Is it possible to achieve horizontal scaling of long-context LLMs in this process and make it close to users? These challenges await researchers and engineers to solve them. Since the integration of large language models and edge devices has already become an industry-wide consensus~\citep{Qualcomm2023QualcommLlama, Apple2024AppleIntelligence, Qualcomm2024QualcommMistral, Google2024PixelGemini, lu2024bluelm}, we hope that they will all be resolved in the near future.

\paragraph{Q7}\label{q8_balance}\textbf{Long-Context Training from Scratch}\quad 
From a perspective of model capability, training with long-context data from the start offers several advantages. It naturally enhances LLM's ability to handle longer context~\citep{gao2024train}. Following the "the best part is no part" philosophy, it eliminates the need for complex length adaptation techniques. Training with mixed-length texts in the same batch allows LLMs to learn from text length distributions that better reflect real-world scenarios~\citep{gao2024train, chatglm2024glmlong}. 

The challenges of training with mixed-length sequences in the same batch are primarily engineering-related rather than theoretical. Traditional training frameworks require extensive padding when processing texts of varying lengths, which wastes computational resources and reduces training throughput. The disparity in computational load between long and short texts creates load imbalance issues in distributed training environments. Sophisticated runtime dynamic schedulers may be needed to address these challenges. Therefore, improving long-context training efficiency remains a critical engineering challenge, with a particular focus on enhancing the efficiency of mixed-length text training.

\paragraph{Q8}\label{q7_scarce}\textbf{Quantity and Quality of Long Data}\quad
As reported by Ilya, existing corpora have almost been exhausted for pre-training\footnote{Ilya Sutskever's talk at NeurIPS 2024. Sequence to Sequence Learning with Neural Networks. \url{https://www.youtube.com/watch?v=qo-ZjF_LAz8}}. The scarcity of data is more severe for long context~\citep{chatglm2024glmlong,gao2024quest}. Although researchers have constructed longer textual data by fancy concatenation~\citep{shicontext,chatglm2024glmlong,zhao2024longskywork} or task-oriented synthesis~\citep{an2024make,pham2024suri}, concerns about the effectiveness of synthetic data have never ceased~\citep{gao2024quest,zhao2024analysing,que2024hellobench}.

Besides quantity, quality also matters. Unfortunately, the definition of long-context data quality has not been thoroughly explored~\citep{lv2024longwanjuan,chen2024long}, and more researchers are trying to optimize the quality of data mixing between long and short corpora~\citep{xiong2024effective,chatglm2024glmlong,gao2024train}. However, training with limited long texts fails to guarantee the short-to-long generalization.~\citep{levy2024same,hsieh2024ruler,anil2024many,huang2024longsafetybench}, and how to guarantee effective training in long contexts is also a challenge~\citep{an2024does,gao2024train,hsieh2024ruler}.

In the multi-modal domain, the scarcity of long video is also significant~\citep{qianmomentor,yin2024t2vid,ren2024vista}. Moreover, although both text and video share sequential features, the generalization from long-context text to long-context video, and from long-context reasoning in text to long-context reasoning in video~\citep{li2024temporal}, remains a topic that requires further research and discussion.

\paragraph{Q9}\label{q9_output}\textbf{Long Output and Reasoning}\quad In the last two questions, we will finally discuss how to enhance the model capabilities with long context. Long context involves long input and long output and we start with the latter. Compared to short outputs, long outputs involve more complex dependencies and exposure bias resulting from inconsistencies between the previously generated content and the ongoing output~\citep{an2022cont}. These factors make training long-output LLMs particularly challenging. Although some evaluation work on long outputs has been conducted~\citep{tan2024proxyqa, que2024hellobench}, there is still a lack of effective metrics for assessing long outputs. Manual scoring remains subjective and difficult, and LLM-as-a-Judge requires further in-depth exploration~\citep{dubois2024length, que2024hellobench}. Thus, evaluating long-output LLMs remains a significant challenge.

Furthermore, the expectations for LLM outputs go beyond mere content generation. There is a need for LLMs to solve complex reasoning problems. The rapid rise of o1 has also highlighted the tremendous potential of long reasoning~\citep{OpenAI2024o1, zeng2024scaling, guo2025deepseekr1, team2025kimi}, and long output, as a core capability of long reasoning, needs to achieve better performance to support the advancement of related work. Moreover, \citet{snell2024scaling} indicates that scaling at test-time is crucial, and long output is a promising method to achieve it. Despite the vast application potential of long-output generation, it still faces numerous challenges. For instance, maintaining logical and informational consistency during long generation, and effectively controlling aspects such as style, tone, and emotion in the generated content, remain key issues for researchers~\citep{bai2024longwriter, quan2024language}. Additionally, these questions are more severe in MLLMs for multi-modal dependencies~\citep{tan2019lxmert, zhang2021visually} and cross-modal consistency~\citep{zhang2024cross, chou2024mm}.

\paragraph{Q10}\label{q10_icl} \textbf{Long In-Context Learning and More}\quad Long In-Context Learning is a method for enhancing LLM performance through long inputs~\citep{brown2020language,pan2023context,agarwal2024many}. Current discussions on long ICL mainly focus on benchmarks that use it to analyze long-context capabilities of LLMs~\citep{li2024long,wang2024benchmarking}. However, there is still a lack of attempts to treat long in-context learning as a means to overcome LLM limitations through long contexts~\citep{bertsch2024context,agarwal2024many}. Although some LLMs~\citep{reid2024gemini} have been shown to achieve translations of a brand new language using its grammar book and numerous demonstrations, discussions on related technical roadmap remain limited and require more open-source reproductions. Some studies also attempt to establish scaling or interpreting mechanisms between long ICL and SFT~\citep{dai2023can,mosbach2023few}, but certain theoretical analyses still need to be based on the separability assumption of softmax operations~\citep{dai2023can}, leaving a gap in practice.

Furthermore, some works also explore test-time training~\citep{sun2020test,sun2024learning}, the idea of training certain parameters of LLMs through a long context to enhance LLM capabilities or perceive user preferences. However, the context lengths involved in these works are still not sufficiently long and similarly lack corresponding scaling mechanisms. In the research on long outputs, concepts such as test-time scaling~\citep{snell2024scaling,OpenAI2024o1,zeng2024scaling} have emerged to enhance LLM performance by increasing the computational overhead of inference. However, the source of computational overhead, from long inputs or long outputs, is not clearly defined. Whether scaling inputs or scaling outputs yields more benefits is also a topic for discussion. Finally, these learning paradigms represent attempts to treat LLMs as humans, striving for ultimate life-long learning. This process will also compel us to rethink the architecture, infrastructure, and training strategy to suit these training paradigms, allowing LLMs to learn in the interactions until the Ewigkeit.


\section*{Acknowledgement}

This work is supported by the project cooperated with Huawei Noah's Ark Lab, \textit{Research on New and Efficient Architectures for Large-scale Language Models}, under the direction of Prof Qiu and Prof Guo in OpenMOSS FNLP as well as InternLM Team.

This survey is conducted for the self-validation of the author team, to present our sincere passion and commitment to long-context research, and to extract the development trajectory of long-context from the extensive literature.

The idea of combining this survey with a symphonic poem is proposed by Xiaoran Liu, inspired by the appreciation of "Thus Spake Zarathustra," to infuse this survey with soul and provide readers with an intuitive introduction.

If possible, I would like to dedicate this work to Prof Guo, who has worked hard over the past six months, to the memorable year with Boss Hang and Master Zhang in AI Lab, to the unforgettable moments in XinKingBo, and to the invaluable recognition from Prof Qiu.

Due to the authors' limited knowledge, this survey may contain omissions. We welcome constructive comments from readers. 

\bibliography{colm2024_conference}
\bibliographystyle{colm2024_conference}


\end{document}